\documentclass[a4paper,fleqn]{cas-dc}

\usepackage[numbers]{natbib}
\usepackage{multirow}
\usepackage{adjustbox}
\usepackage{hyperref}
\hypersetup{breaklinks=true}
\usepackage{cite}
\usepackage{amsmath,amssymb,amsfonts}
\usepackage{graphicx}
\usepackage{subcaption}
\usepackage{textcomp}
\usepackage{xcolor}
\usepackage{url}
\usepackage{float}
\usepackage{algorithm}
\usepackage{algorithmicx}
\usepackage[noend]{algpseudocode}%

\def\tsc#1{\csdef{#1}{\textsc{\lowercase{#1}}\xspace}}
\tsc{WGM}
\tsc{QE}
\tsc{EP}
\tsc{PMS}
\tsc{BEC}
\tsc{DE}

\begin{document}
\let\WriteBookmarks\relax
\def\floatpagepagefraction{1}
\def\textpagefraction{.001}

\shorttitle{Drift-Aware Variational Autoencoder-based Anomaly Detection with Two-level Ensembling}

\shortauthors{Jin Li et~al.}

\title [mode = title]{Drift-Aware Variational Autoencoder-based Anomaly Detection with Two-level Ensembling}                      

\tnotetext[1]{This paper was supported by the European Research Council (ERC) under grant agreement No 951424 (Water-Futures) and the Republic of Cyprus through the Deputy Ministry of Research, Innovation and Digital Policy.}


%
\author[1,2]{Jin Li}[
                        orcid=0000-0002-3534-524X]



\ead{li.jin@ucy.ac.cy}



\affiliation[1]{organization={KIOS Research and Innovation Center of Excellence},
    addressline={University of Cyprus}, 
    city={Nicosia},
    country={Cyprus}}

\author[1]{ Kleanthis Malialis} [orcid=0000-0003-3432-7434]
\cormark[1]
\ead{malialis.kleanthis@ucy.ac.cy}
\author[1,2]{ Christos G. Panayiotou}[orcid=0000-0002-6476-9025]

\ead{christosp@ucy.ac.cy}


\affiliation[2]{organization={Department of Electrical and Computer Engineering},
    addressline={University of Cyprus}, 
    city={Nicosia},
    country={Cyprus}}

\author[1,2]{ Marios M. Polycarpou}[orcid=0000-0001-6495-9171]
\ead{mpolycar@ucy.ac.cy}


\cortext[cor1]{Corresponding author}



\begin{abstract}
In today's digital world, the generation of vast amounts of streaming data in various domains has become ubiquitous. However, many of these data are unlabeled, making it challenging to identify events, particularly anomalies. This task becomes even more formidable in nonstationary environments where model performance can deteriorate over time due to concept drift. To address these challenges, this paper presents a novel method, VAE++ESDD, which employs incremental learning and two-level ensembling: an ensemble of Variational AutoEncoder(VAEs) for anomaly prediction, along with an ensemble of concept drift detectors. Each drift detector utilizes a statistical-based concept drift mechanism. To evaluate the effectiveness of VAE++ESDD, we conduct a comprehensive experimental study using real-world and synthetic datasets characterized by severely or extremely low anomalous rates and various drift characteristics. Our study reveals that the proposed method significantly outperforms both strong baselines and state-of-the-art methods.

\end{abstract}







\maketitle

\section{Introduction}

Over the past few years, there has been a significant surge in the amount of streaming data across diverse fields of application. However, incorporating online learning algorithms to handle streaming data in real-world scenarios poses significant obstacles, such as anomalous events, unavailability of labels, and nonstationary environments.

\textbf{Class imbalance}: The presence of infrequent/anomalous events can severely impact model performance, leading to biased predictions favoring the normal class \citep{krawczyk2016learning}. Handling this issue becomes even more intricate when working with streaming data.

\textbf{Unavailability of labels}: The majority of anomaly detection systems commonly utilize either signature-based methods or data mining-based methods, relying on labeled training data \citep{eskin2002geometric, gomes2017survey}. However, obtaining labeled data in real-time applications can be costly or, in some cases, labeled data may not be available at all.

\textbf{Nonstationary environments}: Frequently, we make the assumption that the process generating streaming data remains stationary. However, in numerous real-world scenarios, the generating process exhibits an inherent nonstationary phenomenon known as concept drift.
Concept drift may result from a variety of factors, including seasonal patterns, periodic fluctuations, evolving user preferences, or behavioral changes, among others \citep{ditzler2015learning,gama2014survey}.

These three challenges frequently coexist in real-world domains. 
In Water Distribution Networks \citep{eliades2024smart}, rare but critical events such as sensor faults or contamination coincide with scarce labeled data and demand-driven nonstationarity. 
In the financial sector \citep{ahmed2016survey}, fraudulent transactions are infrequent, labels for new fraud patterns are limited, and market conditions evolve rapidly. 
In the health domain \citep{ghassemi2018opportunities}, rare diseases and evolving patient conditions exacerbate labeling difficulties and nonstationary behavior. 
These cases illustrate the broad relevance of jointly addressing all three challenges.

\textbf{Interplay of challenges and illustrative applications:} In nonstationary streaming environments with limited label availability, class imbalance renders anomalies rare, while concept drift continuously reshapes normal behavior, making it difficult to distinguish true anomalies from natural distributional shifts. Consequently, drift detection and adaptation are essential to maintain alignment with evolving data distributions, reduce false alarms, prevent model degradation, and ensure long-term reliability under dynamic and imbalanced streams. These intertwined challenges arise across many real-world systems, including (a) WDN contamination monitoring, where sensor aging and operational changes induce drift while true contamination events remain rare; (b) industrial equipment monitoring, where vibration and temperature signals vary with operating conditions despite infrequent failures; and (c) network security monitoring, where benign traffic evolves with user behavior while attacks are scarce and often abrupt.

To address these challenges, an ideal predictive model should: (i)  effectively identify rare events or anomalies; (ii) adapt to concept drift to ensure sustained performance in nonstationary environments. (iii) learn from unlabelled data, considering that obtaining labeled data may not be feasible in certain real-world scenarios. While prior studies have addressed these challenges individually, they often co-occur and interact in real-world scenarios. Few online methods handle rare events, unlabeled data, and nonstationary dynamics simultaneously. Our approach addresses these challenges while also detecting the timing of distributional drifts. The main contributions of this study are as follows:

\begin{enumerate}

	\item We propose VAE++ESDD, a novel unsupervised online anomaly detection framework that leverages two-level ensembling to address nonstationary data streams. The framework combines incremental learning (++) and an explicit concept drift detection module (DD) for robust adaptation. The first level ensembles multiple VAEs for accurate detection, while the second level uses drift detection ensembling to enhance adaptability and reduce false alarms.

    \item We evaluate VAE++ESDD on diverse datasets using tfour metrics, including an ablation study of key components. Compared to baseline and state-of-the-art methods, our approach achieves superior performance, benefiting from its ensemble design: multiple VAEs capture variations in data distribution, while multiple detectors enhance drift detection accuracy.

\end{enumerate}

VAE++ESDD was first introduced in our brief conference paper \citep{li2023autoencoder} with the name strAEm++DD, and this work provides a significant extension. Unlike our preliminary paper: (i) We propose a method that incorporates ensembling in two ways—one with multiple incremental learners and the second with multiple concept drift detectors. (ii) Additional methodological differences include the use of VAEs, and the integration of an adaptive threshold for anomaly detection; (iii) We have included extensive ablation studies to examine the role of various components and hyper-parameters; (iv) An extensive study is performed comparing the proposed method to baselines and state-of-the-art methods; (v)	The “Related work” section has been considerably enriched.

The paper is organized as follows. Sec.~\ref{sec:background} provides the background material, necessary to understand the contributions of this paper. Related work is described in Sec.~\ref{sec:related}. The proposed method and it's computation analysis are presented in Sec.~\ref{sec:method}. The experimental setup is described in Sec.~\ref{sec:exp_setup}. The empirical analysis of the proposed approach and comparative studies of learning methods are provided in Sec.~\ref{sec:exp_results}. Some concluding remarks are discussed in Sec.~\ref{sec:conclusion}. For the reproducibility of our results, the data sets used and our code are made publicly available to the community\footnote{https://github.com/Jin000001/VAEESDD}.

\section{Preliminaries}\label{sec:background}

Online learning deals with a data generation process that provides a batch of examples at each time step $t$, denoted as $S = \{B^t\}_{t=1}^T$, where each batch is defined as $B^t = \{(x^t_i,y^t_i)\}^M_{i=1}$. The total number of steps is represented by $T \in [1, \infty)$, and the data is typically sampled from a long, potentially infinite sequence. The number of examples in each step is denoted as $M$. When $M=1$, it is referred to as \textbf{one-by-one online learning}, whereas for $M > 1$, it is known as \textbf{batch-by-batch online learning}  \citep{ditzler2015learning}. This work specifically focuses on one-by-one learning, denoted as $B^t = (x^t, y^t)$, which is particularly relevant for continuous or sequential data processing scenarios. $x^t \in \mathbb{R}^d$ represents a $d$-dimensional vector in the input space $x \subset \mathbb{R}^d$, $y^t \in \{1, ..., K\}$ represents the class label, and $K \geq 2$ is the number of classes. In the context of this work, which focuses on anomaly detection, the number of classes is $K=2$ (``normal'' and ``anomalous''). The examples are drawn from an unknown time-varying probability distribution $p^t(x,y)$, where $p^t$ denotes the joint distribution at time $t$ between the set of input variables $x$ and the target variable $y$.

In one-by-one online classification, a model $h$ receives an instance $x^t$ at time $t$ and produces a prediction $\hat{y}^t = h(x^t)$. In \textbf{online supervised learning}, the true label $y^t$ is then revealed and used to update the model. This process repeats at each time step. When the model is updated sequentially using newly arriving data without full retraining, i.e., $h^t = h^{t-1}.\text{train}(\text{instances})$, the process is referred to as \textbf{incremental learning}~\citep{losing2018incremental}.

In online data streaming applications, it is often impractical or impossible to obtain class labels in a timely manner. To address this, the research community has explored alternative learning methods. One such paradigm is \textbf{online semi-supervised learning} \citep{dyer2013compose}, which initially relies on a small portion of labeled data. Another approach is \textbf{online active learning} \citep{malialis2022nonstationary, vzliobaite2013active}, where strategies are employed to intelligently decide when to query a human expert for ground truth information, such as class labels, for selected examples. While effective, these paradigms rely on labeled data. In this study, we focus on one-by-one \textbf{online unsupervised learning}, which operates without class labels, i.e., $B^t = (x^t)$.

\textbf{Data nonstationarity} is a prominent challenge observed in certain streaming applications, often attributed to concept drift, which refers to a change in the underlying joint probability distribution \citep{ditzler2015learning, gama2014survey}. \textbf{Concept drift} can lead to shifts in the data characteristics over time. Specifically, the drift between two time steps $t_i$ and $t_j$, is defined as $p^{t_i}(x,y) \neq p^{t_j}(x,y)$. Data distribution changes, or drifts, over time can take various forms. First, \textit{abrupt drifts} occur suddenly, where one concept shifts immediately to another. Second, \textit{incremental drifts} unfold gradually, progressing through multiple intermediate concepts. Finally, \textit{recurrent drifts} may bring back previously observed concepts after a period of absence \citep{gama2014survey}. 


\section{Related Work}\label{sec:related}

\subsection{Learning in nonstationary environments}\label{sec:conceptdrift}

Two primary families of strategies are commonly employed for learning concept drift: passive and active approaches \citep{ditzler2015learning,han2022survey}. These families differ in their adaptation mechanisms for coping with changes. 

\subsubsection{Passive methods}
Passive approaches typically adapt models using incoming data without explicitly detecting drifts. They are often categorized into memory-based and ensemble-based methods. Memory-based models \citep{ditzler2015learning} employ sliding windows to retain recent data, which helps the model adapt incrementally.  Ensemble methods, such as DWM \citep{kolter2007dynamic}, maintain a pool of models that evolve over time based on performance. Methods have been extended to handle class imbalance, such as, separate memories per class \citep{malialis2018queue}, adaptive rebalancing \citep{malialis2020online}, few-shot learning \citep{malialis2022nonstationary}, data augmentation \citep{malialis2025siameseduo++}, and ensembling \citep{cano2022rose,wang2014resampling}. However, the vast majority of passive methods still rely on the availability of labelled data.

\subsubsection{Active Methods}

Active approaches trigger model adaptation by detecting changes in the data distribution~\citep{ditzler2015learning}. Such detection strategies generally fall into two categories: statistical tests that monitor distributional shifts in the input space~\citep{friedrich2023unsupervised, jaworski2020concept}, and threshold-based methods that track performance indicators such as prediction errors~\citep{li2024unsupervised, greco2024unsupervised, han2024concept, cerqueira2023studd}.

Autoencoders have been utilized as drift detectors. One approach based on autoencoders is presented in the work by Jaworski et al. \citep{jaworski2020concept}, which focuses on detecting concept drift by monitoring two distinct cost functions: the cross-entropy and the reconstruction error. The variation of these cost functions is utilized as an indicator for drift detection. Another example is the ADD method proposed by Menon et al. \citep{menon2020concept}. ADD is a threshold-based method for batch-by-batch data that detects concept drift using reconstruction errors. By comparing current and previous batch errors against a threshold, it identifies incremental or abrupt drifts and triggers retraining accordingly.

\subsection{Anomaly detection}\label{sec:anoamlydetection}
Anomaly detection in data analysis typically involves training a model on normal data to establish a baseline of ``normality", enabling deviations from this baseline to be identified as anomalies \citep{chandola2009anomaly}. Most studies remain limited to static databases rather than evolving data streams \citep{li2019faad}. In this study, we focus on anomaly detection in scenarios with limited labeled data, emphasizing high-dimensional rather than one-dimensional sensor signals. Accordingly, the methods presented are either in semi-unsupervised or unsupervised settings.

\subsubsection{Traditional machine learning methods}
Several techniques have been developed for anomaly detection. LOF \citep{breunig2000lof} identifies outliers by comparing the local density of a point to its neighbors. OC-SVM \citep{scholkopf2001estimating} constructs a boundary around normal data, classifying points outside the boundary as anomalies. iForest \citep{liu2008isolation} focuses on isolating anomalies rather than modeling normal data, using an ensemble of trees to detect anomalies based on shorter path lengths. Some recent works have been proposed. RRCF \citep{guha2016robust} is designed for dynamic data streams, leveraging random forests to detect anomalies. Two ensemble learning methods are proposed for anomaly detection. DELR \citep{zhang2019delr} is a two-tier anomaly detection framework that enhances generalization by preserving critical features across subspaces and combining model outputs via a weighted strategy. It also introduces a diversity loss to encourage varied model perspectives. \citep{yu2016bayesian} proposes a theoretical framework that employs Bayesian classifier combination for integrating individual detectors. This method allows for the straightforward interpretation of posterior distributions as anomaly probability distributions while also enabling the assessment of bias, variance, and individual error rates of the detectors. However, these two ensembling methods cannot adapt to evolving data streams.


\subsubsection{Deep learning methods}
Several deep learning approaches for anomaly detection have been proposed. Zhang et al.~\citep{zhang2021feature} introduce a feature-aligned stacked autoencoder that exploits unlabeled data during fine-tuning, while the Teacher--Student Uncertainty Autoencoder~\citep{yang2022teacher} reduces representation discrepancies for fault detection. Ensemble-based methods combine heterogeneous models to enhance robustness, such as ENAD~\citep{liao2021enad}, which integrates autoencoders and GANs via weighted model fusion, and GASN~\citep{chen2025generative}, a two-stage GAN-based framework that assigns anomaly scores through neighborhood analysis. Recent studies also focus on image-based anomaly detection~\citep{lin2025survey, cui2023survey}, employing adversarial autoencoders~\citep{zhang2022unsupervised}, multi-scale feature learning with augmentation~\citep{sun2024ramfae}, masked autoencoders with pseudo-abnormal modeling~\citep{georgescu2023masked}, and dual student--teacher distillation frameworks such as DMDD~\citep{liu2024dual}.

\subsection{Anomaly detection in nonstationary environments}\label{sec:anoamlydetectioncon}
While the above works address either concept drift or anomaly detection, this section focuses on approaches that consider both.

INSOMNIA~\citep{andresini2021insomnia} reduces update latency via incremental learning with actively estimated labels. PWPAE~\citep{yang2021pwpae} employs an ensemble of drift-adaptive models with performance-based weighting, but requires ground-truth labels or prediction errors for drift detection. Mustafa et al.~\citep{mustafa2017unsupervised} detect change points using statistical tests on deep feature representations and retrain models accordingly, though labeled data are needed for low-confidence instances. ReCDA~\citep{yang2024recda} combines self-supervised representation learning with weakly supervised classifier tuning for drift adaptation, but depends on anomalous data during offline training. The Streaming Autoencoder (SA)~\citep{dong2018threaded} adopts an ensemble of incrementally trained autoencoders and distinguishes drift from anomalies using buffered future instances, limiting real-time detection. ARCUS~\citep{yoon2022adaptive} performs drift-aware adaptation through adaptive model pooling, but operates in a batch-based manner. MemStream~\citep{bhatia2022memstream} leverages a denoising autoencoder with a memory module to adapt to evolving streams without labeled data. METER~\citep{zhu2023meter} addresses concept drift via a hypernetwork that generates parameter shifts guided by evidential learning. CPOCEDS~\citep{jafseer2024cpoceds} preserves past concepts by incorporating representative samples across windows and supports incremental class discovery, though we consider its binary setting here. SEAD~\citep{shah2025sead} maintains an ensemble of base detectors and adapts to concept drift by dynamically reweighting or replacing detectors based on anomaly frequency.


\subsubsection{Open challenges} 

Key challenges persist when anomaly detection must operate under data non-stationarity or concept drift. Most existing methods address either concept drift (Sec.~\ref{sec:conceptdrift}) or anomaly detection (Sec.~\ref{sec:anoamlydetection}) in isolation. Anomaly detection approaches typically assume a stationary normal data distribution, causing drifted normal samples to be misclassified as anomalies. Conversely, drift-handling methods often focus on adapting to evolving data distributions and do not explicitly consider anomaly discriminability or class imbalance. Although a few studies attempt to jointly address drift and anomaly detection (Sec.~\ref{sec:anoamlydetectioncon}), most rely on access to streaming ground-truth labels for drift detection or model adaptation \citep{andresini2021insomnia,mustafa2017unsupervised}, an assumption that is frequently violated in real-world monitoring scenarios.

To address these limitations, we propose a unified framework that explicitly incorporates concept drift detection into anomaly detection. By detecting drift and triggering timely model adaptation in a fully unsupervised manner, the proposed approach prevents post-drift normal samples from being misclassified as anomalies and preserves the separability between normal and anomalous data over time, thereby mitigating the reliance on stationarity assumptions and labeled data.

Some recent unsupervised methods, such as ARCUS \citep{yoon2022adaptive}, MemStream \citep{bhatia2022memstream}, METER \citep{zhu2023meter}, and strAEm++DD \citep{li2023autoencoder}, represent important steps toward addressing these challenges by combining anomaly detection with adaptation mechanisms. However, these approaches typically rely on single-model architectures, which may lack robustness under complex, evolving conditions. Among ensemble-based approaches, SA and ARCUS primarily rely on passive adaptation, whereas SEAD adapts to evolving concepts by dynamically reweighting or replacing detectors. In contrast, VAE++ESDD integrates both active and passive strategies, which is absent from existing ensemble-based approaches and enables more reliable reasoning under non-stationary and imbalanced data streams. Unlike prior ensembles that achieve diversity through data partitioning, model pooling, or performance-based voting, VAE++ESDD introduces multi-level diversity via heterogeneous window scales and employs reconstruction-loss–driven drift detectors that are more sensitive to subtle distributional shifts.


\begin{algorithm}[!h]

	\caption{VAE++ESDD}
	\label{alg:method1}
	\begin{algorithmic}[1]

		\Statex \textbf{Input:} 
        \State $D$; $W_{drift}(i)$; $W_{train}$; $\gamma$; $expiry\_time$; $D_{thre}$; $P_{thre}$;$P_{warn}$;$P_{alarm}$ (Descriptions can be found in Table~\ref{tab:varia}) 

		\Statex \textbf{Init:}
        \State{
        $ES_{warntrig}$; $ES_{alarmtrig}$; $mov_{ESwarn}$; $mov_{train}(i)$; $mov_{driftx}(i)$;}
        \State Initialize  $h(i)$;  $ref_{driftx}(i)$; $\theta^t(i)$ for $i \in \{1,...,n$\} with $D$

	  \Statex \textbf{Start:}
		\For{each time step $t \in [1,...,\infty)$}
        \State receive instance $x^t \in \mathbb{R}^d$
        \State append $x^t$ to $mov^t_{train}(i)$
        \For{$i \in \{1,...,n\}$}  
        \State append $x^t$ to $mov^t_{driftx}(i)$
        \State predict $\hat{y}^t(i) \in \{0,1\}$ and get $mov^t_{driftl}(i)$ \Comment Eq.~\ref{eq:predict} and \ref{eq:ref_mov_def}
         \State calculate $P^t(i)$ and raise $flag(i)$ \Comment Eq.~\ref{eq:u} to \ref{eq:pvalue}

        \If{$flag_{warn}(i)$ raised first time} 
        \State $ES_{warntrig} \gets t$
           \EndIf

        \If{$flag_{alarm}(i)$ raised first time} 
        \State $ES_{alarmtrig} \gets t$
           \EndIf

\If{$t \%  W_{train}(i) == 0$}
\Comment \textcolor{blue}{Incr. learning}

            \State train $h^t(i)$ and update $\theta^t(i)$ \Comment Eq.~\ref{eq:adtthreshold},\ref{eq:update}

        \EndIf
        \EndFor

        \If {$ES_{warntrig} > 0$ and $ES_{alarmtrig}==0$} 
            \State Buffer $x^t$ to $mov_{ESwarn}$
             \If{$ t-ES_{warntrig} > expiry\_time$}
            \State Reset $ES_{warntrig}$, clear $mov_{ESwarn}$ 
        \EndIf
        \EndIf
        $\hat{y}^t = 1$ if  $\sum_{i=1}^{n}(\hat{y}^t(i))>= P_{thre}$\Comment \textcolor{blue}{Prediction voting}

     \If {$\sum_{i=1}^{n}(flag_{alarm}(i))>=D_{thre}$} \Comment \textcolor{blue}{DD voting}
     \State update models and re-initialize windows


     \EndIf
\EndFor
    \end{algorithmic}
\end{algorithm}

\section{The VAE++ESDD Method}\label{sec:method}

\begin{figure*}[t!]
	\centering
	\includegraphics[scale=0.4]{ 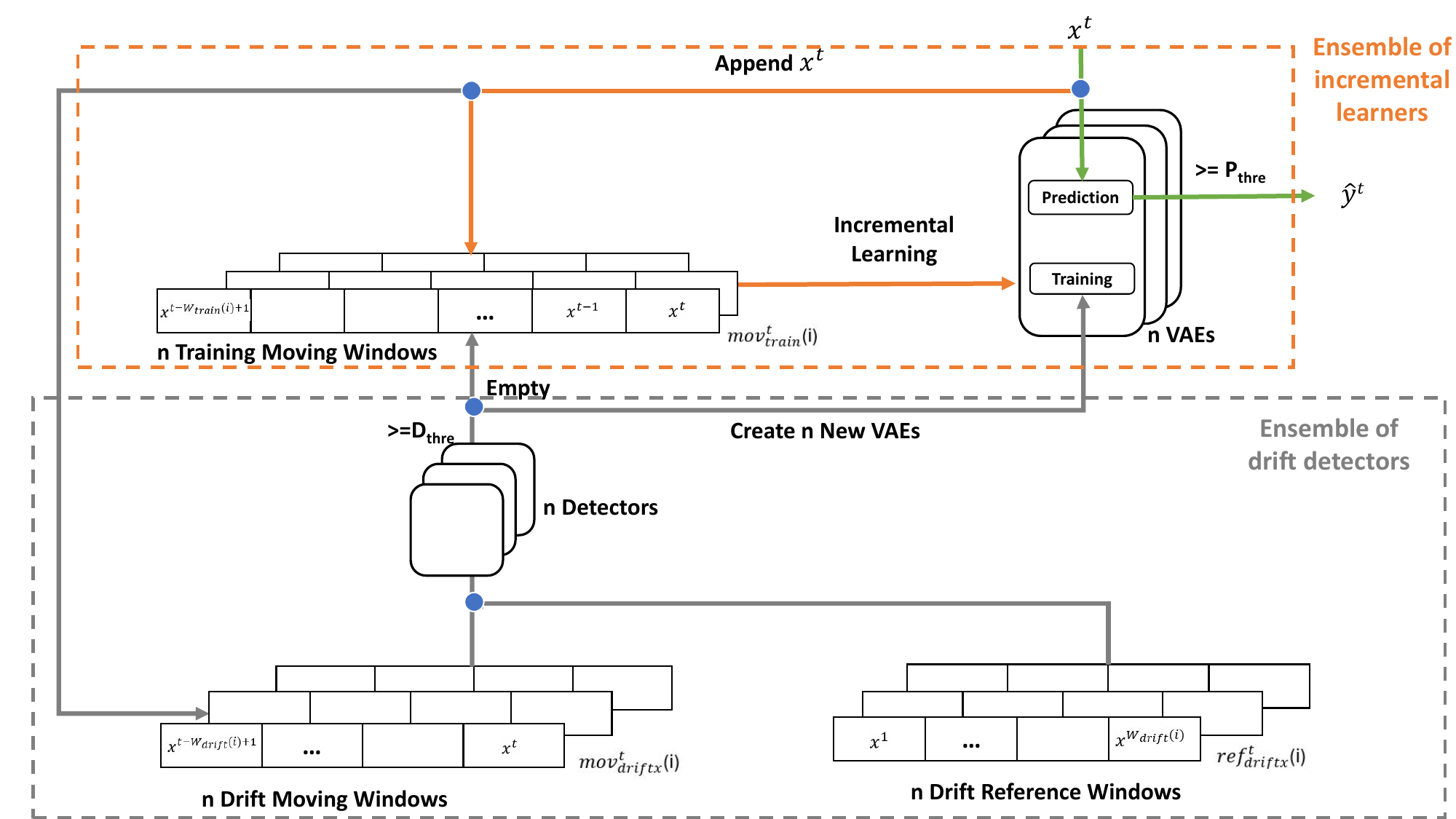}
	\caption{An overview of VAE++ESDD}
	\label{fig:VAE++ESDD}
\end{figure*}

\begin{table}[!t]

\caption{Nomenclature}\label{tab:varia}
\begin{adjustbox}{width=0.5\textwidth}
\begin{tabular}{cc}
\hline
\textbf{Variable} & \textbf{Description}                                                                                                                                                                   \\ \hline
\multicolumn{2}{c}{\textbf{Concept Drift Detection}}                                                                                                                                                       \\ \hline
$W_{drift}$       & window size for drift detection.                                                                                                                                                        \\
$mov_{driftx}$    & moving window of instances, size=$W_{drift}$.                                                                                                                                            \\
$ref_{driftx}$    & reference window of instances, size=$W_{drift}$.                                                                                                                                        \\
$mov_{driftl}$    & moving window of losses, size=$W_{drift}$.                                                                                                                                              \\
$ref_{driftl}$    & ref window of losses, size=$W_{drift}$.        

                                                              \\
$P_{warn}$    & threshold to trigger warn flag for each individual model.   

                                                              \\
$P_{alarm}$    & threshold to trigger alarm flag for each individual model.   

\\ \hline
\multicolumn{2}{c}{\textbf{Incremental Learning}}                                                                                                                                                          \\ \hline
$W_{train}$       & window size for incremental learning.                                                                                                                                                   \\
$mov_{train}$     & training moving window of size $W_{train}$.                                                                                                                                             \\
$\gamma$          & \begin{tabular}[c]{@{}c@{}}parameter to decide size range of $mov_{train}(i)$, \\ where $mov_{train}(i)$ randomly vary in the range \\ $(W_{train} - \gamma, W_{train})$.\end{tabular} 
                                                                                                                                                   \\ \hline
\multicolumn{2}{c}{\textbf{Ensemble Learning}}                                                                                                                                                             \\ \hline
$ES_{warntrig}$   & warning trigger for ensemble learning.                                                                                                                                                  \\
$ES_{alarmtrig}$  & alarm trigger for ensemble learning.                                                                                                                                                    \\
$expiry\_time$    & expiry time of $ES_{warntrig}$.                                                                                                                                                         \\
$mov_{ESwarn}$    & moving window storing instances with warning flags.                                                                                                                                     \\
$P_{thre}$        & anomaly prediction threshold. 
\\
$D_{thre}$        & decision threshold of drift detection.                                                                                                \\ \hline
\multicolumn{2}{c}{\textbf{Prediction}}                                                                                                                                                                    \\ \hline
$b$               & \begin{tabular}[c]{@{}c@{}}percentile value to set anomaly threshold \\ (only for ablation study).\end{tabular}                                                                         \\
$\theta$          & anomaly prediction threshold.                                                                                                                                                           \\ \hline
\multicolumn{2}{c}{\textbf{Data}}                                                                                                                                                                          \\ \hline
$D$               & unlabeled pre-training data.                                                                                                                                                            \\
$x$               & input instance, dimension:${R}^d$.                                                                                                                                                                         \\
$\hat{x}$         & reconstructed instance, dimension:${R}^d$.                                                                                                                                                                   \\
$y$               & label of $x$.                                                                                                                                                                           \\
$\hat{y}$         & predicted label of $x$.                                                                                                                                                                 \\ \hline
\end{tabular}
\end{adjustbox}
\end{table}

The hybrid method adeptly combines the benefits of passive and active approaches to address concept drift efficiently. By leveraging ensemble learning, each component's performance is optimized, resulting in improved overall model efficacy. Drawing inspiration from these concepts, we introduce VAE++ESDD, a novel approach that integrates ensemble incremental learning—a passive method—and ensemble drift detectors—an active method. The pseudocode can be found in the Algorithm~\ref{alg:method1} and the descriptions of variables in this method is summerized in Table~\ref{tab:varia}.

The overview of the proposed method VAE++ESDD is shown in Fig.~\ref{fig:VAE++ESDD}. In this method, we perform ensemble learning of $n$ VAEs. The prediction part is displayed in green color. The system first observes the instance $x^t\in{R}^d$ at time $t$ and puts it in each model for class prediction. Then the final prediction $\hat{y}^t$ is determined by a voting strategy; i.e., the final prediction is anomalous ($\hat{y}^t=1$) only if at least $P_{thre}$ models predict this instance as anomalous, where $P_{thre}$ is the decision threshold of prediction. The most recent instances are stored in a sliding window of size $W_{train}$. Depending on whether or not a condition is satisfied (discussed below), each VAE is incrementally updated independently with a training moving window $mov_{train}(i)$ of size $W_{train}(i) < W_{train}$. This process is displayed in orange color. Notably, except the random initialization of each model, each $mov_{train}(i)$ has a different size as shown in the figure to achieve ensemble diversity. The proposed method is completely unsupervised as the ground truth $y^t$ is not available. Then we incorporate an ensemble of drift detectors, outlined by the grey dashed line in Fig.~\ref{fig:VAE++ESDD}. For each model in the ensemble of drift detectors, the instance $x^t$ is continuously appended into the drift moving window $mov_{driftx}(i)$, which is used to compare the VAE's reconstruction loss with the reference window $ref_{driftx}(i)$ for drift detection. The window size $W_{drift}(i)$ of each $mov_{driftx}(i)$ and $ref_{driftx}(i)$ pair varies to promote diversity. Then we perform the voting strategy. If at least $D_{thre}$ detectors raise alarms, where $D_{thre}$ is the decision threshold of detection, we consider it as a real alarm and all VAEs are discarded and new ones are created, while all windows are emptied.

\subsection{Ensemble of incremental learners}\label{sec:ensem_il}
For each ensemble member $i$, an update is performed only when the condition $t \% W_{\text{train}}(i)=0$ is satisfied, as shown in Lines 15–16 of Algorithm~1.

\textbf{Ensemble Model.}
Each ensemble member employs a variational autoencoder (VAE)~\citep{kingma2014auto}, where the encoder learns a latent distribution $q(z \mid x)$ regularized toward a multivariate standard Gaussian. For a $k$-dimensional latent variable $z \in \mathbb{R}^k$, the encoder outputs the mean $\mu \in \mathbb{R}^k$ and standard deviation $\sigma \in \mathbb{R}^k$, with diagonal covariance $\sigma^2$. Regularization is enforced via the Kullback--Leibler (KL) divergence~\citep{kullback1951information} between $q(z \mid x)$ and the unit Gaussian $N(0, I_k)$:

\begin{equation}\label{eq:kl}
\begin{aligned}
l_{K L}(x) & =K L\left(q(z \mid x) \| N\left(0, I_k\right)\right) \\
& =\frac{1}{2} \sum_{i=1}^k \left(\mu_i^2+\sigma_i^2-\log \left(\sigma_i^2\right)-1\right)
\end{aligned}
\end{equation}

The total loss is determined by the combination of the reconstruction loss and the regularization loss, represented by Eq.~(\ref{eq:vae}), in which $\beta$ is a weight parameter that adjusts the importance of the KL divergence loss in the overall loss. $\hat{x}$ represents the reconstructed value of $x$ through the VAE. The $l_{A E}(x, \hat{x})$ depends on the input's nature. For real-valued inputs, it is defined as the sum of squared differences, while for binary inputs it is defined as the cross-entropy.

\begin{equation}\label{eq:vae}
l_{V A E}(x, \hat{x})=l_{A E}(x, \hat{x})+ \beta * l_{K L}(x)
\end{equation}

\textbf{Memory}. The most recent instances are stored using a sliding window of size \( W_{train} \). During each training iteration for each ensemble member, a random subset of instances of size \( W_{train}(i) < W_{train}\) from this window is selected as a training window \( mov_{train}(i) \) for incremental training, where $i\in{1, ..., M}$ is a member of ensemble of size $M$. The reason for employing a sliding window is two-fold. First, it is unreasonable to assume that all arriving instances could be stored and be available at all times. Second, the sliding window implicitly addresses the problem of concept drift, as obsolete examples will eventually drop out of the window. It is expected that the window will be mostly populated by normal instances, due to the class imbalance encountered in an anomaly detection problem. 

\textbf{Ensemble diversity.}
To promote ensemble diversity, the sizes of \( mov_{train}(i) \) randomly vary in the range $(W_{train} - \gamma, W_{train})$. For instance, if $W_{train} = 3000$ and $\gamma = 2000$, \( W_{train}(i) \) falls within the range of \( (1000, 3000) \). At any time \( t \), each model maintains a window \( mov_{train}(i)^t = \{x^t\}^t_{t-W_{train(i)}+1} \), where $x_t$ is the instance observed at time $t$. Due to different window sizes, models update at different paces. Shorter windows adapt quickly but fluctuate more, while longer ones are steadier but slower to react. This temporal diversity improves robustness to both abrupt and gradual changes—even in stationary settings.

\textbf{Anomaly detection (prediction).}
Autoencoder-based detection exploits the fact that anomalous instances typically produce higher reconstruction losses than normal ones. For model $i$, an instance at time $t$ is labeled as anomalous if its total reconstruction loss (Eq.~\ref{eq:vae}) exceeds the adaptive threshold $\theta^{t}(i)$, which is updated after each model refinement. Threshold selection is non-trivial due to the ambiguous boundary between normal behavior and outliers~\citep{clark2018adaptive}.

To address this, we employ an adaptive thresholding strategy that tracks the reconstruction error distribution. Once $t$ reaches the training period, the loss set
$L^{t}(i)=\{l_{\mathrm{VAE}}(x^k,\hat{x}^k)\}_{k=t-W_{\mathrm{train}}(i)+1}^{t}$
is computed over the sliding training window $mov^{t}_{\mathrm{train}}(i)$, and the anomaly threshold is defined as the mean plus one standard deviation of $L^{t}(i)$ (Eq.~\ref{eq:adtthreshold}).

\begin{equation}\label{eq:adtthreshold}
\theta^t(i) = mean(L^t(i))+std(L^t(i))
\end{equation}

The adaptive threshold in Eq.~\ref{eq:adtthreshold} is designed as a data-driven heuristic decision rule. In unsupervised and non-stationary settings, the reconstruction error distribution and anomaly rate are unknown and time-varying, making formal statistical thresholding impractical. The mean-plus-standard-deviation rule offers an approximation of the upper tail of the error distribution by jointly capturing the central tendency and dispersion of reconstruction errors within a local window. This formulation allows the threshold to adapt naturally to shifts in the error distribution while remaining insensitive to small fluctuations. Similar heuristic thresholding strategies have been widely adopted in reconstruction-based anomaly detection methods based on autoencoders and VAEs \citep{sakurada2014anomaly,an2015variational}.

Once having set a threshold, we can proceed with anomaly prediction. Suppose a new instance $x^{t+\Delta}$, $\Delta > 0$ is now observed. 
\begin{equation}\label{eq:predict}
\hat{y}^{t+\Delta}(i) =
\begin{cases}
1 \ \text{(anomaly)} & \text{if} \ l_{V A E}(x^{t+\Delta}, \hat{x}^{t+\Delta}) > \theta^t(i)\\
0 \ \text{(normal)} & \text{otherwise}
\end{cases}
\end{equation}
\noindent where $t$ is the time of the most recent training, and $t+\Delta$ is the current time.

If the number of classifiers that have detected anomalies is equal to or greater than $P_{thre}$, we infer the presence of anomalies:   $\hat{y}^t = 1$ if  $\sum_{i=1}^{n}(\hat{y}^t(i))>= P_{thre}$.

In the empirical analysis, we examine the benefit of an adaptive versus fixed threhsolding approach.

\textbf{Incremental learning (training)}. Each model is continuously updated through incremental learning, which refers to the gradual adaptation of a model without complete re-training. Specifically, when the current time step $t+\Delta$ reaches the scheduled re-training point, the model parameters are updated as shown in Eq.~\ref{eq:update}. 

\begin{equation}\label{eq:update}
h^{t+\Delta}(i) = h^{t}(i).train(mov^t_{train}(i))
\end{equation}


The incremental learning step in Eq.~\ref{eq:update} corresponds to a bounded fine-tuning process rather than full retraining. At each update time $t+\Delta$, the model is initialized from $h^{t}(i)$ and optimized on a fixed sliding training window $mov^t_{\mathrm{train}}(i)$ for a limited number of epochs, resulting in a finite stochastic optimization problem. While the sliding window evolves over time, the objective function is fixed within each update step. Therefore, convergence properties can be discussed at the level of individual updates, but not globally across updates. Under mild smoothness assumptions, gradient-based methods are known to exhibit stable descent behavior and converge to a stationary point of the local objective for such bounded optimization problems~\citep{bottou2018optimization,ghadimi2013stochastic}.


\subsection{Ensemble of drift detectors}\label{sec:ensem_dd}

\textbf{Memory.} Each ensemble member maintains its own memory, consisting of a drift reference window $ref^t_{driftx}(i)$ and a drift moving window $mov^t_{driftx}(i)$ at time $t$, both of size $W_{drift}(i)$.

The instances of $ref^t_{driftx}(i)$ are obtained from the first window of arriving data and $mov^t_{driftx}(i)$ is used to store the most recent instances, i.e., for ensemble member $i$ at time $t$,  $mov^t_{driftx}(i)=\{x^t\}^t_{t-W_{drift}(i)+1}$ and $ref^t_{driftx}(i)=\{x^t\}^{W_{drift}(i)}_1$. These two windows are used for statistical comparison to detect concept drift. Another window $mov_{ESwarn}$ of size $W_{drift}(i)$ stores instances with warning flags, which are used to train the new VAE and get the new threshold once the flag alarm is triggered.

\begin{equation}\label{eq:ref_mov_def}
\begin{aligned}
\mathrm{ref}^{t}_{\mathrm{driftl}}(i)
&= \big\{\, l_{\mathrm{VAE}}(x^{t}, \hat{x}^{t})
\;\big|\; x^{t} \in \mathrm{ref}^{t}_{\mathrm{driftx}}(i) \,\big\} \\[4pt]
\mathrm{mov}^{t}_{\mathrm{driftl}}(i)
&= \big\{\, l_{\mathrm{VAE}}(x^{t}, \hat{x}^{t})
\;\big|\; x^{t} \in \mathrm{mov}^{t}_{\mathrm{driftx}}(i) \,\big\}
\end{aligned}
\end{equation}

As defined in Eq.~\ref{eq:ref_mov_def}, 
$\mathrm{ref}^{t}_{\mathrm{driftl}}(i)$ and $\mathrm{mov}^{t}_{\mathrm{driftl}}(i)$ 
represent the sets of total losses corresponding to the instances within 
the reference and moving windows, 
$\mathrm{ref}^{t}_{\mathrm{driftx}}(i)$ and $\mathrm{mov}^{t}_{\mathrm{driftx}}(i)$, respectively. 
The drift detection mechanism operates by comparing these two loss sets.

\textbf{Statistical test.} As a non‐parametric test, the Mann‐Whitney U Test does not depend on assumptions on the distribution \citep{nachar2008mann}. Benefiting from this property, researchers have applied this method to fault detection \citep{yang2019new}, \citep{swain2019complete}. The test is performed by two hypotheses (H0 and H1).

H0: There is no difference between $ref^t_{driftl}(i)$ and $mov^t_{driftl}(i)$ (the two windows of samples come from the same distribution).

H1: There is a difference between $ref^t_{driftl}(i)$ and $mov^t_{driftl}(i)$ (the two windows of samples do not come from the same distribution).

The Mann-Whitney U test proceeds as follows. Samples from the two windows are merged, sorted, and assigned ranks. In case of ties, each tied value is assigned the average of the ranks they would occupy. The sum of the ranks refers to the total of the ranks assigned to the values originating from a specific window. And the sum of ranks of $ref_{driftl}(i)$ and $mov_{driftl}(i)$ are represented as $Rank(ref^t_{driftl}(i))$ and $Rank(mov^t_{driftl}(i))$ respectively. Based on these rank sums, we compute the U values for the reference and moving windows, respectively, as shown in Eq.~\ref{eq:u}.

\begin{equation}\label{eq:u}
\begin{aligned}
U^{t}_{\mathrm{ref}}(i) &= 
n_{\mathrm{ref}} n_{\mathrm{mov}}
+ \frac{n_{\mathrm{ref}}\left(n_{\mathrm{ref}} + 1\right)}{2}
- \operatorname{Rank}\!\left(\mathrm{ref}^{t}_{\mathrm{driftl}}(i)\right), \\
U^{t}_{\mathrm{mov}}(i) &= 
n_{\mathrm{ref}} n_{\mathrm{mov}}
+ \frac{n_{\mathrm{mov}}\left(n_{\mathrm{mov}} + 1\right)}{2}
- \operatorname{Rank}\!\left(\mathrm{mov}^{t}_{\mathrm{driftl}}(i)\right),\\
\text{where } &n_{\mathrm{ref}} = n_{\mathrm{mov}} = W_{\mathrm{drift}}(i)
\end{aligned}
\end{equation}

\begin{equation}\label{eq:z}
\begin{aligned}
Z^t(i)&=\frac{U^t(i)-\operatorname{mean}(U^t(i))}{\operatorname{std}(U^t(i))},\\
\text{where } &U^t(i)=\operatorname{min}(U^{t}_{\mathrm{ref}}(i) ,U^{t}_{\mathrm{mov}}(i) )
\end{aligned}
\end{equation}

The Z-value $Z^t(i)$ is computed using Eq.~\ref{eq:z}, from which the P-value $P^t(i)$ is obtained via Eq.~\ref{eq:pvalue_def}. Two thresholds, $P_{warn}$ and $P_{alarm}$, define warning and alarm levels, respectively, ensuring that $flag_{warn}(i)$ precedes $flag_{alarm}(i)$. Upon $flag_{alarm}(i)$ is raised according to Eq.~\ref{eq:pvalue}, H0 (null hypothesis) is rejected in favor of H1 (alternative hypothesis). $flag_{warn}(i)$ indicates a potential concept drift, while $flag_{alarm}(i)$ triggers an actual concept drift alarm. Each ensemble member employs these flags for drift detection.

\begin{equation}\label{eq:pvalue_def}
P^t(i)=2 \int_{-\infty}^t \frac{1}{\sqrt{2 \pi \operatorname{var}(Z^t(i))}} \times e^{\left(-\frac{Z^t(i)-\operatorname{mean}(Z^t(i))}{2 \operatorname{var}(Z^t(i))}\right)}dt
\end{equation}


\begin{equation}\label{eq:pvalue}
{flag(i)} =
\begin{cases}
warn \  & if \ P^t(i) < P_{warn}\\
alarm \ &  if \ P^t(i) < P_{alarm}\\
\end{cases}
\end{equation}

Algorithm 1 outlines the high-level procedure, where individual model flags can activate ensemble learning's warning and alarm triggers, $ES_{warntrig}$ and $ES_{alarmtrig}$ (Line 11 to 14). Further elaboration will follow.

\textbf{Warning flag raised}. The first time $flag_{warn}(i)$ is raised in any dimension $i$, the warning trigger for ensemble learning, $ES_{warntrig}$, is triggered. Then, the process begins to store examples into the moving window called $mov_{ESwarn}$ for re-training the models as detailed in Lines 17 and 18. To avoid false alarms, we set a parameter $expiry\_time$. As shown in Lines 19 and 20, if $ES_{warntrig}$ is raised for more than $expiry\_time$ and $ES_{alarmtrig}$ is still false, we regard this as false warnings and then reset the status of $ES_{warntrig}$ and empty $mov_{ESwarn}$.

\textbf{Alarm flag raised}. When an alarm flag is triggered, a final drift decision is made via majority voting across the ensemble. If the number of models detecting drift reaches or exceeds $D_{\mathrm{thre}}$, drift is declared. In this case, each ensemble member replaces its current autoencoder with a newly initialized one, which is trained using $mov_{\mathrm{ESwarn}}$, and the detection threshold is updated accordingly. All state windows ($ref_{\mathrm{drift}x}(i)$, $mov_{\mathrm{train}}(i)$, $mov_{\mathrm{drift}x}(i)$, and $mov_{\mathrm{ESwarn}}$) are then cleared, the alarm and warning flags are reset, and a new post-drift reference window of size $W_{\mathrm{drift}}(i)$ is constructed from incoming data. The integration of drift detection (DD) directly enhances outlier detection (OD) by explicitly signaling when distributional shifts occur and when model adaptation should be performed. Through the warn–alarm mechanism, DD retains post-drift samples for adaptive retraining, enabling rapid realignment with the new data distribution and preventing immediate performance degradation.

\subsection{Computational aspects}\label{sec:com_ana}

\subsubsection{Training stage}

Incremental learning maintains a shared sliding window \(mov_{\text{train}}\) of size \(W_{\text{train}}\), resulting in a memory cost of \(W_{\text{train}}d\). Each ensemble member \(i\) technically uses its own window size \(W_{\text{train}}(i)\); for simplicity, we use \(W_{\text{train}}\) to denote their average and compute memory accordingly. For drift detection, each ensemble member keeps two windows, \(ref_{\text{driftx}}(i)\) and \(mov_{\text{driftx}}(i)\), each with size \(W_{\text{drift}}(i)\). Similar to the training window, we use \(W_{\text{drift}}\) to denote the average drift-detection window size across members. Hence, the memory required for drift detection is approximately \(2nW_{\text{drift}}d\). Upon drift detection, retraining uses the warning buffer \(mov_{\text{ESwarn}}\), also of size \(W_{\text{drift}}\), adding an additional \(W_{\text{drift}}d\) memory independent of \(n\).  
Therefore, the total memory cost in the training stage is $(2n + 1)W_{\text{drift}}d + W_{\text{train}}d$.

During incremental learning, each ensemble member \(i\) updates every \(W_{\text{train}}(i)\) steps using its recent window \(mov_{\text{train}}(i)\) for \(E\) epochs, with a cost
 $T_{\text{incr}}^{(i)} \approx E\,W_{\text{train}}(i)\,\alpha\,C_{\text{fwd}}$, where \(\alpha\) accounts for backpropagation and optimizer updates. These updates occur periodically, scale roughly linearly with \(n\), and can be executed in parallel. When drift is detected, each member is retrained using \(W_{\text{drift}}\) samples from the warning buffer for \(E\) epochs, resulting in $T_{\text{drift}} \approx n\,E\,W_{\text{drift}}\,\alpha\,C_{\text{fwd}}$. Although more expensive, this cost is incurred only after drift alarms, and parallel training can reduce the effective wall-clock time to \(n/s_{\text{train}}(n)\), where \(s_{\text{train}}(n)\) denotes the parallel speedup factor.

\subsubsection{Prediction stage}

At each time step, each VAE performs only forward propagation. With \(P_{\text{vae}}\) parameters per VAE, the total model-related memory is \(nP_{\text{vae}}\).

Streaming operations consist of  
(i) anomaly scoring via forward passes of the \(n\) VAEs, and  
(ii) drift detection via \(n\) statistical tests.  
The per-step computational complexity is \(O\!\big(n(C_{\text{fwd}} + C_{\text{test}})\big)\), where \(C_{\text{fwd}}\) is the cost of a single forward pass and \(C_{\text{test}}\) the cost of one drift test. The wall-clock time per step is $T_{\text{stream}} = T_{\text{fixed}} + nT_{\text{fwd}} + nT_{\text{test}}$, where \(T_{\text{fixed}}\) includes ensemble-independent preprocessing. In practice, \(T_{\text{fwd}}\) and \(T_{\text{test}}\) can be parallelized, allowing the runtime to grow slower than linearly with the ensemble size $n$.

\begin{table*}[!h]
\caption{Description of Synthetic and Real-World Datasets}
\label{tab:dataset}
\begin{adjustbox}{width=0.95\textwidth}
\begin{tabular}{|c|c|c|c|c|c|c|c|}

\hline
Type                       & Dataset                    & \# Features          & \# Instances           & Drift Time            & Drift Type              & Before Drift                                                                                                                                 & After Drift                                                                                                                                  \\ \hline
\multirow{4}{*}{Synthetic} & Sea (2 variations)         & 2                    & \multirow{4}{*}{20000} & 10000, 15000          & Recurrent               & \begin{tabular}[c]{@{}c@{}}C0: $x_1 + x_2 > 7$, C1: $x_1 + x_2 \leq 7$\\ $x_1, x_2 \in [0, 10]$\end{tabular} & \begin{tabular}[c]{@{}c@{}}C0: $x_1 + x_2 \leq 7$, C1: $x_1 + x_2 > 7$\\ $x_1, x_2 \in [0, 10]$\end{tabular}                               \\ \cline{2-3} \cline{5-8} 
                           & Circle (2 variations)      & 2                    &                        & 10000                 & Abrupt                  & \begin{tabular}[c]{@{}c@{}}center=(0.4,0.5), radius=0.2\\ C0: inside, C1: outside\\ $x_1, x_2 \in [0, 1]$\end{tabular} & \begin{tabular}[c]{@{}c@{}}center=(0.4,0.5), radius=0.2\\ C0: outside, C1: inside\\ $x_1, x_2 \in [0, 1]$\end{tabular}                     \\ \cline{2-3} \cline{5-8} 
                           & Sine (2 variations)        & 2                    &                        & 10000                 & Abrupt                  & \begin{tabular}[c]{@{}c@{}}C0: $x_2 > \sin(x_1)$\\ C1: $x_2 < \sin(x_1)$\\ $x_1 \in [0, \pi]$, $x_2 \in [-1, 1]$\end{tabular} & \begin{tabular}[c]{@{}c@{}}C0: $x_2 < \sin(x_1)$\\ C1: $x_2 > \sin(x_1)$\\ $x_1 \in [0, \pi]$, $x_2 \in [-1, 1]$\end{tabular}               \\ \cline{2-3} \cline{5-8} 
                           & Vib (2 variations)         & 10                   &                        & 10000-11000           & Incremental                 & \begin{tabular}[c]{@{}c@{}}C0: mean=0, std=1\\ C1: mean=5, std=1\end{tabular}                           & \begin{tabular}[c]{@{}c@{}}C0: mean=3, std=1\\ C1: mean=5, std=1\end{tabular}                           \\ \hline
\multirow{6}{*}{Real}      & MNIST-01 (2 variations)    & \multirow{3}{*}{784} & \multirow{3}{*}{5000}  & \multirow{3}{*}{2500} & \multirow{3}{*}{Abrupt} & \begin{tabular}[c]{@{}c@{}}C0: digit 0\\ C1: digit 1\end{tabular}                                       & \begin{tabular}[c]{@{}c@{}}C0: digit 0\\ C1: digit 1\\ width \& height shift $\pm10\%$\end{tabular}                    \\ \cline{2-2} \cline{7-8} 
                           & MNIST-multi (2 variations) &                      &                        &                       &                         & \begin{tabular}[c]{@{}c@{}}C0: digit 0\\ C1: digits 1-9\end{tabular}                                   & \begin{tabular}[c]{@{}c@{}}C0: digit 0\\ C1: digits 1-9\\ width \& height $\pm10\%$\end{tabular}                 \\ \cline{2-2} \cline{7-8} 
                           & MNIST-23 (2 variations)    &                      &                        &                       &                         & \begin{tabular}[c]{@{}c@{}}C0: digit 0\\ C1: digit 1\end{tabular}                                       & \begin{tabular}[c]{@{}c@{}}C0: digit 2\\ C1: digit 3\end{tabular}                                      \\ \cline{2-8} 
                           & Fraud (2 variations)       & 29                   & 15000                  & 5000, 10000           & Recurrent               & \citep{dal2015calibrating}                                                                                                     & values of all features $\times 0.1$                                                                  \\ \cline{2-8} 
                           & Forest (2 variations)      & 52                   & 100000                 & 50000                 & Abrupt                  & \begin{tabular}[c]{@{}c@{}}C0: covertype 1-2\\ C1: covertype 3-7\end{tabular}                         & \begin{tabular}[c]{@{}c@{}}C0: covertype 1-4\\ C1: covertype 5-7\end{tabular}                         \\ \cline{2-8} 
                           & Arrhy                      & 187                  & 80000                  & Unknown               & Unknown                 & \multicolumn{2}{c|}{\citep{moody2001impact}}                                                                                                                                                       \\ \hline
\end{tabular}
\end{adjustbox}
\end{table*}

\section{Experimental Setup}\label{sec:exp_setup}

\subsection{Datasets}

Our experimental study considers (i) synthetic (Sea, Circle, Sine, Vib) and real-world (MNIST-01, MNIST-multi, MNIST-23, Forest, Fraud, Arrhy) datasets, and (ii) cases with imbalance rates of 1\% (severe) and 0.1\% (extreme). Recurrent drift and incremental drift are also considered in different datasets. The description of all datasets can be found in Table \ref{tab:dataset}, where `C0' and `C1' represent normal and anomalous classes respectively.

\noindent\textbf{Synthetic datasets:}

\textbf{Sea} \citep{street2001streaming} has two features and its class boundary is set by inequations, as shown in Table~\ref{tab:dataset}. 

\textbf{Circle} \citep{gama2004learning} has two features and its class boundary is in the shape of a circle.

\textbf{Sine} \citep{gama2004learning} consists of two features and the sine function is the classification boundary.

\textbf{Vib} \citep{li2024unsupervised} consists of simulated equipment vibration data in industrial manufacturing.

\noindent\textbf{Real-world datasets:}

\textbf{MNIST}~\citep{lecun1998gradient} is a dataset of handwritten digit images of size $28\times28$ (784 features). Although commonly used for image classification, its high dimensionality poses challenges for streaming methods. We consider three variants—\textbf{MNIST-01}, \textbf{MNIST-23}, and \textbf{MNIST-multi}—which include different digit subsets, as detailed in Table~\ref{tab:dataset}.

\textbf{Fraud} \citep{dal2015calibrating} contains credit card transactions from September 2013 by European cardholders. This dataset corresponds to transactions that occurred over two days. 

\textbf{Forest} \citep{blackard1999comparative} contains cartographic information from the U.S. Forest Service. The task is to predict the forest cover type for given 30x30m cells from the Roosevelt National Forest in Colorado.

\textbf{Arrhy}~\citep{moody2001impact} (MIT-BIH Arrhythmia Database) is a widely used ECG dataset consisting of 30-minute recordings from 47 patients, annotated with five arrhythmia types and locations. The dataset is highly imbalanced (around 18\%), and its concept drift characteristics are unknown.

As summarized in Table~\ref{tab:dataset}, Sea, Circle, and Sine induce decision-boundary shifts with unchanged input distributions, corresponding to changes in $P(y\!\mid\!x)$. Vib, MNIST-01, MNIST-multi, and Fraud involve input distribution shifts with fixed label mappings, reflecting changes in $P(x)$. MNIST-23 and Forest exhibit joint changes in $P(x)$ and $P(y\!\mid\!x)$, while Arrhy contains unknown drift sources typical of real-world physiological data. Although the proposed method is based on VAE reconstruction loss, drift is detected by comparing loss distributions between reference and current windows rather than absolute errors; thus, any significant variation in the data distribution—arising from shifts in $P(x)$, $P(y\!\mid\!x)$, or both—can be effectively captured, including cases where only $P(y\!\mid\!x)$ changes.

The datasets in this study span several common and realistic drift types. Recurrent drift (Sea, Fraud) models periodically alternating behaviors, abrupt drift (Circle, Sine, Forest, MNIST-01, MNIST-multi) reflects sudden regime changes, incremental drift (Vib) captures gradual shifts due to system degradation, and class-substitution drift (MNIST-23) represents complete task changes. Finally, Arrhythmia exhibits irregular or unknown drift typical of real-world monitoring data.

\textbf{Baseline}: This method is pretrained offline using 2000 unlabeled normal samples, with the online threshold determined by an adaptive thresholding strategy. The resulting pretrained model is also used as the initialization for the subsequent methods.

\textbf{iForest++} \citep{liu2008isolation}: A tree-based method discussed in Sec.~\ref{sec:anoamlydetection}. To ensure fairness, we follow the same incremental learning framework as the proposed method.

\textbf{LOF++} \citep{breunig2000lof}: Another popular method to compare is LOF, the detail is provided in Sec.~\ref{sec:anoamlydetection}. As with iForest++, we follow the same incremental learning framework as the proposed method.


\textbf{StrAEm++DD} \citep{li2023autoencoder}, our previous work, serves as the foundation of this study.

\textbf{SEAD} \citep{shah2025sead}, a recent ensemble framework for anomaly detection as described in Sec.~\ref{sec:anoamlydetection}. 

\textbf{CPOCEDS} \citep{jafseer2024cpoceds}, a recent cluster-based anomaly detection method as described in Sec.~\ref{sec:anoamlydetection}.

\textbf{ARCUS} \citep{yoon2022adaptive}: an recent online deep anomaly detection framework as described in Sec.~\ref{sec:anoamlydetection}. 

\textbf{MemStream} \citep{bhatia2022memstream}: a recent DAE-based method which can adapt to evolving data stream. The detailed description can be found in Sec.~\ref{sec:anoamlydetection}.

\textbf{METER} \citep{zhu2023meter}: a recent dynamic concept adaptation framework for online anomaly detection. The detailed description can be found in Sec.~\ref{sec:anoamlydetection}.

\begin{table}[!h]
\caption{Hyper-parameter values for (V)AE++(ES)(DD)}\label{tab:params_nn}
\centering
\begin{adjustbox}{width=0.5\textwidth}
\begin{tabular}{|c|cccccccc|}
\hline
                  & \multicolumn{3}{c|}{Synthetic}                                                        & \multicolumn{5}{c|}{Real-world}                                                                                                                \\ \hline
Datasets          & \multicolumn{1}{c|}{Sea}    & \multicolumn{1}{c|}{Sine} & \multicolumn{1}{c|}{Circle} & \multicolumn{1}{c|}{Vib} & \multicolumn{1}{c|}{Fraud}        & \multicolumn{1}{c|}{MNIST}                & \multicolumn{1}{c|}{Forest} & Arrhy \\ \hline
Learning rate     & \multicolumn{8}{c|}{0.001}                                                                                                                                                                            \\ \hline
Hidden layers     & \multicolumn{1}{c|}{[64,8]} & \multicolumn{3}{c|}{8}                                                             & \multicolumn{1}{c|}{[64,32]}      & \multicolumn{1}{c|}{[512,256]}            & \multicolumn{1}{c|}{64}     & 128   \\ \hline
Mini-batch size   & \multicolumn{8}{c|}{64}                                                                                                                                                                                                                \\ \hline
Weigh initializer & \multicolumn{8}{c|}{He Normal}                                                                                                                                                                                                         \\ \hline
Optimizer         & \multicolumn{8}{c|}{Adam}                                                                                                                                                                                                              \\ \hline
Hidden activation & \multicolumn{8}{c|}{Leaky ReLU}                                                                                                                                                                                                        \\ \hline
Num. of epochs    & \multicolumn{1}{c|}{10}     & \multicolumn{1}{c|}{10}   & \multicolumn{1}{c|}{50}     & \multicolumn{1}{c|}{1}   & \multicolumn{1}{c|}{10}           & \multicolumn{1}{c|}{10}                   & \multicolumn{1}{c|}{50}     & 10    \\ \hline
Output activation & \multicolumn{8}{c|}{Sigmoid}                                                                                                                                                                                                           \\ \hline
Loss function     & \multicolumn{4}{c|}{Binary cross-entropy}                                                                        & \multicolumn{1}{c|}{Square error} & \multicolumn{1}{c|}{Binary cross-entropy} & \multicolumn{2}{c|}{Square error}   \\ \hline
Beta (VAE)        & \multicolumn{8}{c|}{1.0}                                                                                                                                                                                                               \\ \hline

\end{tabular}
\end{adjustbox}
\end{table}


\textbf{VAE++ES (proposed)}: This variant represents the ensemble-based incremental learning module described in Sec.~\ref{sec:ensem_il}.

\textbf{VAE++ESDD (proposed)}: This is the complete proposed method, combining the VAE++ES incremental learner (Sec.\ref{sec:ensem_il}) with an ensemble of drift detectors (Sec.\ref{sec:ensem_dd}).

\begin{table}[H]
\caption{Summary of ablation studies}
\label{tab:comp_resource}
\begin{adjustbox}{width=0.5\textwidth}
\centering
\begin{tabular}{|p{4cm}|p{11cm}|}

\hline
Ablation Study & Description \\ \hline

Sec. 6.1.1 Different Autoencoder Types 
& To compare different autoencoder architectures and justify the selection of the VAE.\\ \hline

Sec. 6.1.2 Selection of $\beta$ 
& To analyze the influence of the KL regularization weight on detection performance. \\ \hline

Sec. 6.1.3 Role of the Training Window Size 
& To investigate how the training window size affects model learning, stability, and detection accuracy. \\ \hline

Sec. 6.1.4 Role of the P-value 
& To examine the effect of the significance thresholds $P_{warn}$ and $P_{alarm}$ on detection sensitivity and false alarm control. \\ \hline

Sec. 6.1.5 Fixed Percentile vs. Adaptive Threshold 
& To evaluate the benefits of the proposed adaptive thresholding strategy in comparison with a fixed percentile-based approach. \\ \hline

Sec. 6.1.6 Role of the Prediction Threshold 
& To investigate the impact of the prediction threshold $P_{thre}$ on anomaly detection. \\ \hline

Sec. 6.1.7 Role of the Drift Detection Window Size 
& To investigate how the drift detection window size $W_{drift}$ affects detection performance and warning frequency. \\ \hline

Sec. 6.1.8 Impact of Ensemble Learning 
& To evaluate the impact of ensemble learning on performance and computational cost, considering both predictor and drift-detector ensembles. \\ \hline

Sec. 6.1.9 Model Reset Performance 
& To analyze the model’s recovery behavior and performance after a reset under non-stationary conditions. \\ \hline

\end{tabular}
\end{adjustbox}
\end{table}

\subsection{Evaluation methodology}




To evaluate the performance under class-imbalanced scenarios, we adopt five metrics: \textbf{recall}, \textbf{specificity}, \textbf{geometric mean (G-mean)}, \textbf{Prequential AUC (PAUC)}, and \textbf{Receiver operating characteristic (ROC)}.

\begin{itemize}
    \item \textbf{Recall (Sensitivity)} measures the ability to correctly identify positive instances and is defined as:
    \begin{equation}
        \text{Recall} = R^+ = \frac{TP}{P},
    \end{equation}
    where $TP$ is the number of true positives and $P$ is the total number of actual positive instances.

    \item \textbf{Specificity} reflects the ability to correctly identify negative instances and is given by:
    \begin{equation}
        \text{Specificity} = R^- = \frac{TN}{N},
    \end{equation}
    where $TN$ is the number of true negatives and $N$ is the total number of actual negative instances.

    \item \textbf{G-mean} combines recall and specificity into a single metric to provide a balanced evaluation, especially in imbalanced settings. It is defined as:
    \begin{equation}\label{eq:gmean}
        G\text{-}mean = \sqrt{R^+ \times R^-}.
    \end{equation}
    G-mean is considered high when both recall and specificity are high and similar in value. It is robust to class imbalance and has been widely adopted in imbalanced learning research~\citep{sun2006boosting, he2008learning}.

\item 
\textbf{Prequential ROC (PROC)} extends the conventional receiver operating characteristic (ROC) curve~\citep{metz1978basic} to data stream settings by computing ROC curves over a sliding window of recent observations in a prequential manner~\citep{brzezinski2017prequential}, thereby capturing the classifier’s sensitivity-specificity trade-off under the current data distribution. Formally, at time $t$, let $W_t = \{(x_i, y_i, s_i)\}_{i=t-d+1}^{t}$ denote the most recent $d$ instances. The PROC curve is obtained by varying the decision threshold over the anomaly scores $\{s_i\}$ within $W_t$ and computing the corresponding true positive rate (TPR) and false positive rate (FPR):
\begin{equation}
\mathrm{TPR} = \frac{\mathrm{TP}}{\mathrm{TP} + \mathrm{FN}}, 
\qquad
\mathrm{FPR} = \frac{\mathrm{FP}}{\mathrm{FP} + \mathrm{TN}},
\end{equation}

where TP, FP, TN, and FN denote the numbers of true positives, false positives, true negatives, and false negatives, respectively.

\item 
\textbf{Prequential AUC (PAUC)~\citep{brzezinski2017prequential}} provides a compact, threshold-independent summary of the prequential ROC curve by measuring the area under the PROC within the same sliding window $W_t$. 
It estimates the probability that a randomly selected positive instance receives a higher anomaly score than a randomly selected negative instance under the current concept:
\begin{equation}
\mathrm{PAUC}_t = \frac{1}{|P_t| \cdot |N_t|}
\sum_{x^+ \in P_t} \sum_{x^- \in N_t}
\mathbf{1}\big[s(x^+) > s(x^-)\big],
\end{equation}
where $P_t$ and $N_t$ denote the sets of positive and negative samples in $W_t$, respectively.
In this study, the window size $d$ is set to 1000 to balance drift sensitivity and statistical stability.

\end{itemize}

The \textit{prequential evaluation with fading factors} method is widely used to assess sequential learning algorithms and has been shown to converge to the Bayes error in stationary data~\citep{gama2013evaluating}. In this study, it is applied to compute Recall, Specificity, and G-mean, with a fading factor set to $0.99$. For PAUC and PROC, however, a \textit{sliding window} mechanism is employed instead, enabling drift-aware and temporally adaptive performance evaluation.

\begin{figure}[!h]
 \begin{subfigure}{.5\columnwidth}
  \centering
\includegraphics[width=0.95\columnwidth]{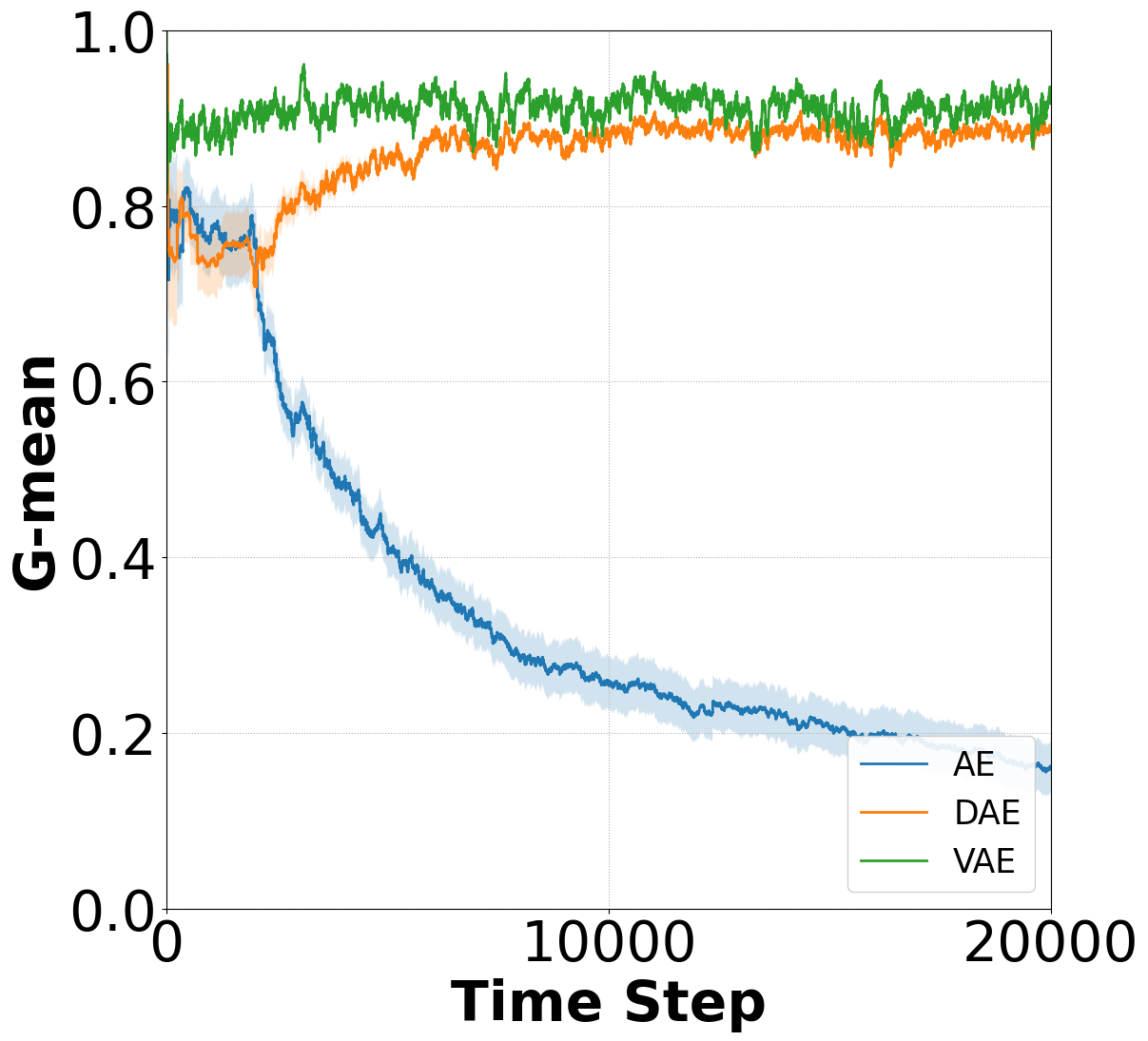} 
  \caption{Sea severe}
  \label{fig:sea_severe_diff_ae}
 \end{subfigure}%
   \begin{subfigure}{.5\columnwidth}
  \centering
\includegraphics[width=0.95\columnwidth]{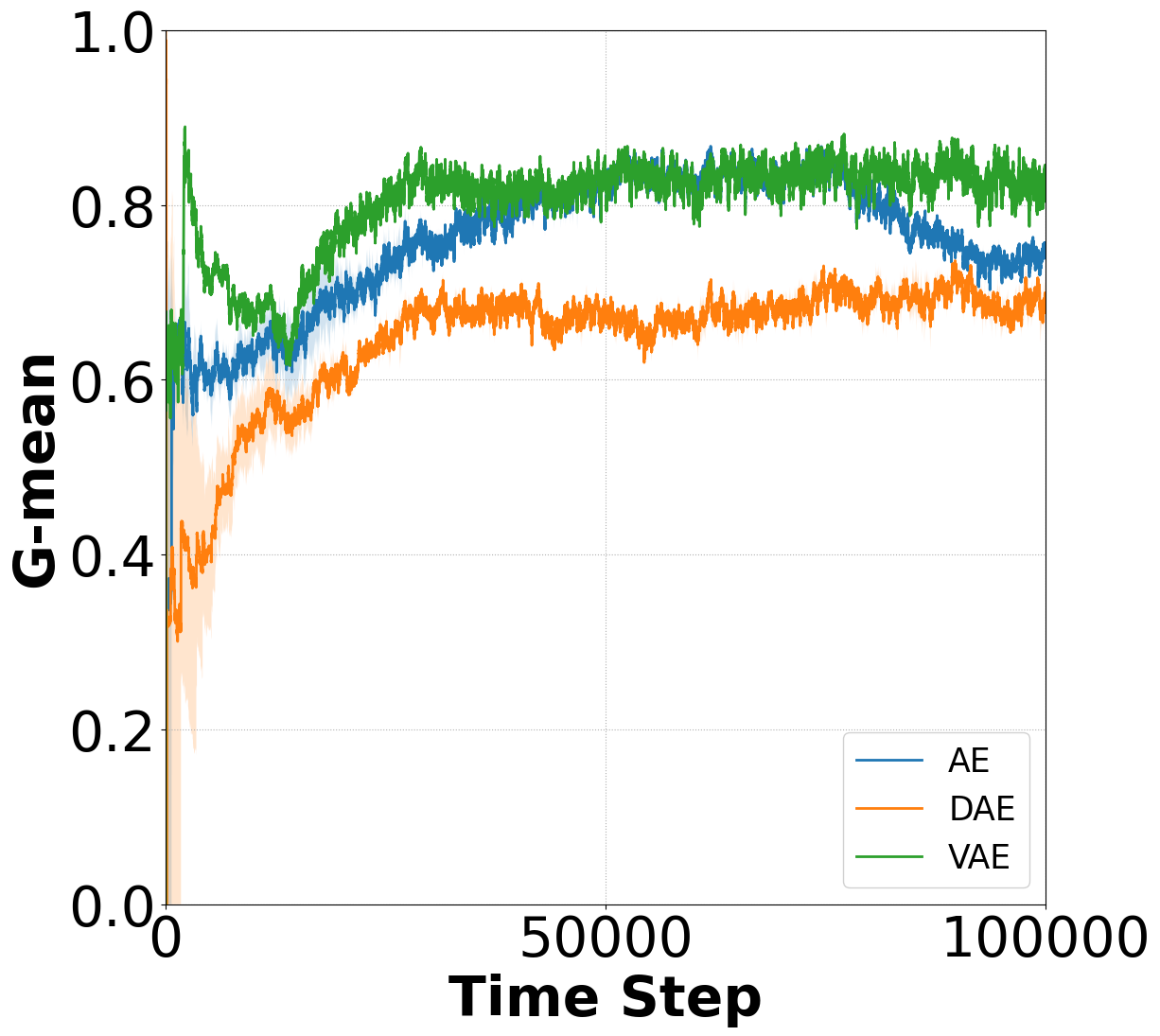} 
  \caption{Forest severe}
\label{fig:forest_severe_diff_ae}
 \end{subfigure}%
  \centering
  
  \begin{subfigure}{.5\columnwidth}
  \centering
\includegraphics[width=0.95\columnwidth]{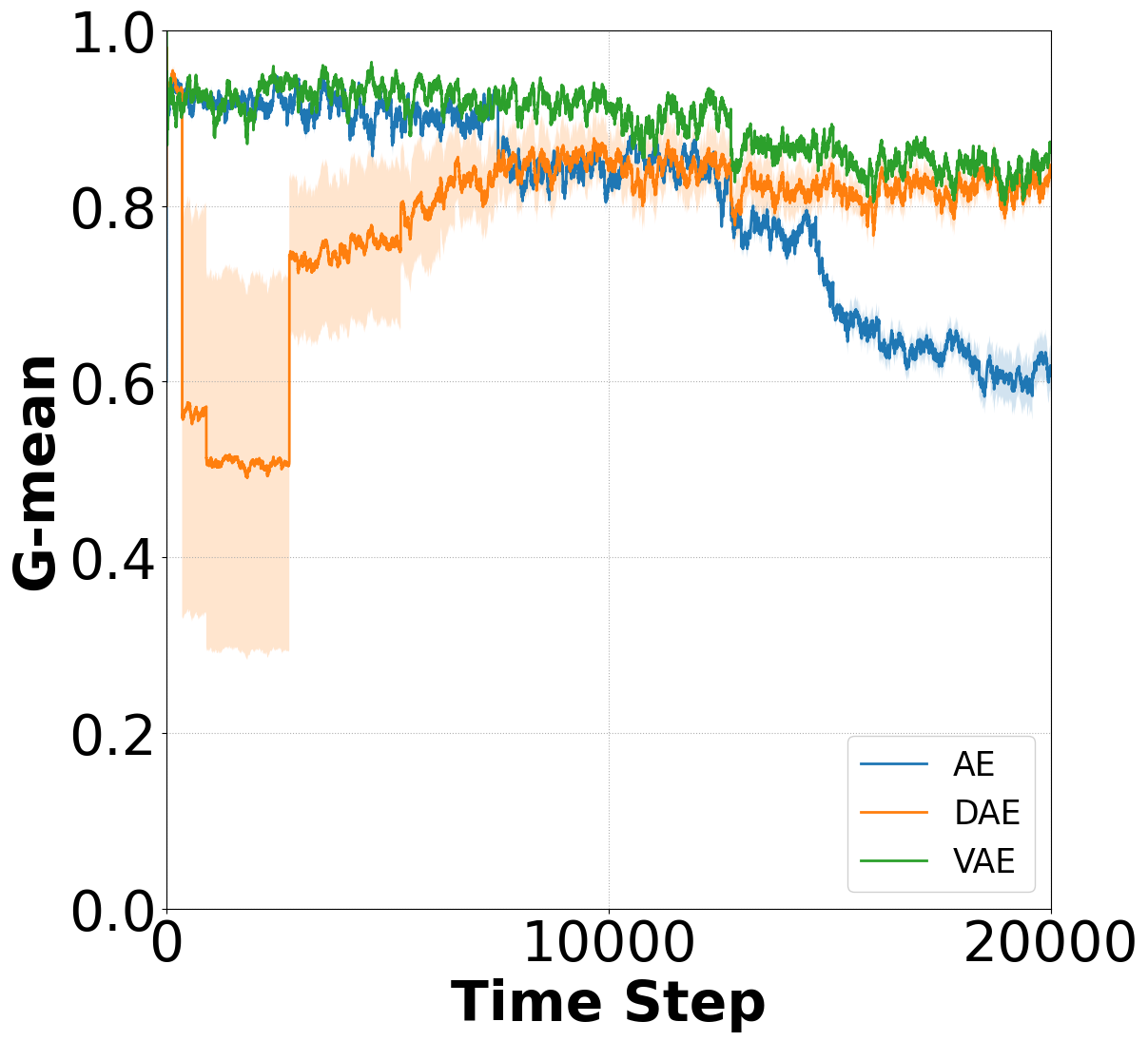} 
  \caption{Sine extreme}
  \label{fig:sine_extreme_diff_ae}
 \end{subfigure}%

\caption{Comparison of different autoencoder architectures in stationary environments.}
\label{fig:compare_ae}
\end{figure}

\begin{figure}[!h]
\centering

\begin{subfigure}{.5\linewidth}
  \centering
  \includegraphics[width=.95\linewidth]{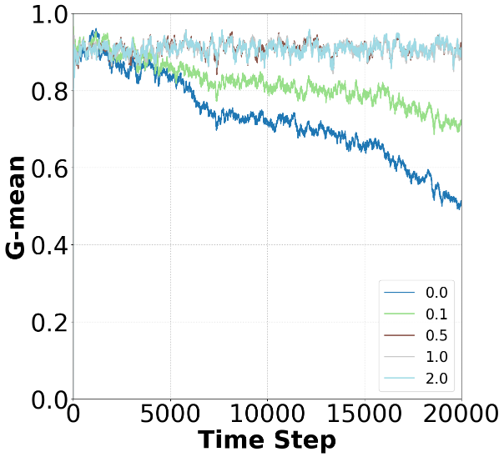}
  \caption{Sea severe}
\end{subfigure}%
\begin{subfigure}{.5\linewidth}
  \centering
  \includegraphics[width=.95\linewidth]{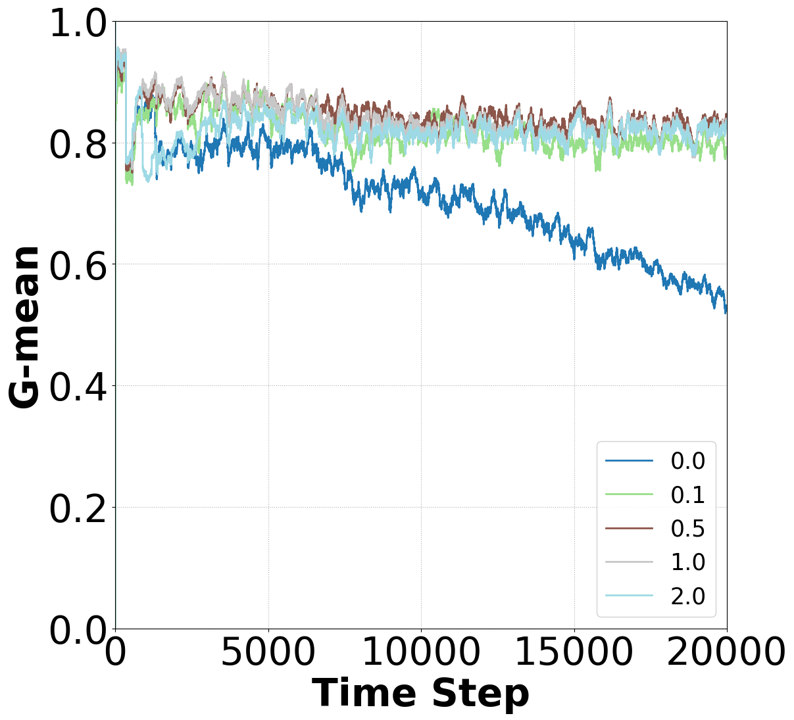}
  \caption{Sine severe}
\end{subfigure}

\begin{subfigure}{\linewidth}
  \centering
  \includegraphics[width=.6\linewidth]{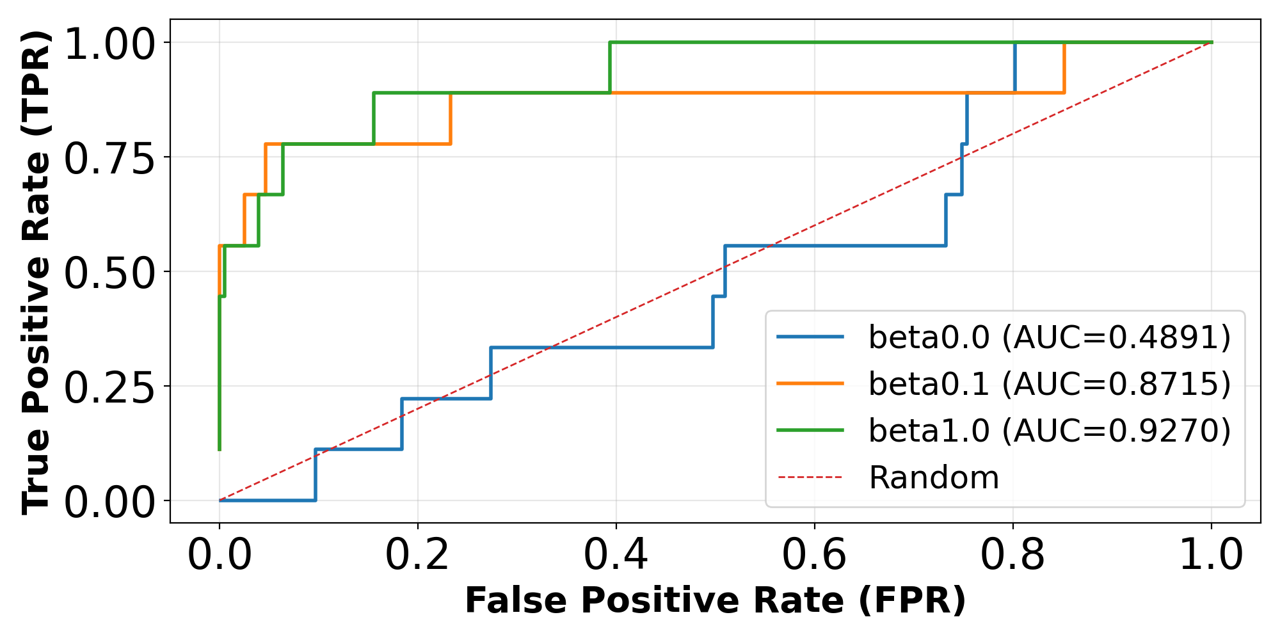}
  \caption{ROC comparison (Sine severe)}
  \label{fig:beta_roc}
\end{subfigure}

\caption{Model performance in stationary environments with different $\beta$.}
\label{fig:beta_compare}
\end{figure}

\begin{figure}[!h]
\begin{subfigure}{.5\columnwidth}
  \centering
  \includegraphics[width=.95\columnwidth]{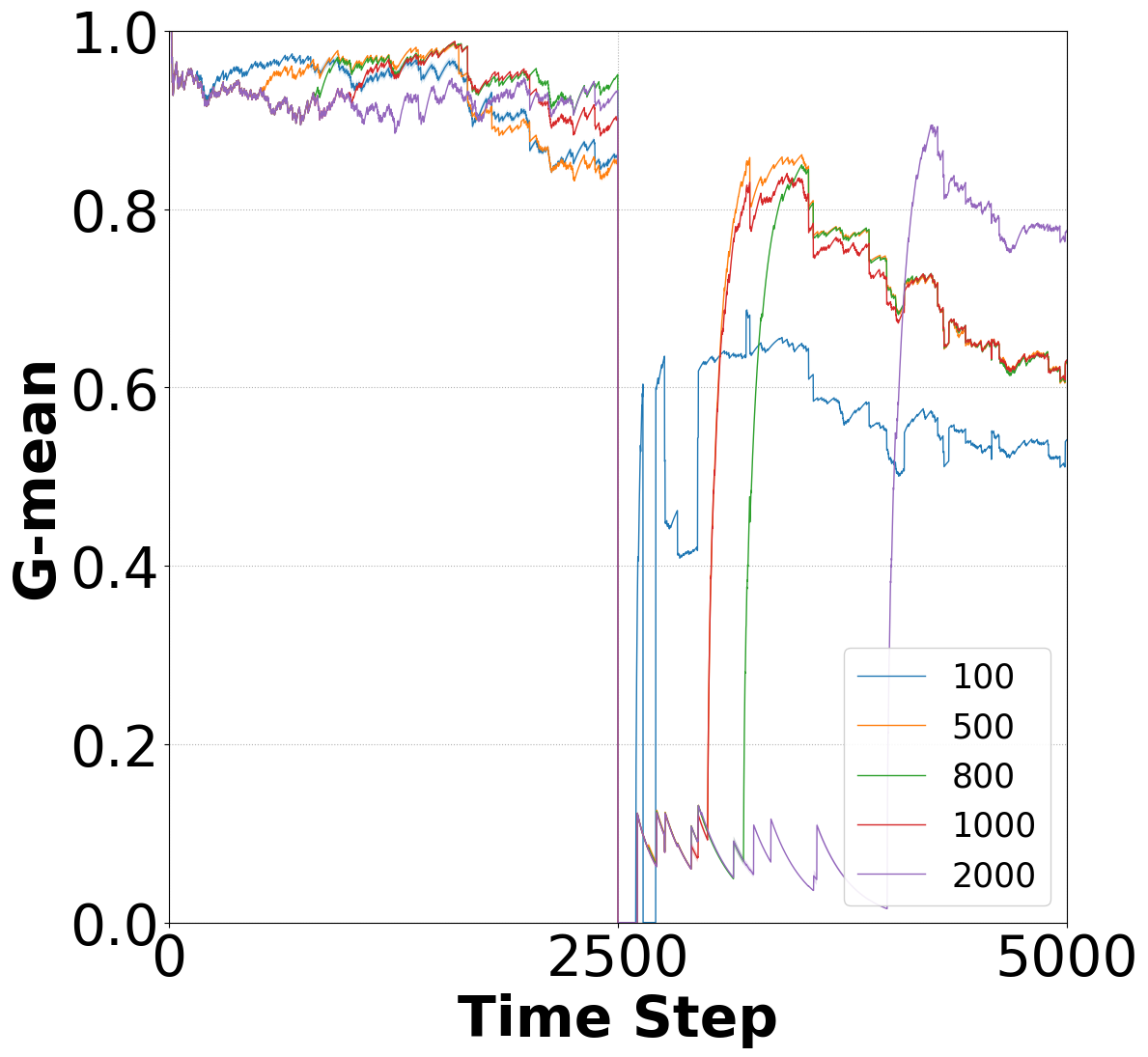}
  \caption{MNIST-23 severe}
  \label{fig:fig_mnist23_severe_win}
\end{subfigure}%
\begin{subfigure}{.5\columnwidth}
  \centering
  \includegraphics[width=.9\columnwidth]{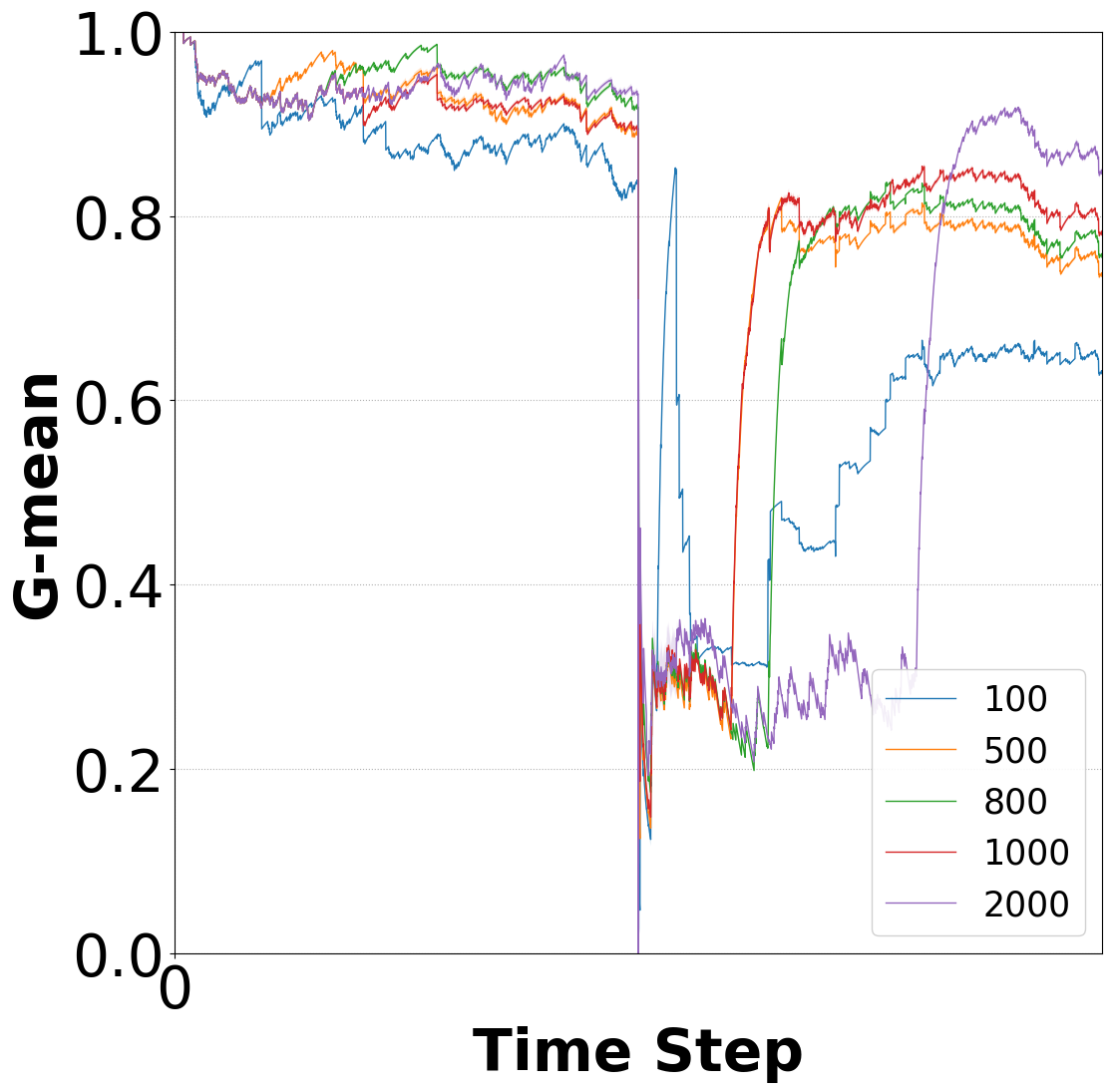}
  \caption{MNIST-01 severe}
  \label{fig:sfig_mnist_m01_severe_win}
\end{subfigure}
\caption{Model performance of in non-stationary environments with different $W_{train}$.}
\label{fig:compare_win}
\end{figure}

\begin{figure}[!h]
 \begin{subfigure}{.5\columnwidth}
  \centering
  \includegraphics[width=0.95\columnwidth]{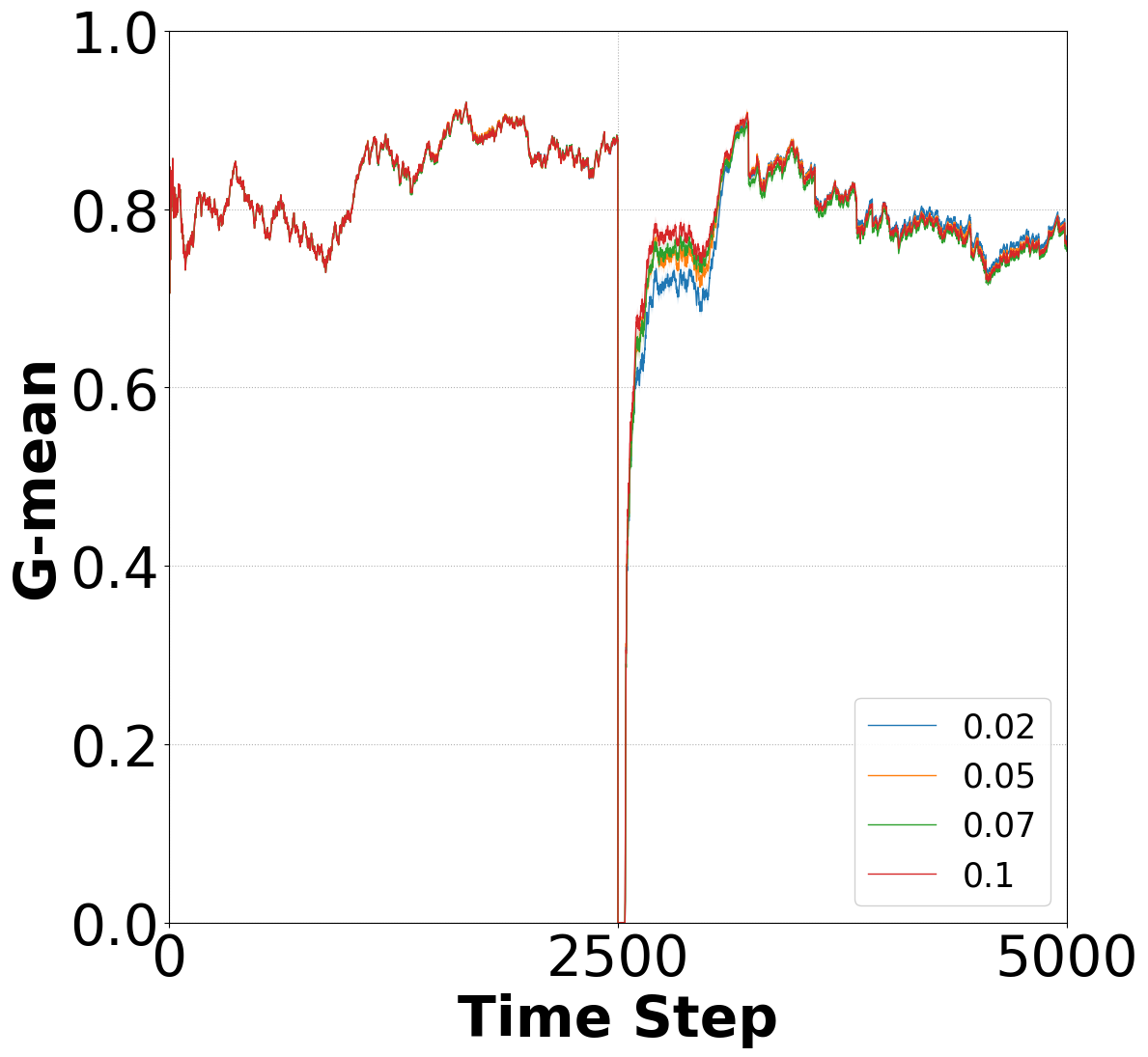} 
  \caption{Model performnce}
 \end{subfigure}%
   \begin{subfigure}{.5\columnwidth}
  \centering
  \includegraphics[width=0.95\columnwidth]{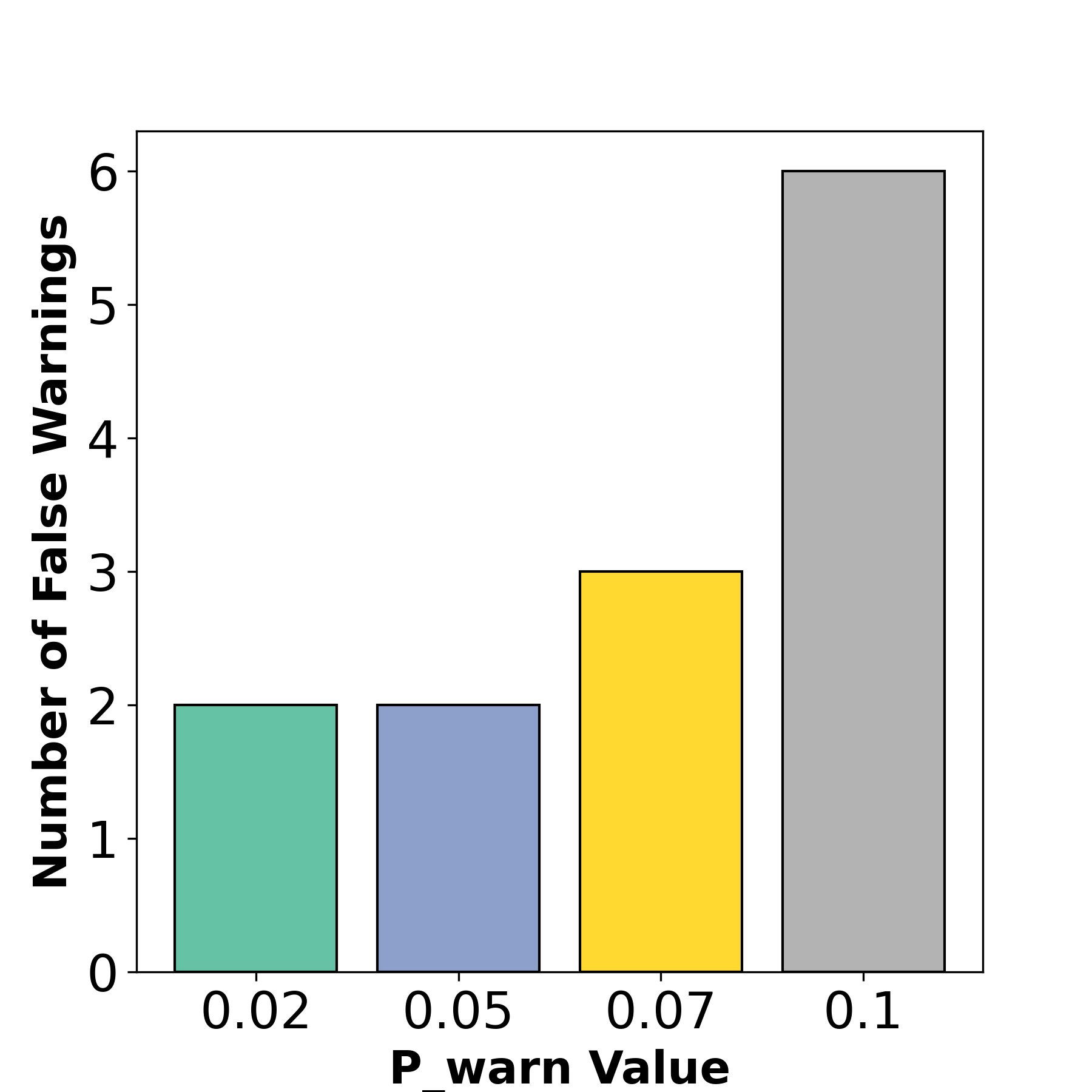} 
  \caption{Number of warnings}
 \end{subfigure}%
 
\caption{Effect of $P_{\text{warn}}$ on performance and warning count (MNIST-23).}
\label{fig:pwarn}
\end{figure}

\begin{figure}[!h]
 \begin{subfigure}{.5\columnwidth}
  \centering
  \includegraphics[width=0.96\columnwidth]{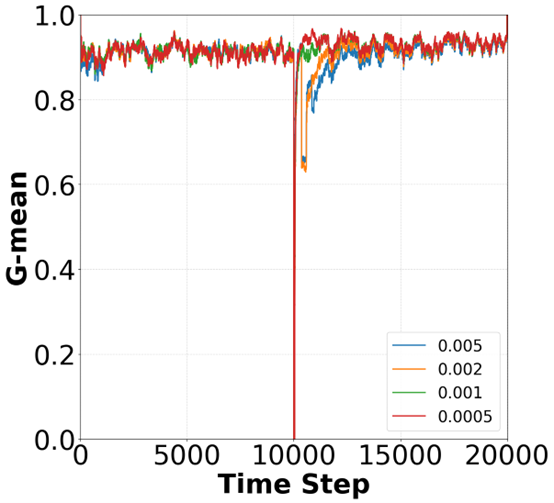} 
  \caption{Model performance}
 \end{subfigure}%
   \begin{subfigure}{.5\columnwidth}
  \centering
  \includegraphics[width=0.89\columnwidth]{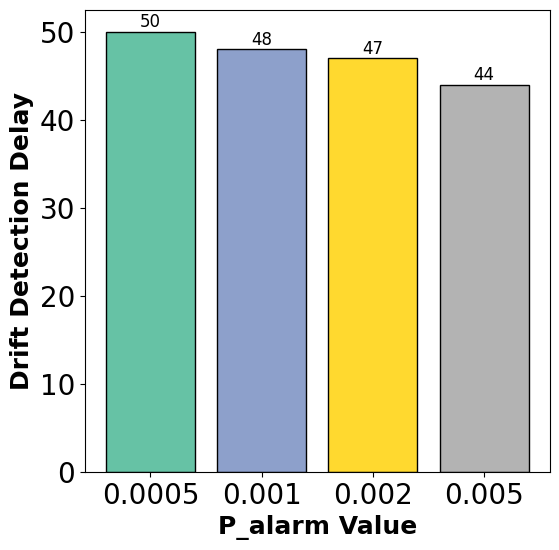} 
  \caption{Drift detection delay}
 \end{subfigure}%
 
\caption{Effect of $P_{\text{alarm}}$ on performance and drift detection delay (Sea).}
\label{fig:palarm}
\end{figure}

\begin{figure}[!h]
 \begin{subfigure}{.5\columnwidth}
  \centering
  \includegraphics[width=0.95\columnwidth]{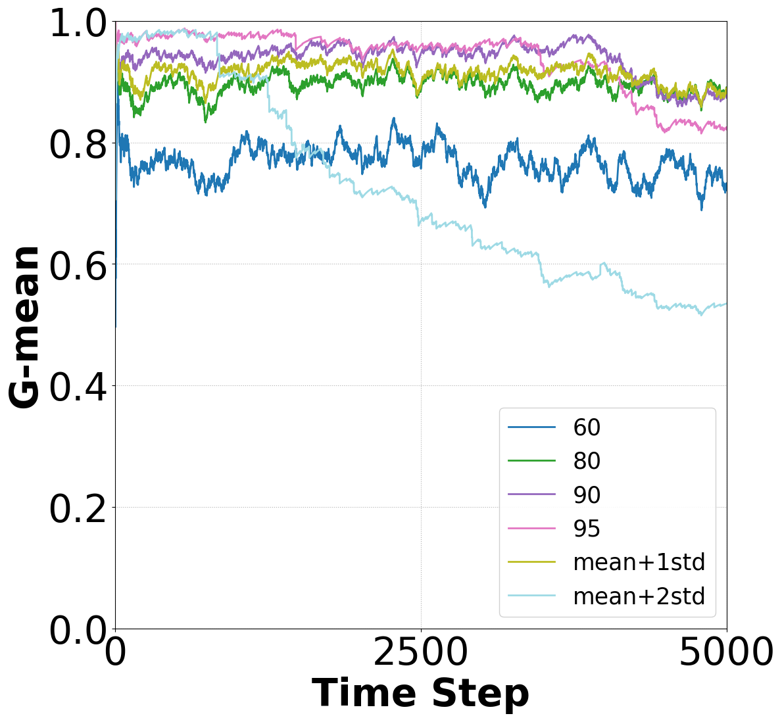} 
  \caption{MNIST-01}
  \label{fig:mnist_01_severe_adap_fix}
 \end{subfigure}%
\begin{subfigure}{.5\columnwidth}
  \centering
  \includegraphics[width=0.98\columnwidth]{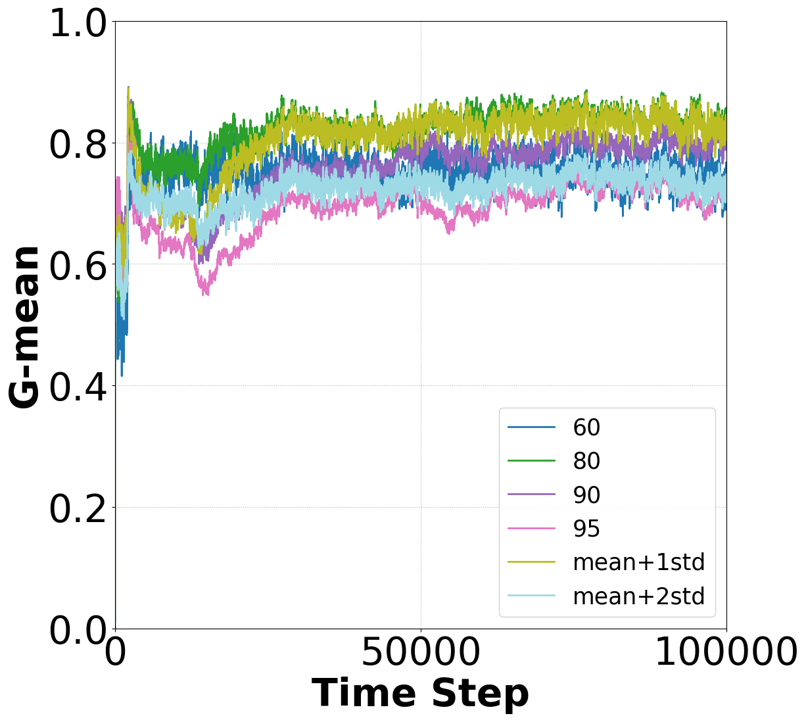}
  \caption{Forest}
  \label{fig:forest_severe_adap_fix}
\end{subfigure}%
 
 \begin{subfigure}{.5\columnwidth}
  \centering
  \includegraphics[width=0.95\columnwidth]{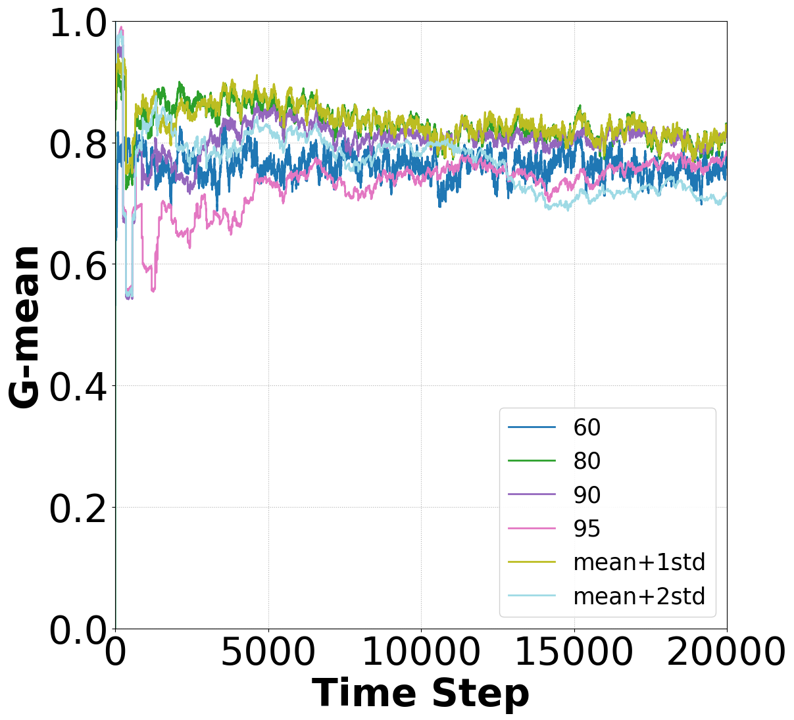} 
  \caption{Sine}
  \label{fig:sine_severe_adap_fix}
 \end{subfigure}%
   \begin{subfigure}{.5\columnwidth}
  \centering
  \includegraphics[width=0.95\columnwidth]{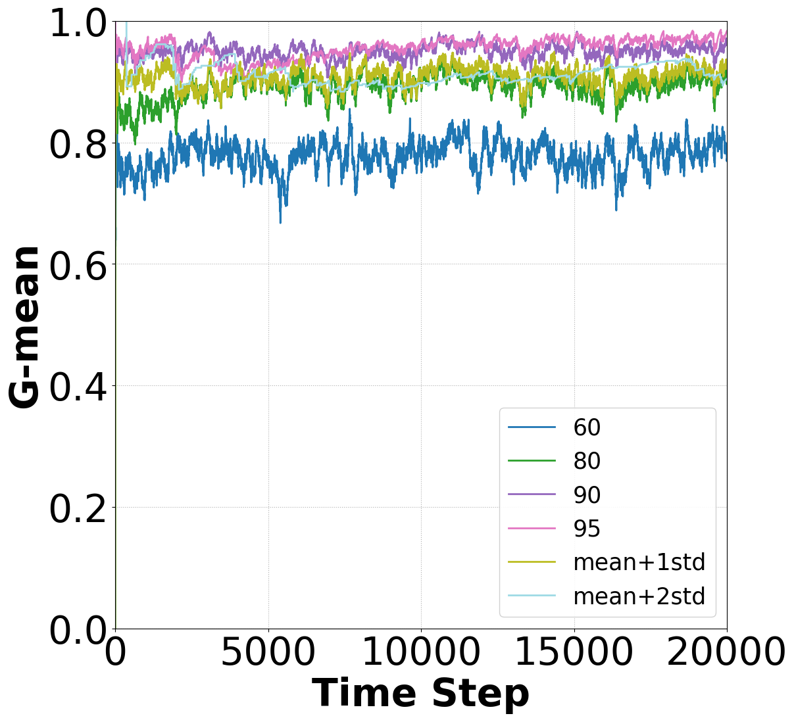} 
  \caption{Sea}
  \label{fig:sea_severe_adap_fix}
 \end{subfigure}%
 
\caption{Comparison between fixed percentile and adaptive threshold in stationary environments.}
\label{fig:thre_severe}
\end{figure}

\section{Experimental Results}\label{sec:exp_results}

In the following ablation studies and comparative study, we investigate several key mechanisms of the proposed framework, including the impact of drift detection (VAE++ vs.\ VAE++DD), incremental learning (Baseline vs.\ VAE++ES), predictor ensembling (VAE++ vs.\ VAE++ES), and drift detector ensembling (VAE++ES vs.\ VAE++ESDD). Furthermore, we compare the proposed method against representative approaches with distinct methodological characteristics: tree- and density-based anomaly detection methods (iForest++, LOF++) \textit{vs.} neural network (NN)-based approaches; clustering- and prototype-driven methods (CPOCEDS) \textit{vs.} non-clustering approaches; ensemble-based frameworks (ARCUS, SEAD, VAE++ES, VAE++ESDD, Baseline) \textit{vs.} single-model approaches; and methods with explicit drift detection (StrAEm++DD, VAE++ESDD) \textit{vs.} passive-adaptation methods.

Hyper-parameter tuning has been performed for each dataset to determine the most suitable values in each case. To facilitate the reprducibility of our results, Table~\ref{tab:params_nn} provides the hyper-paramerters for (V)AE++(ES)(DD) where fully connected neural networks are used. Importantly, the proposed method appears to be robust and many values remain the same for all cases, for instance, the learning rate has been set to 0.001. We set ensemble size $n$ as 10, $P_{thre}$ to 1 and $D_{thre}$ to 10. The choice of $D_{thre}$ as 10 is aimed at minimizing false alarms. During the experiments, no true drift was missed with this setting. Since smaller values of \(D_{\text{thre}}\) only introduce additional false alarms, we keep \(D_{\text{thre}} = 10\). $W_{train}$ is set to 3000 and $\gamma=2000$. For the `MNIST' dataset, considering its volume, we set $W_{train}=1500$ and $\gamma=1000$. Each window $mov_{train}(i)$ is assigned a random size $W_{train}(i)$ accordingly. Other parameters include $expiry\_time=100$, $P_{warn}=0.01$, and $P_{alarm}=0.001$. The initialization of $mov_{ESwarn}$, $ref_{driftx}$, and $mov_{driftx}$ is performed with $W_{drift}(i)=random (180,220)$. The relevant information for other methods can be found in the released code. To ensure statistical significance, we conduct simulation experiments with 20 repetitions for each scenario. In each repetition, we calculate the prequential metric at every time step. The results are then averaged, and error bars are added to represent the standard error around the mean.

\subsection{Ablation studies}

In this section, we conduct a series of ablation studies to analyze the impact of key design choices in the proposed framework, as summarized in Table~\ref{tab:comp_resource}.
For datasets with recurrent drifts, the ablation study focuses on the data stream before the second drift, while the comparative study considers the entire stream with all recurrent drifts.

\subsubsection{Different Autoencoder types}
Utilizing the stationary datasets with severe and extreme class imbalance rates and adopting the adaptive threshold method, we compare here incremental learning in the stationary environment with various autoencoder architectures.

As illustrated in Fig.~\ref{fig:compare_ae}, the results demonstrate the superior stability of incremental learning with VAE compared to AE or Denoising Autoencoder (DAE). In this experiment, we consider the DAE with Gaussian distributions $G(0, 2)$. The model performance of AE++ and DAE++ exhibits significant variations across the datasets, whereas VAE++ consistently achieves excellent performance across all datasets. One possible explanation for this phenomenon is that the variability component in the architecture of VAE allows it to cover a wider range of possible input values for the decoder. In contrast, a regular autoencoder may have significant gaps or regions in the input space where the decoder cannot accurately reconstruct the corresponding input values. The reduced presence of these gaps in the VAE results in a more comprehensive coverage of input values, leading to improved performance compared to the AE.

\subsubsection{Selection of $\beta$}
In this experiment, we investigate the effect of the parameter $\beta$ on anomaly detection performance in a stationary environment using a single model without drift detection. The parameter $\beta$ controls the contribution of the KL divergence term in the total loss, as defined in Eq.~\ref{eq:vae}.

As shown in the first two subfigures of Fig.~\ref{fig:beta_compare}, $\beta=0$ (i.e., a standard autoencoder) yields the poorest performance, while introducing KL regularization significantly improves and stabilizes detection accuracy. In particular, $\beta=0.5$, $\beta=1.0$, and $\beta=2.0$ exhibit comparable and consistently strong performance, in contrast to the substantially weaker results observed for $\beta=0$ and $\beta=0.1$. Accordingly, we focus the PROC analysis on representative settings with clear performance differences, namely $\beta=0$, $\beta=0.1$, and $\beta=1.0$. As shown in Fig.~\ref{fig:beta_roc}, the discriminative capability improves monotonically with increasing $\beta$ under stationary conditions. The PROC analysis is conducted by evaluating only the final 1000 samples of the data stream. As $\beta$ increases, the AUC rises from 0.4891 ($\beta=0$) to 0.8715 ($\beta=0.1$), and reaches its maximum of 0.9270 for $\beta=1.0$. This trend indicates that stronger KL regularization enhances latent-space separability and improves anomaly score ranking without sacrificing detection sensitivity. Therefore, $\beta=1.0$ is adopted in this study, as it shows the best discriminative performance under stationary conditions.

\subsubsection{Role of the training window size}
In this experiment, we investigate the effect of the training window size on anomaly detection performance using a single model without a drift detection mechanism. 

As shown in Fig.~\ref{fig:compare_win}, the model’s performance is evaluated across different training window sizes (100, 500, 800, 1000, and 2000). The results indicate that the window size has minimal influence on performance before the drift, primarily due to the model’s pre-training. However, after the drift occurs, the model with the largest window size (2000) adapts more slowly to the new concept than that with the smallest window (100), yet ultimately achieves higher performance. Because neither extreme is desirable in practice, the training window size must balance these two competing objectives. In our experiments, a window of 2000 delivered the best post-drift recovery. A window of 1000 produced the next-best results, though with a clear performance gap relative to 2000, but with a considerably shorter adaptation delay. 

Therefore, to balance drift adaptation speed and stability, the training window size $W_{\text{train}}$ is set to 3000, with $\gamma = 2000$, ensuring that $W_{\text{train}}(i)$ varies between 1000 and 3000 during training. This design enables some models to adapt quickly (smaller windows) while others provide more stable long-term performance (larger windows), effectively capturing both ends of the trade-off.

\subsubsection{Role of the P-value}

In this experiment, we investigate the effect of the significance thresholds $P_{warn}$ and $P_{alarm}$ on a single model.

\paragraph{Effect of $P_{warn}$.}
We first analyze the effect of $P_{warn}$ on the MNIST-23 dataset with severe class imbalance, where drift occurs at $t=2500$, while fixing $P_{alarm}=0.001$. As shown in Fig.~\ref{fig:pwarn}, varying $P_{warn}$ has little impact on overall performance but increases the number of false warnings as $P_{warn}$ grows, consistent with Eq.~\ref{eq:pvalue}. Accordingly, we set $P_{warn}=0.01$ in subsequent experiments to limit unnecessary warnings.

\paragraph{Effect of $P_{alarm}$.}

We evaluate the effect of $P_{alarm}$ on the Sea dataset with severe imbalance, where drift occurs at $t=10000$. As shown in Fig.~\ref{fig:palarm}, smaller $P_{alarm}$ values improve post-drift performance, while larger values reduce detection delay. Balancing this trade-off, we set $P_{alarm}=0.001$ for subsequent experiments. More generally, overly small $P_{alarm}$ values may miss true drifts, whereas large values can increase false alarms; thus, $P_{alarm}$ can be tuned based on prior knowledge of expected drift severity when available.

\subsubsection{Fixed percentile vs adaptive threshold}

In this experiment, we compare the impact of fixed percentile thresholds and adaptive thresholds on model performance in stationary environments under severe class imbalance. A VAE is adopted as the base model. The fixed percentile threshold parameter $b$ is set to 60, 80, 90, and 95. In addition, two adaptive thresholding strategies are evaluated: one based on Eq.~\ref{eq:adtthreshold} using a `mean+1std' rule, and a more conservative variant using `mean+2std'.

Fig.~\ref{fig:thre_severe} shows that for the MNIST-01 and Sea datasets, model performance generally improves as the percentile threshold increases. However, a different trend is observed for the Forest dataset. Notably, the adaptive thresholding strategy based on `mean+1std' consistently achieves strong and stable performance across datasets. This is because the adaptive mechanism dynamically adjusts to changes in the reconstruction error distribution. When anomalies produce larger reconstruction errors, both the mean and standard deviation increase, leading to a higher threshold that improves separation between normal and anomalous samples. Conversely, when anomalies are subtle and induce smaller error shifts, the threshold is reduced, preserving detection sensitivity. By contrast, the `mean+2std' strategy exhibits larger performance variability, particularly on the MNIST-01 dataset. The excessively high threshold reduces the sensitivity of the detector, causing anomalous samples to remain below the decision boundary and thus go undetected, especially when anomalies induce only modest increases in reconstruction error. Overall, the `mean+1std' criterion provides a more favorable trade-off between detection sensitivity and robustness under severe class imbalance. Based on the above analysis, `mean+1std' is selected as the threshold setting.

\subsubsection{Role of the prediction threshold}
\begin{figure*}[!t]
  \centering
  \begin{subfigure}{.3\textwidth}
    \centering
    \includegraphics[width=0.8\columnwidth]{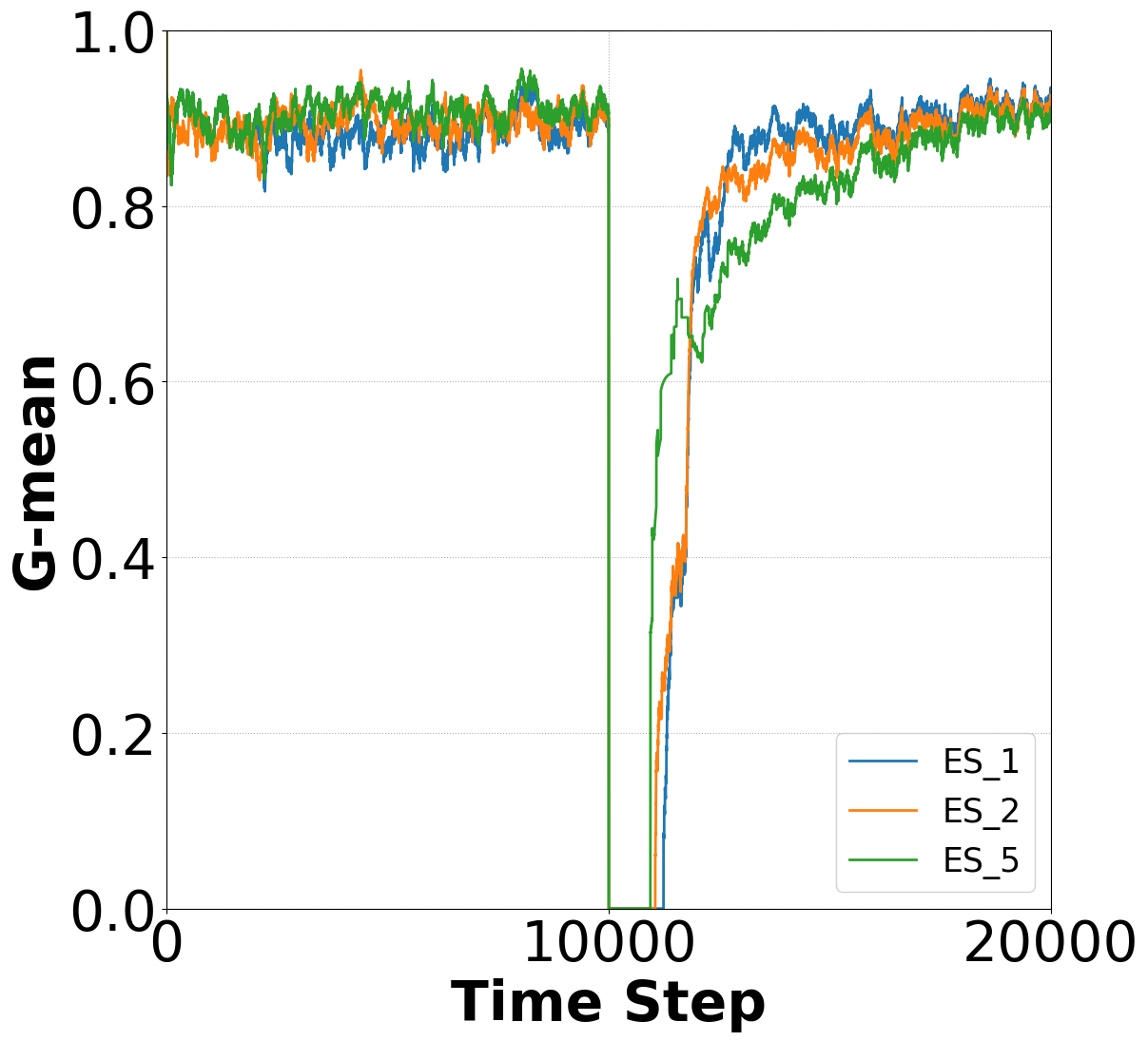}
    \caption{Sea (G-mean)}
  \end{subfigure}%
  \begin{subfigure}{.3\textwidth}
    \centering
    \includegraphics[width=0.8\columnwidth]{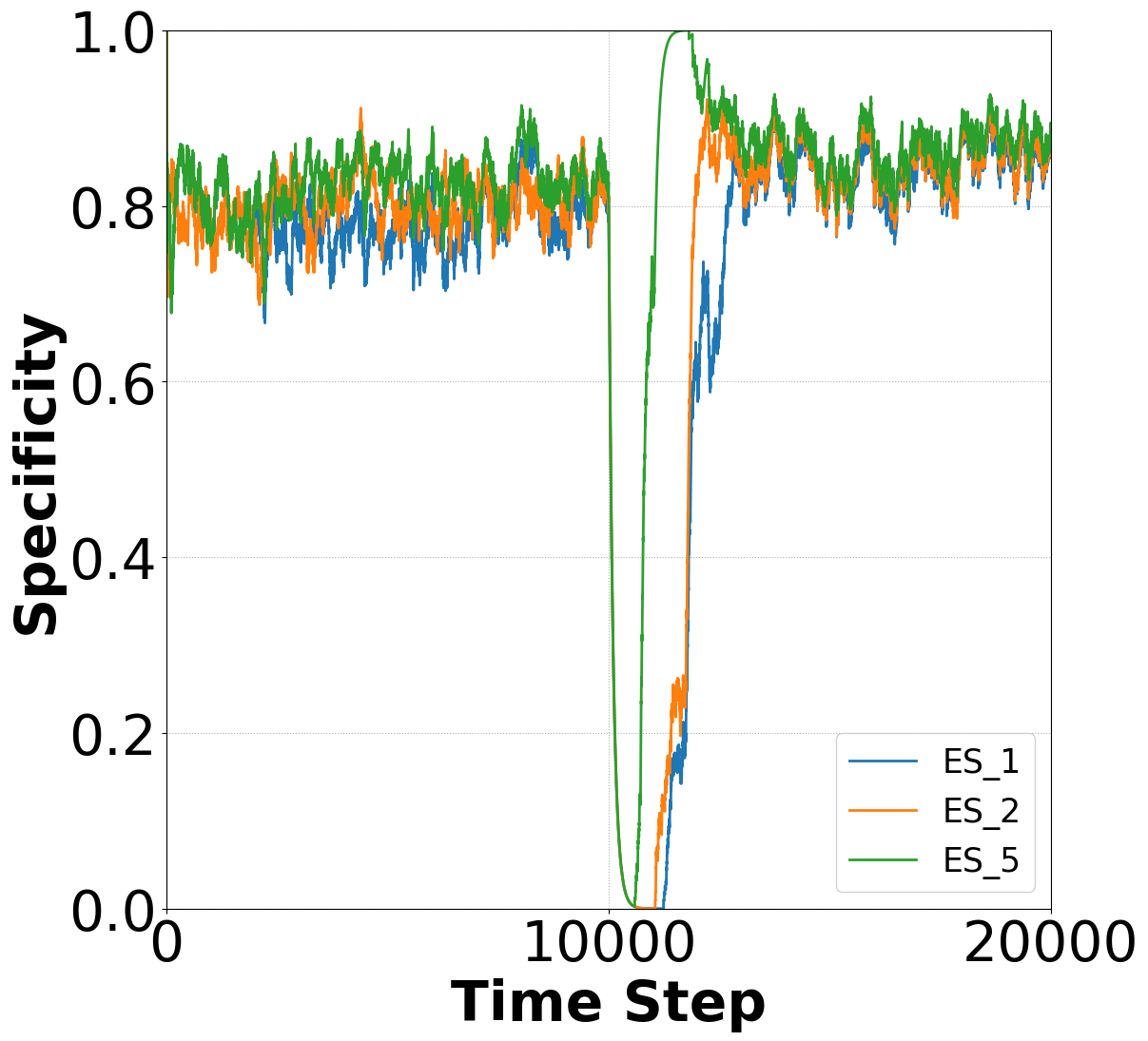}
    \caption{Sea (Specificity)}
  \end{subfigure}%
  \begin{subfigure}{.3\textwidth}
    \centering
    \includegraphics[width=0.8\columnwidth]{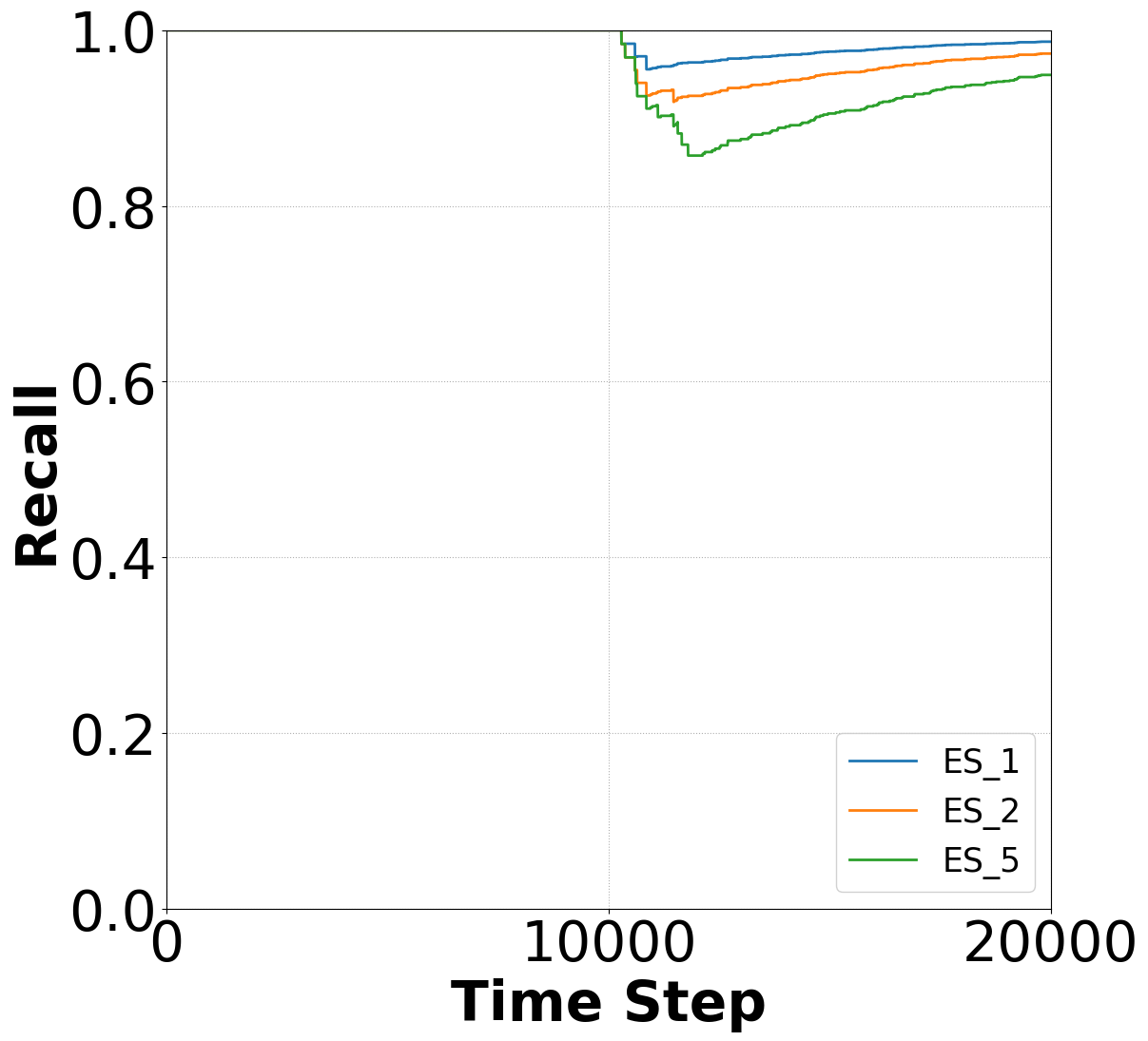}
    \caption{Sea (Recall)}
  \end{subfigure}%

  \vspace{1em}
  \begin{subfigure}{.3\textwidth}
    \centering
    \includegraphics[width=0.8\columnwidth]{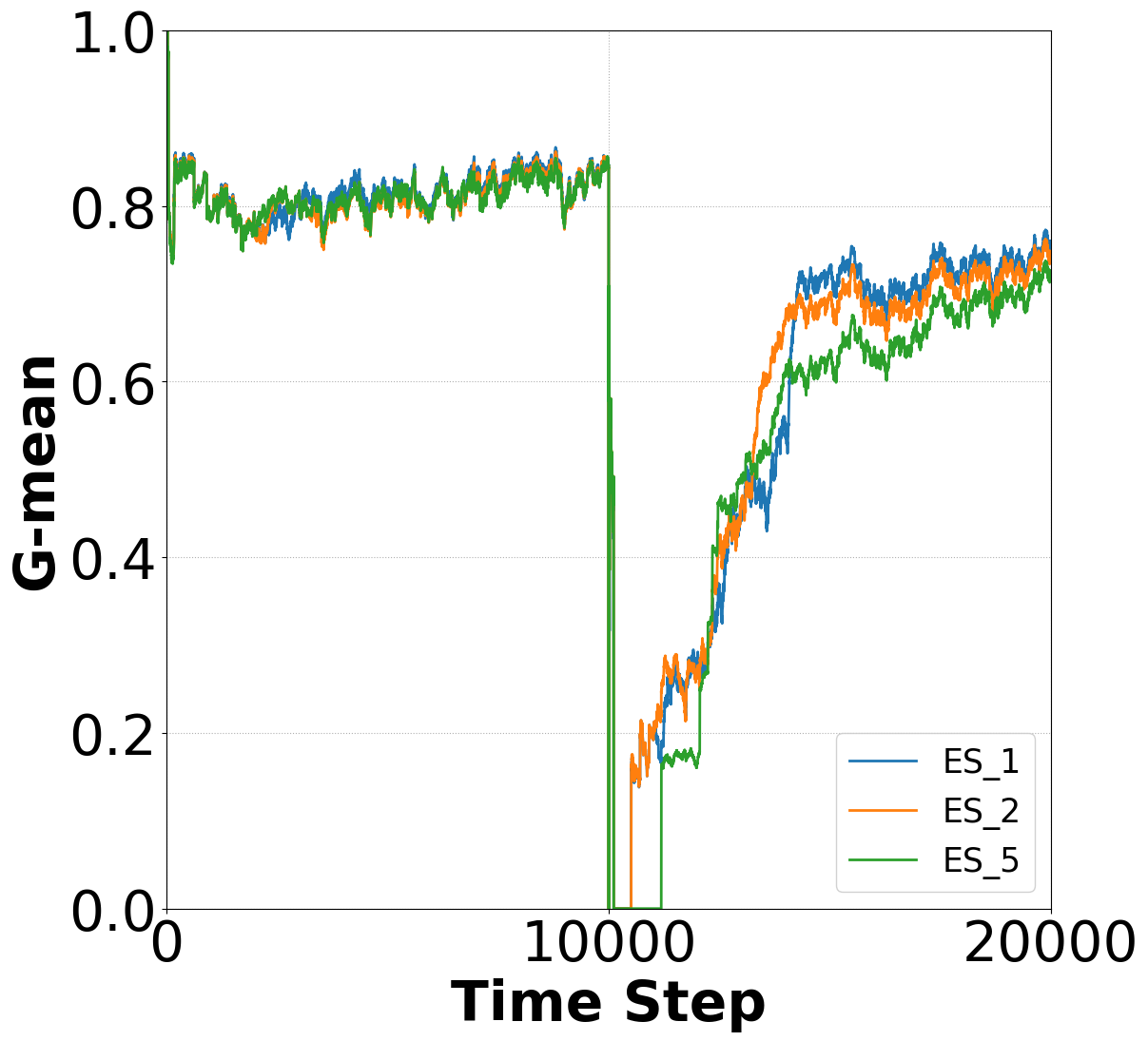}
    \caption{Sine (G-mean)}
  \end{subfigure}%
  \begin{subfigure}{.3\textwidth}
    \centering
    \includegraphics[width=0.8\columnwidth]{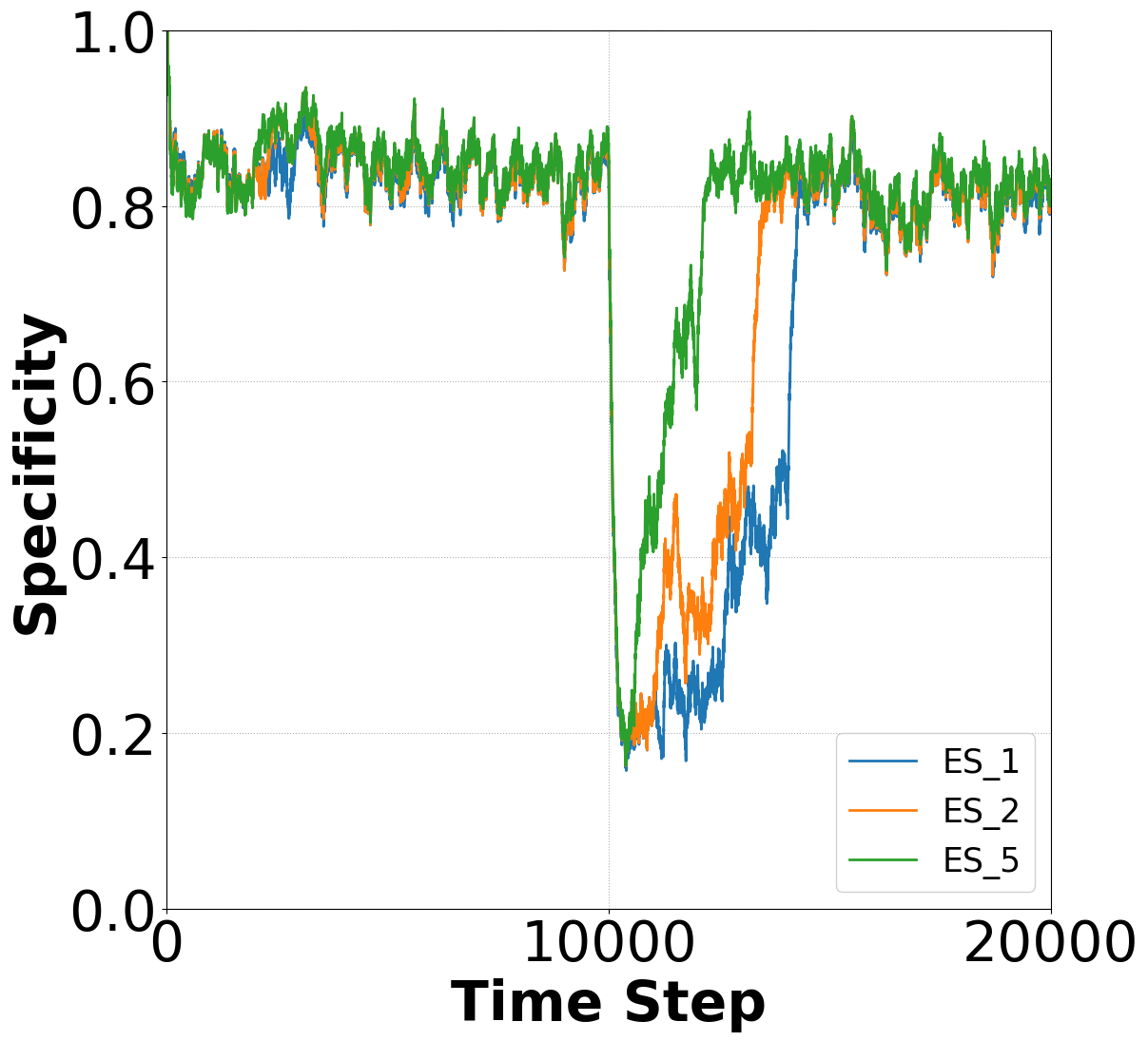}
    \caption{Sine (Specificity)}
  \end{subfigure}%
  \begin{subfigure}{.3\textwidth}
    \centering
    \includegraphics[width=0.8\columnwidth]{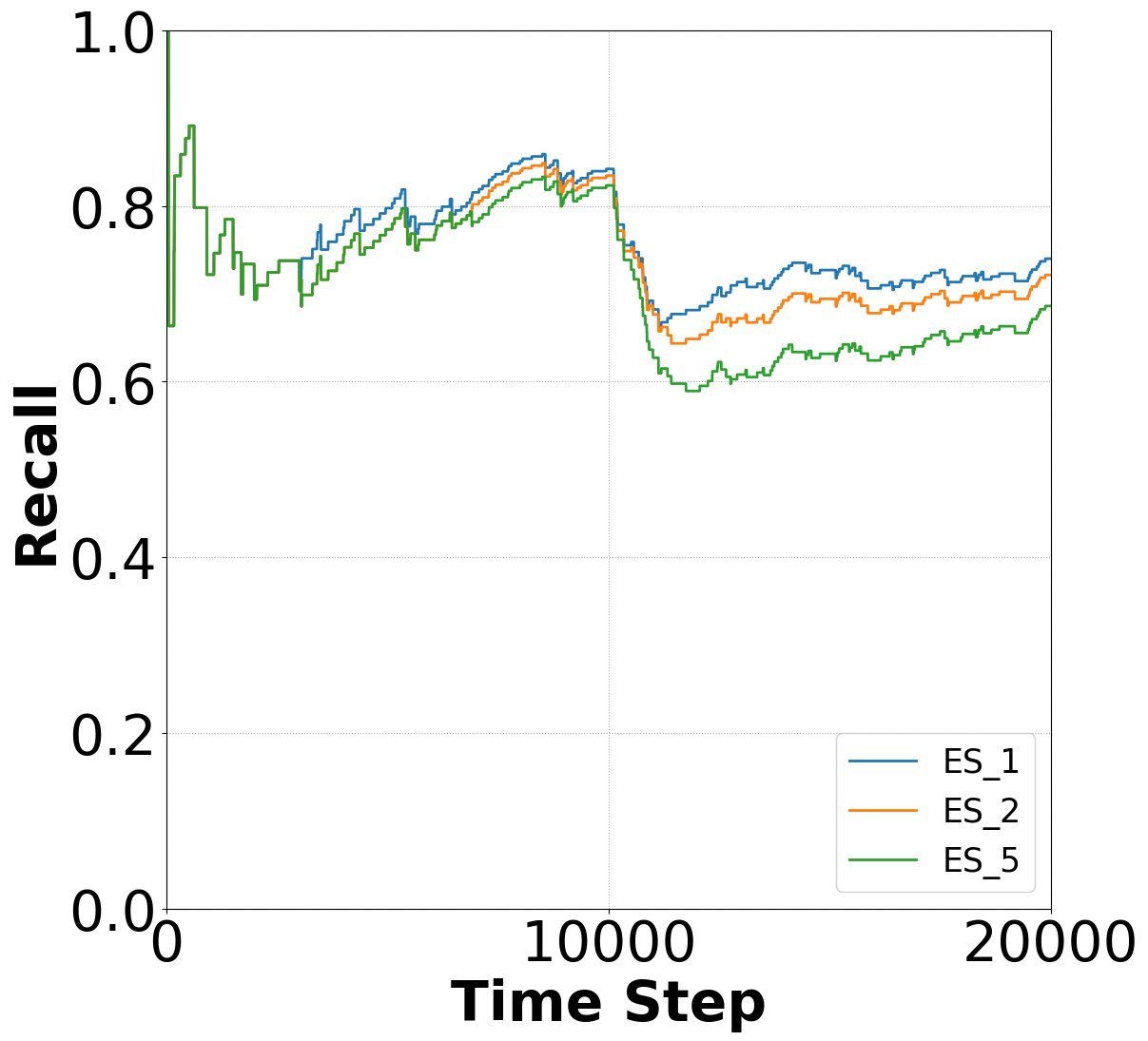}
    \caption{Sine (Recall)}
  \end{subfigure}%

  \vspace{1em}
  \begin{subfigure}{.3\textwidth}
    \centering
    \includegraphics[width=0.8\columnwidth]{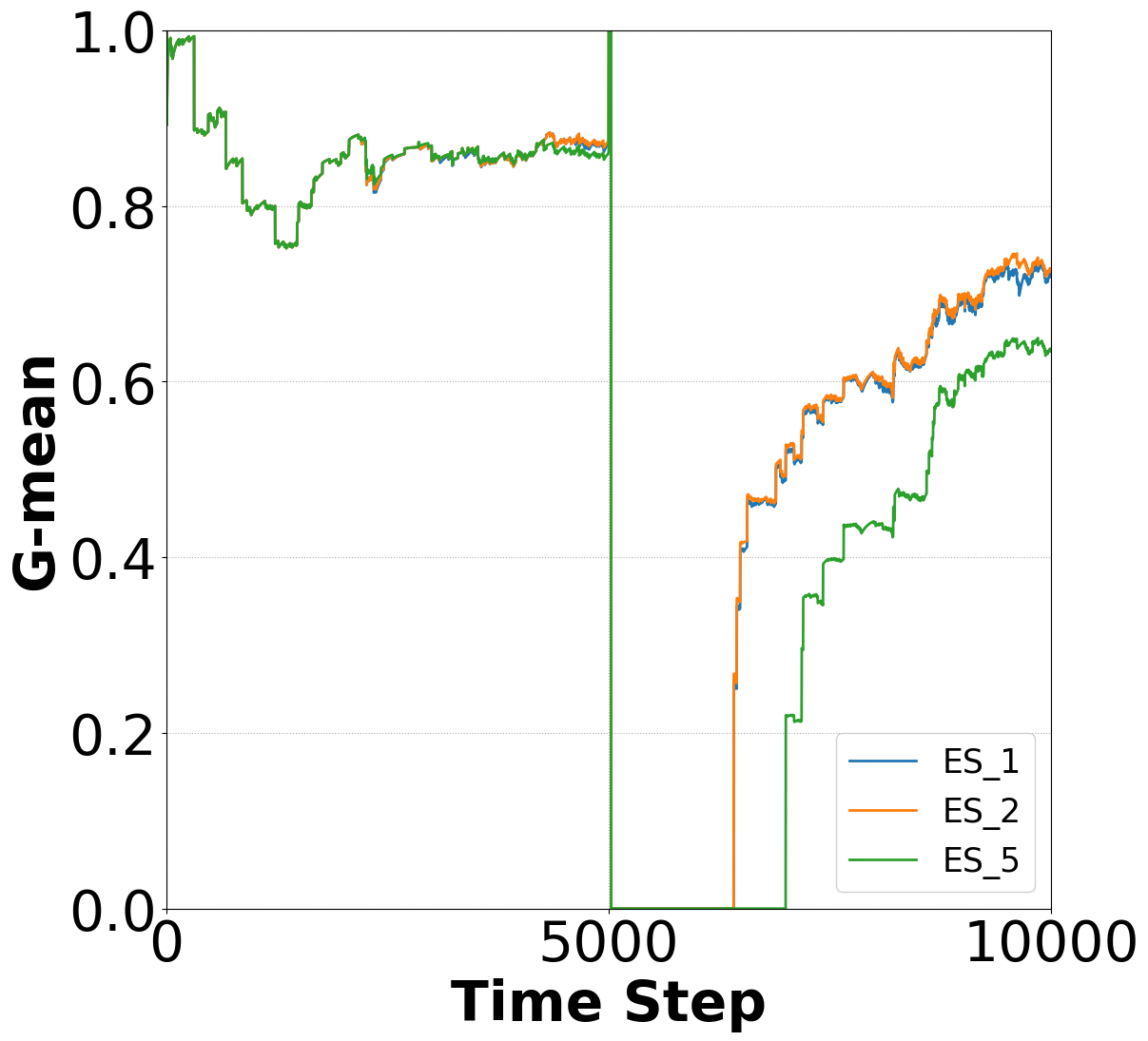}
    \caption{Fraud (G-mean)}
  \end{subfigure}%
  \begin{subfigure}{.3\textwidth}
    \centering
    \includegraphics[width=0.8\columnwidth]{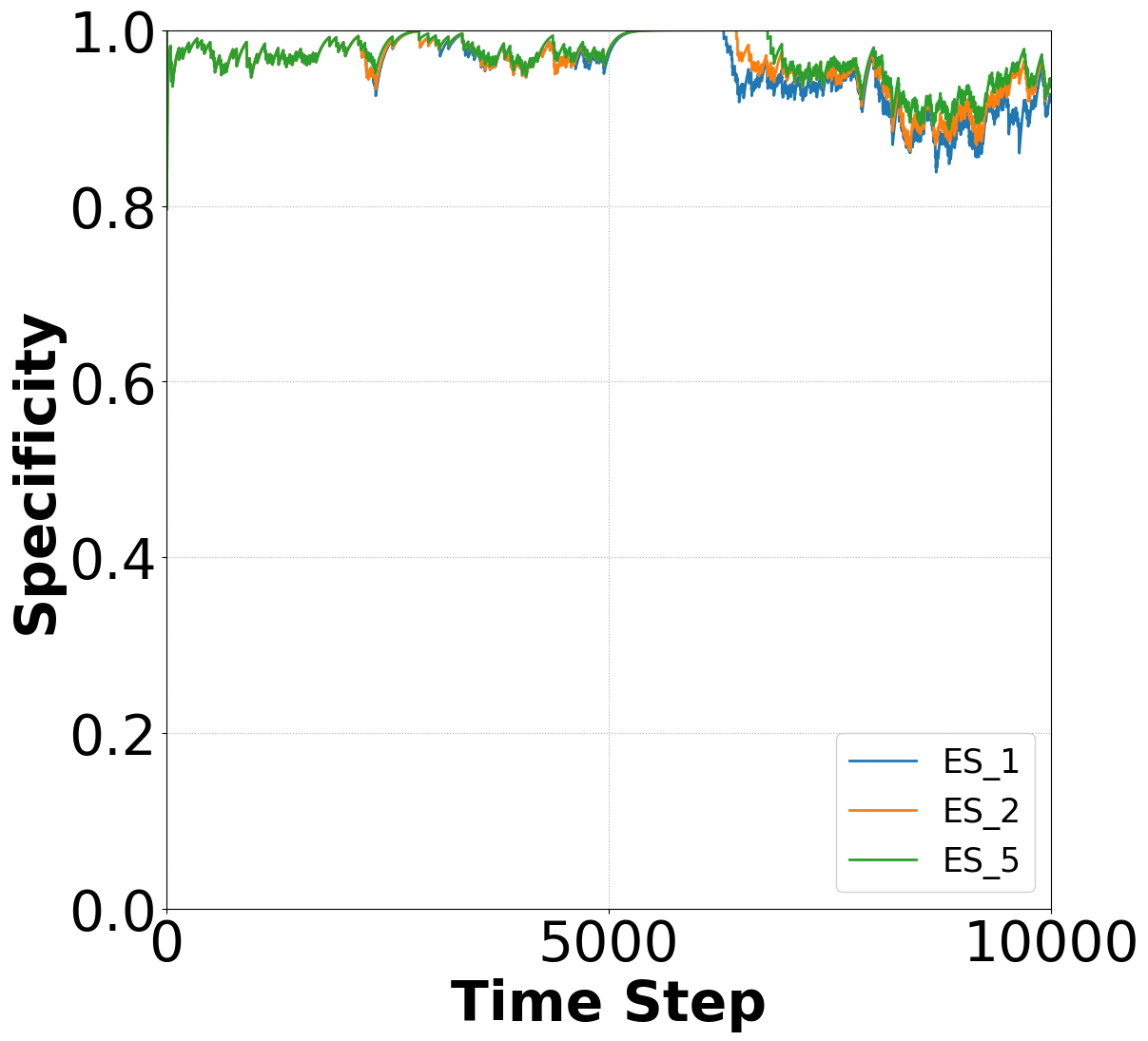}
    \caption{Fraud (Specificity)}
  \end{subfigure}%
  \begin{subfigure}{.3\textwidth}
    \centering
    \includegraphics[width=0.8\columnwidth]{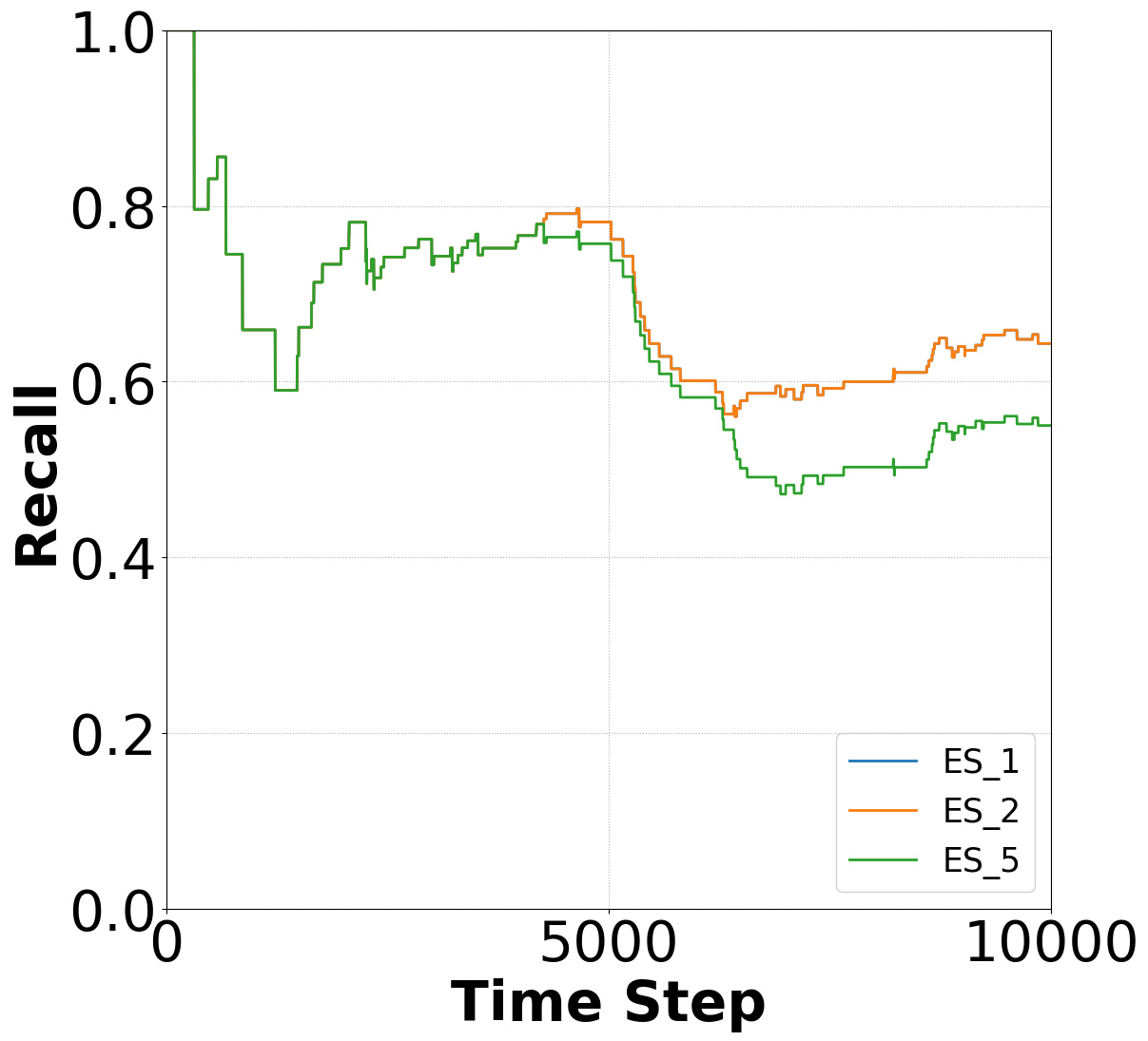}
    \caption{Fraud (Recall)}
  \end{subfigure}%


  \caption{Comparison of different prediction $P_{thre}$ thresholds across datasets in non-stationary environments.}
  \label{fig:compare_pred_index}
\end{figure*}

In this experiment, we investigate the impact of the prediction threshold $P_{thre}$ in VAE++ES. The ensemble size is 10. The labels $ES_1$, $ES_2$, and $ES_5$ correspond to VAE++ES with $P_{thre}$ set to 1, 2, and 5, respectively, as depicted in Fig.~\ref{fig:compare_pred_index}. An instance is classified as anomalous if the number of models predicting it as such exceeds $P_{thre}$. We compare the results with severe imbalance rate using three evaluation metrics.

Different performance metrics are displayed side by side for each dataset to enable clearer visual comparison. From the G-mean results, the model shows inferior performance when $P_{thre} = 5$ compared to the other two settings. For all datasets, decreasing $P_{thre}$ increases recall because more anomalies are detected. Between $P_{thre} = 1$ and $P_{thre} = 2$, the G-mean values are comparable, with slightly better results when $P_{thre} = 1$. In anomaly detection tasks where anomalies are rare, recall is a more important metric since it reflects the system’s ability to identify all true anomalies. Furthermore, the recall obtained with $P_{thre} = 1$ is consistently higher than or equal to that with $P_{thre} = 2$. Considering both G-mean and recall, $P_{thre} = 1$ provides the best balance between detection sensitivity and overall performance. Therefore, $P_{thre} = 1$ is used in the subsequent experiments.

\subsubsection{Role of the drift detection window size}
\begin{figure}[!t]
\begin{subfigure}{.5\columnwidth}
  \centering
  \includegraphics[width=.95\columnwidth]{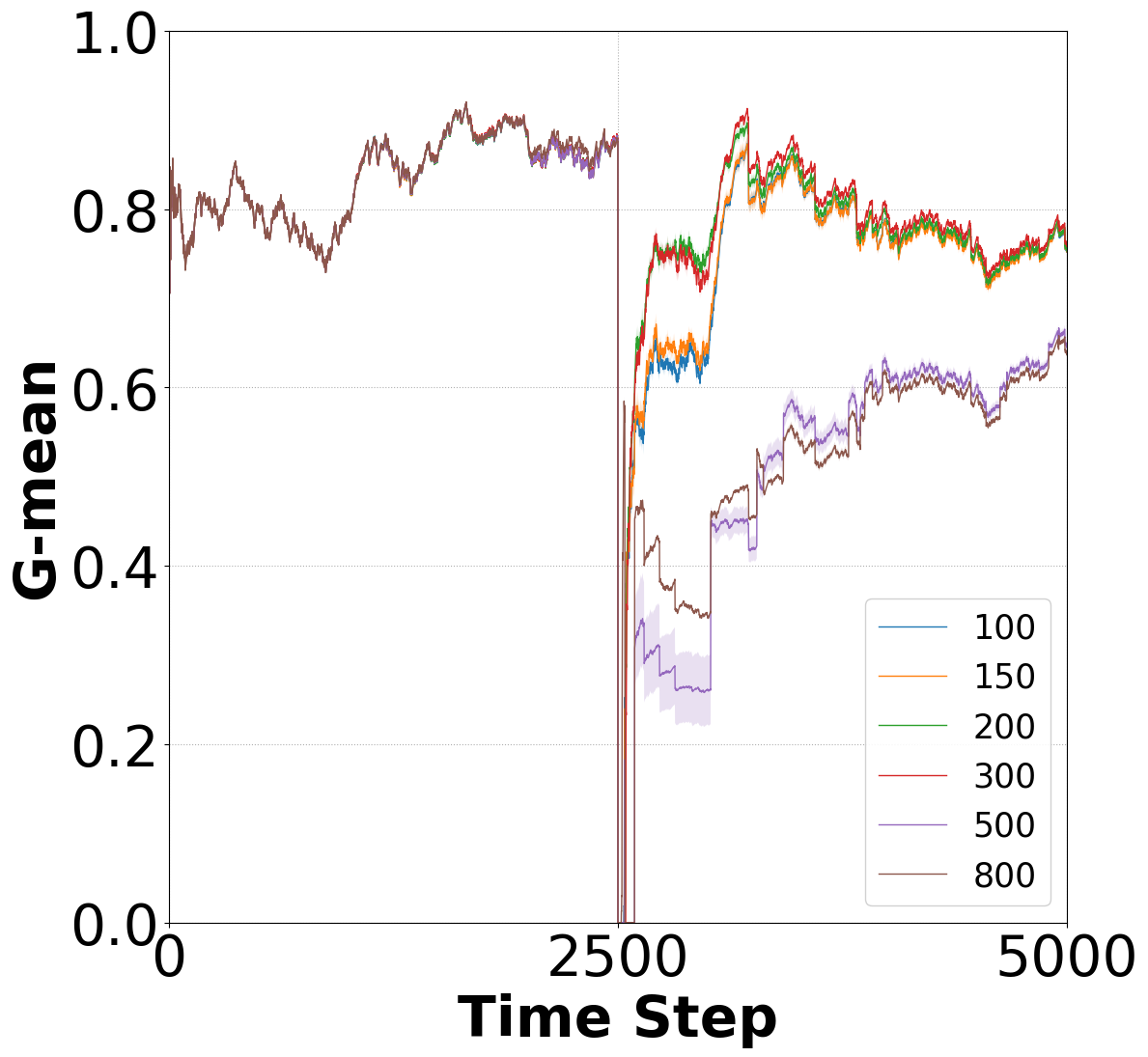}
  \caption{model performance}
  \label{fig:mnist_23_severe_performance_diffwin}
\end{subfigure}%
\begin{subfigure}{.5\columnwidth}
  \centering
  \includegraphics[width=.85\columnwidth]{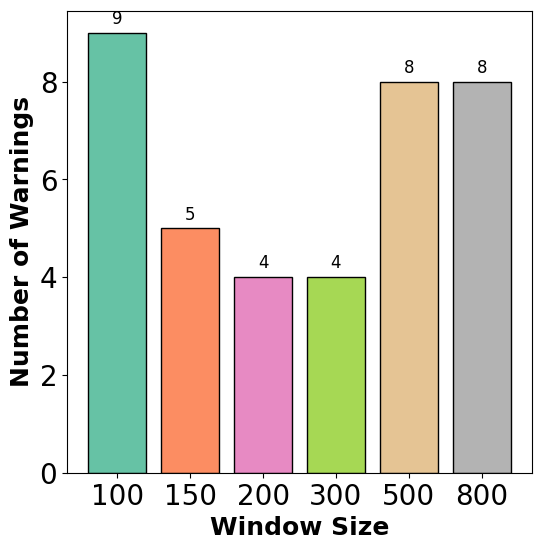}
  \caption{number of warnings}
  \label{fig:sfig_mnist_23_extreme}
\end{subfigure}
\caption{Model performance of in non-stationary environments with different $W_{drift}$ (MNIST-23)}
\label{fig:drift_thre}
\end{figure}

In this experiment, we investigate the influence of the drift detection window size $W_{drift}$ using a single model. A single drift occurs at $t = 2500$. As shown in Fig.~\ref{fig:drift_thre}, $W_{drift} = 200$ and $W_{drift} = 300$ yield the best performance, and both result in only three false warnings, fewer than other window sizes. Therefore, in our proposed framework, we set $W_{drift}(i) = random (180, 220)$ in order to further reduce false warnings and maintain higher detection accuracy.

\begin{table}[H]
\centering
\caption{Computational performance and resource usage of VAE++ESDD under different ensemble sizes.}
\label{tab:comp_resource}
\begin{adjustbox}{width=0.48\textwidth}
\begin{tabular}{|c|c|c|c|c|c|c|}
\hline
Dataset & Ensemble Size & G-mean & Memory(MB) & $T_{\text{stream}}$(s) & $T_{\text{incr}}$(s) & $T_{\text{drift}}$(s) \\ 
\hline
\multirow{2}{*}{Sine}      
 & $n=1$  & 0.704 & 0.040 & 0.105 & 0.256 & 0.828 \\ \cline{2-7} 
 & $n=10$ & 0.791 & 0.400 & 0.106 & 1.630 & 6.086 \\ 
\hline
\multirow{2}{*}{MNIST\_23} 
 & $n=1$  & 0.701 & 4.200 & 0.140 & 1.826 & 1.094 \\ \cline{2-7} 
 & $n=10$ & 0.799 & 42.000 & 0.143 & 11.200 & 8.509 \\ 
\hline
\end{tabular}
\end{adjustbox}
\end{table}

\begin{figure}[!h]
 \begin{subfigure}{.5\columnwidth}
  \centering
  \includegraphics[width=0.95\columnwidth]{ 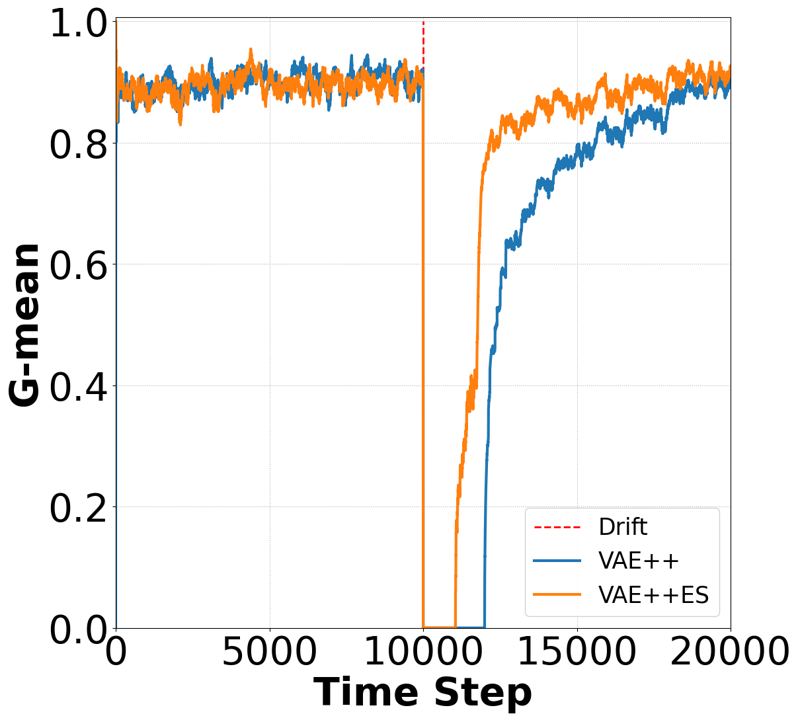}
  \caption{Sea}
  \label{fig:sea_es}
 \end{subfigure}%
  \begin{subfigure}{.5\columnwidth}
  \centering
\includegraphics[width=0.95\columnwidth]{ 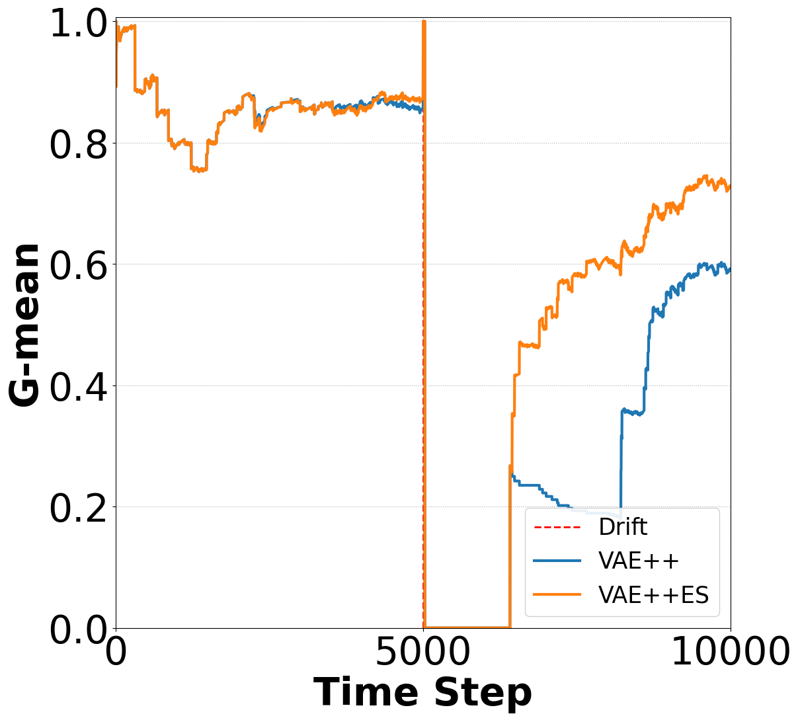}
  \caption{Fraud}
  \label{fig:fraud_es}
 \end{subfigure}%
\caption{Comparison of VAE++ with VAE++ES in nonstationary environments.}
\label{fig:compare_es}
\end{figure}

\begin{figure}[!h]
  \begin{subfigure}{.5\columnwidth}
  \centering
\includegraphics[width=0.95\columnwidth]{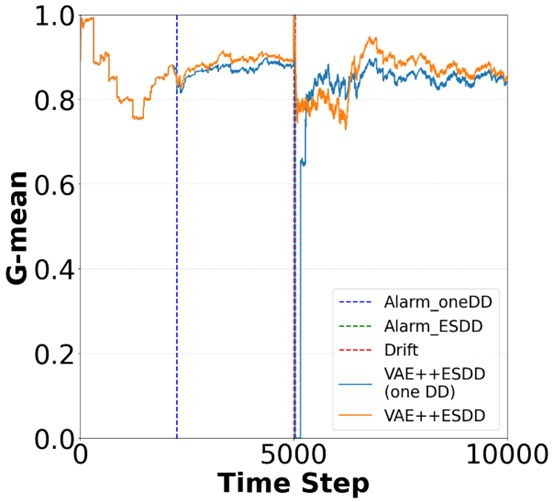}
  \caption{Fraud}
  \label{fig:fraud_es_dd}
 \end{subfigure}%
  \begin{subfigure}{.5\columnwidth}
  \centering
\includegraphics[width=0.935\columnwidth]{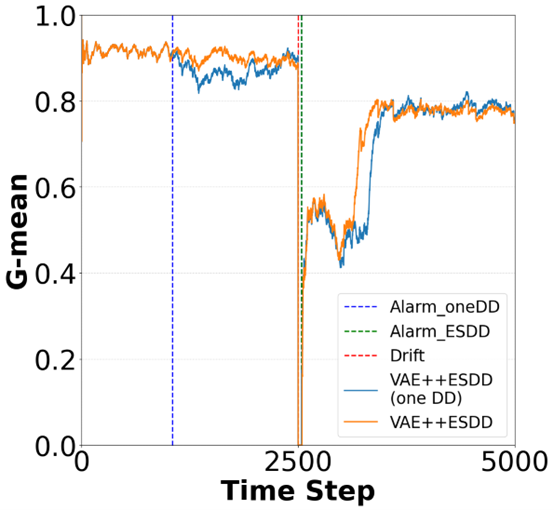} 
  \caption{MNIST-23}
\label{fig:mnist23_es_dd}
 \end{subfigure}%
\caption{Comparison of VAE++ESDD (one DD) with VAE++ESDD in nonstationary environments.}
\label{fig:compare_es_dd}
\end{figure}

\begin{figure}[!h]
\centering

\begin{subfigure}{.5\linewidth}
  \centering
  \includegraphics[width=0.95\linewidth]{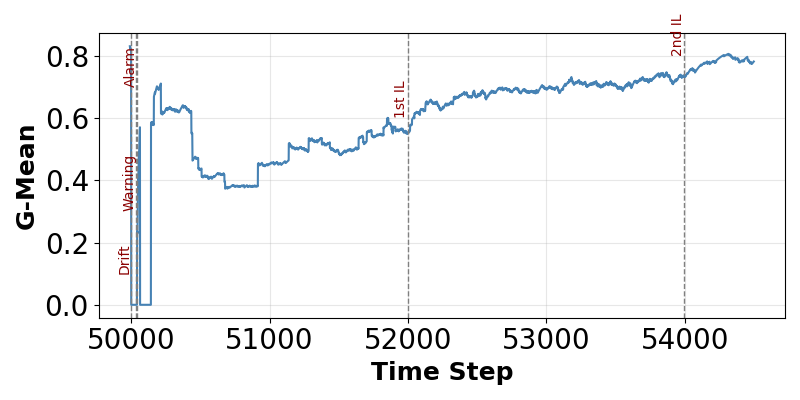}
  \caption{Forest}
    \label{fig:forest_big}
\end{subfigure}%
\begin{subfigure}{.5\linewidth}
  \centering
  \includegraphics[width=0.95\linewidth]{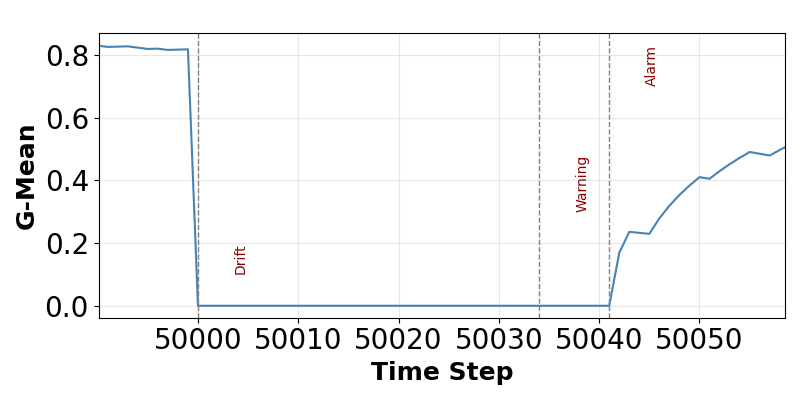}
  \caption{Forest (zoomed)}
        \label{fig:forest_small}
\end{subfigure}

\begin{subfigure}{.5\linewidth}
  \centering
  \includegraphics[width=0.95\linewidth]{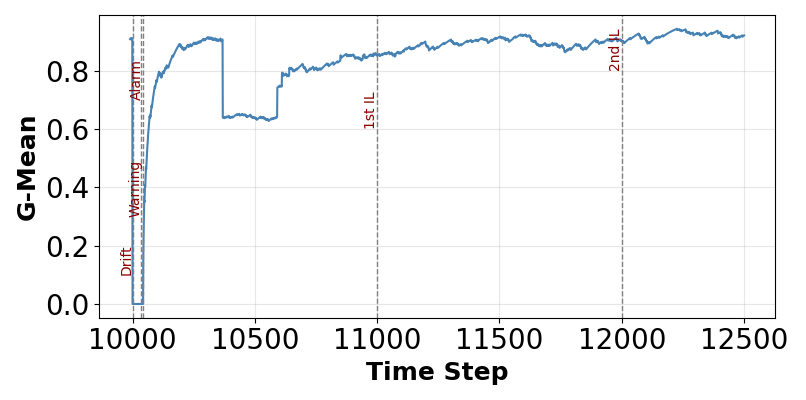}
  \caption{Sea}
      \label{fig:sea_big}
\end{subfigure}%
\begin{subfigure}{.5\linewidth}
  \centering
  \includegraphics[width=0.97\linewidth]{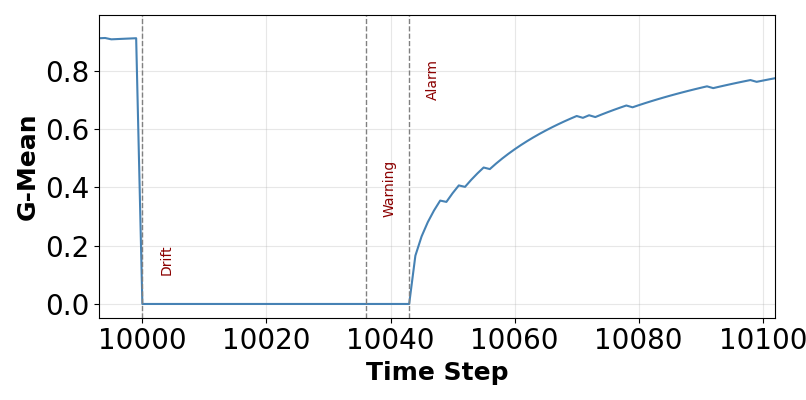}
  \caption{Sea (zoomed)}
      \label{fig:sea_small}
\end{subfigure}

\begin{subfigure}{.5\linewidth}
  \centering
  \includegraphics[width=0.95\linewidth]{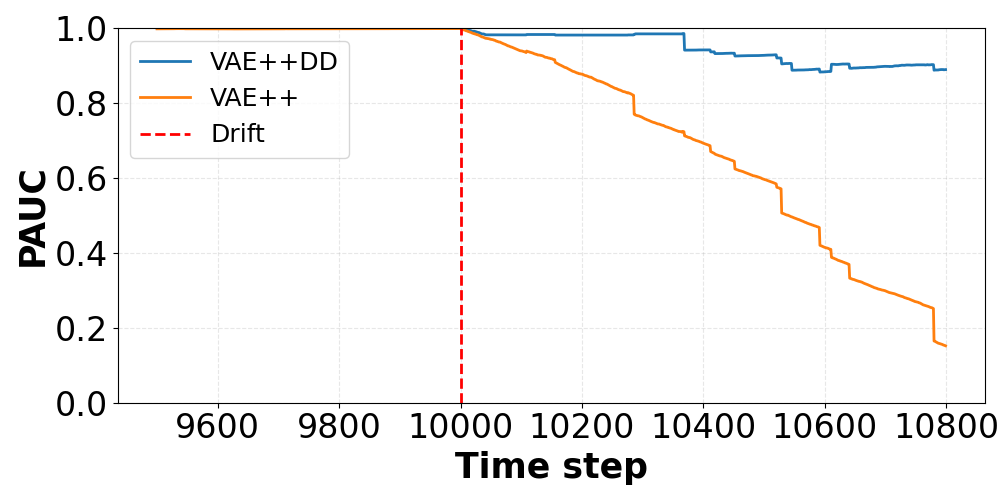}
  \caption{PAUC comparison (Sea)}
  \label{fig:pauc_sea}
\end{subfigure}%
\begin{subfigure}{.5\linewidth}
  \centering
  \includegraphics[width=0.95\linewidth]{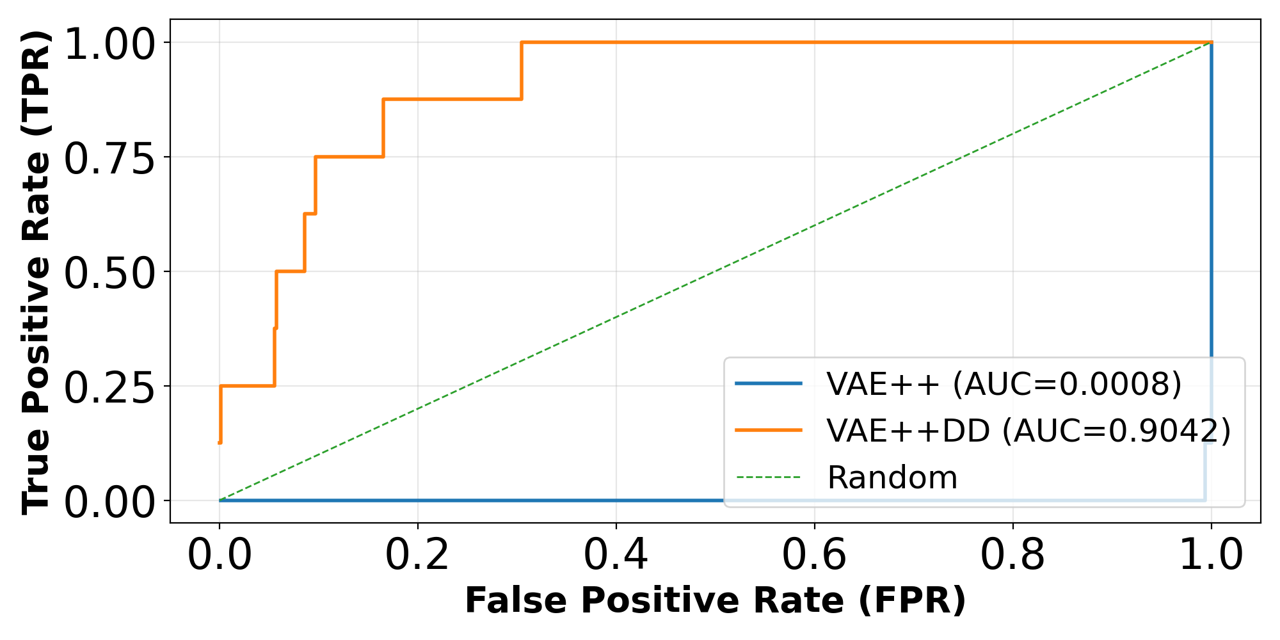}
  \caption{ROC comparison (Sea)}
  \label{fig:roc_sea}
\end{subfigure}

\caption{Model reset Performance.}
\label{fig:after_drift}
\end{figure}

\begin{figure}[!t]
\begin{subfigure}{.5\columnwidth}
  \centering
  \includegraphics[width=1\columnwidth]{ 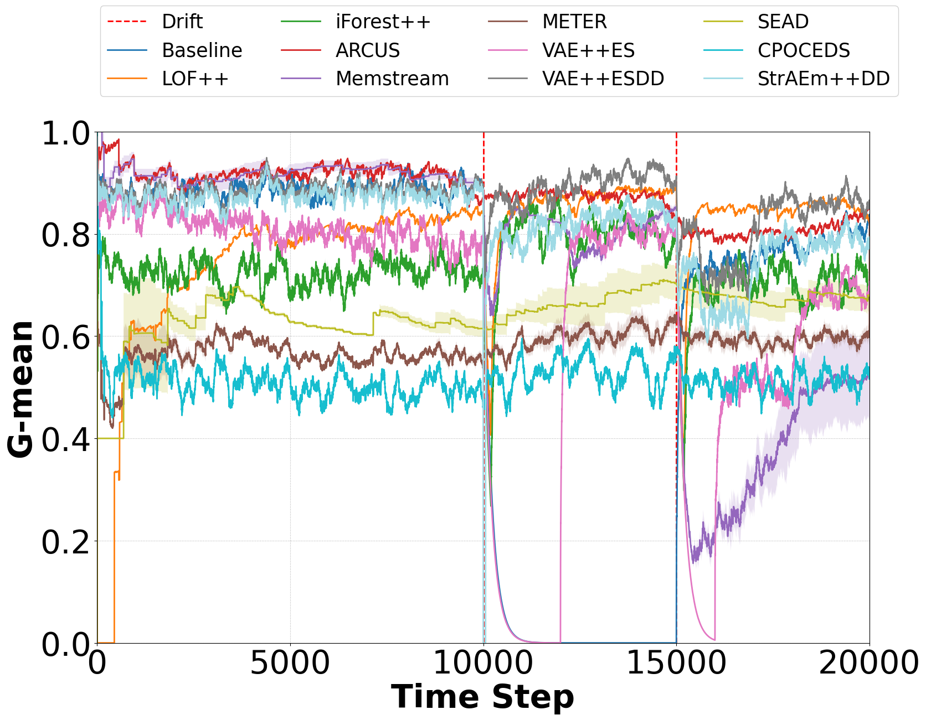}
  \caption{Sea}
  \label{fig:sea_recu}
\end{subfigure}%
\begin{subfigure}{.5\columnwidth}
  \centering
  \includegraphics[width=1\columnwidth]{ 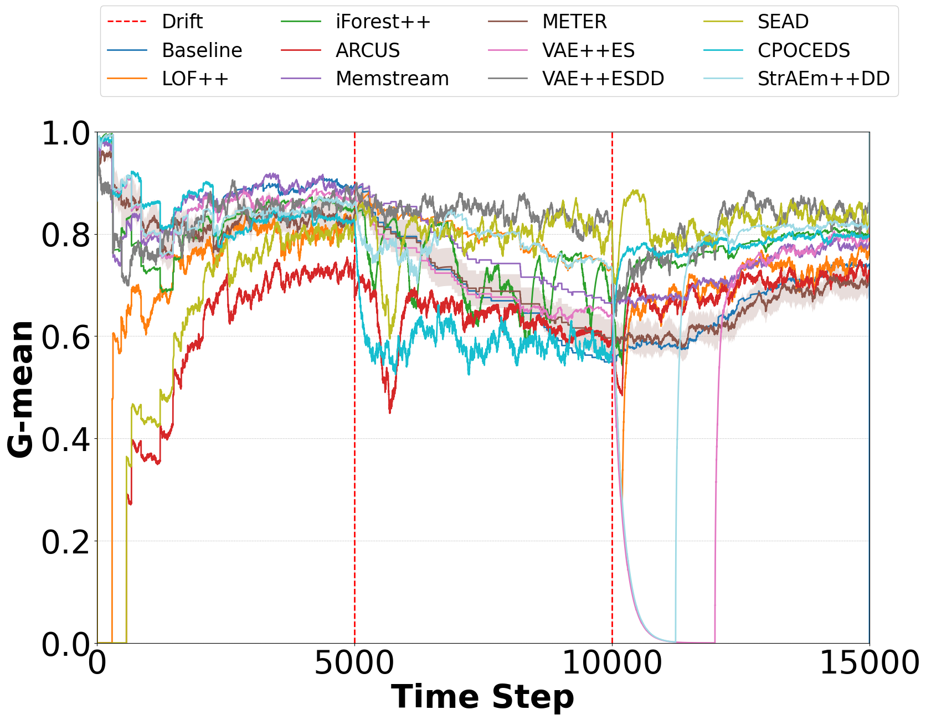}
  \caption{Fraud}
  \label{fig:fraud_recu}
\end{subfigure}
\caption{ G-mean of different methods in nonstationary environments.}
\label{fig:compare_recu}
\end{figure}

\subsubsection{Impact of ensemble learning}\label{sec:ensembleimpact}

This comparison highlights the benefits of ensemble learning. We first examine how the number of ensemble members (n) influences computational cost, resource usage, and performance. We then separately analyze the contribution of the predictor ensemble and the contribution of the drift-detector ensemble.

Table~\ref{tab:comp_resource} summarizes computational resources on the low-dimensional Sine and high-dimensional MNIST\_23 datasets. Memory usage scales linearly with the ensemble size $n$ and remains manageable at $n=10$, yielding about a 13\% performance gain on both datasets. The ensemble size has little effect on \(T_{\text{stream}}\) because fixed computation dominates the per-step runtime. In contrast, \(T_{\text{incr}}\) and \(T_{\text{drift}}\) increase with \(n\), reflecting the extra cost of updating and retraining multiple models, especially after drift detection.

Since incremental updates and retraining occur infrequently, the overall runtime is dominated by the small \(T_{\text{stream}}\), enabling near–real-time operation under normal conditions. The latency introduced during incremental learning or drift-triggered retraining can be mitigated through parallel execution, as ensemble members are independent. All experiments were run on a CPU, suggesting that GPU or distributed implementations would further improve efficiency.

\textbf{Impact of ensemble of predictors.} In this experiment, we compare VAE++ and VAE++ES. VAE++ is a single-model approach that employs incremental learning without ensembling or explicit drift detection. VAE++ES is the ensemble version of VAE++ with multiple anomaly predictors to make decisions. We provide two examples in Fig.~\ref{fig:compare_es}. As observed, VAE++ES demonstrates superior performance over VAE++ in terms of G-mean, especially after drift. This improvement can be explained by the multi-window incremental learning mechanism of the ensemble. Since the predictors are trained with different window sizes, some models perform incremental updates earlier after the drift occurs and thus adapt faster to the new data distribution. As a result, the ensemble as a whole exhibits improved detection performance during and after drift. In addition, with $P_{thre}=1$, an instance is classified as anomalous if any predictor flags it as such, which increases the overall sensitivity to new anomalies and consequently improves the G-mean.

\textbf{Impact of ensemble of drift detectors.} In this experiment, we compare VAE++ESDD (one DD) and VAE++ESDD. VAE++ESDD (one DD) is a variation of VAE++ESDD. In VAE++ESDD (one DD), we adopt an ensemble of predictors but with only one drift detector. In this experiment, the number of predictors is fixed for both methods and we only examine the effect of drift detector ensemble. 

Fig.~\ref{fig:compare_es_dd} compares the G-mean performance of the two methods, with alarm times indicated. For both datasets, around the drift point (red dashed line), the curves for “Alarm\_oneDD,” “Alarm\_ESDD,” and “Drift” nearly coincide, indicating that both methods successfully detected the actual drift. In contrast, VAE++ESDD (one DD), one false alarm is observed in each dataset, followed by a noticeable decrease in G-mean compared to VAE++ESDD. This degradation occurs because the ensemble predictors are reset after the false alarm and retrained using only instances within $mov_{ESwarn}$, while the previous predictors had already been trained on more data through several incremental learning updates. Consequently, the performance after the reset is lower. This highlights the importance of accurate drift detection for stable anomaly detection. VAE++ESDD employs a voting strategy that triggers an alarm only when more than $D_{thre}$ ensemble members detect a drift, thereby reducing false alarms from individual detectors.

\subsubsection{Model reset performance}
In this section, we focus on the model performance immediately after the model reset triggered by drift detection. For clarity, from Fig.~\ref{fig:forest_big} to Fig.~\ref{fig:sea_small}, the actual drift time, the warning time, the alarm time, and the two subsequent incremental updates are marked on the figures. For each dataset, we report the post-drift G-mean performance along with a zoomed view to better illustrate the short-term behavior right after drift. As shown in Fig.~\ref{fig:forest_big} and Fig.~\ref{fig:sea_big}, the model gradually recovers performance after each incremental update. After two updates, the performance returns to a level comparable to that before the drift. By inspecting the zoomed plots in Fig.~\ref{fig:forest_small} and Fig.~\ref{fig:sea_small}, we observe that the performance drops to zero between the drift and the alarm, since the model initially misclassifies the drifted data as anomalies. Once the warning is raised, instances begin to accumulate in $mov_{\text{ESwarn}}$. Approximately ten timesteps later, the alarm is triggered and a new ensemble is created and trained. Immediately after the alarm, the performance begins to recover, demonstrating that the warning buffer effectively mitigates degradation after the drift.


To further illustrate the effect of drift detection on anomaly detection under non-stationary conditions, Fig.~\ref{fig:pauc_sea} and Fig.~\ref{fig:roc_sea} compare the PAUC and PROC performance of VAE++ and VAE++DD on the Sea dataset. Before the occurrence of drift, both methods exhibit comparable PAUC values, indicating similar discriminative capability under stationary conditions. After drift, although a mild performance degradation is observed, VAE++DD maintains a consistently high PAUC, whereas the PAUC of VAE++ rapidly deteriorates. This behavior arises because concept drift increases the reconstruction errors of drifted normal samples, leading to substantial overlap between normal and anomalous score distributions when drift detection is absent. As a result, VAE++ increasingly misclassifies drifted normal samples as anomalies, causing a progressive loss of discriminative power.

To avoid the bias introduced by aggregating predictions over the entire data stream, the PROC analysis in Fig.~\ref{fig:roc_sea} is conducted, focusing exclusively on the first 1000 samples following the detected drift. Within this post-drift interval, VAE++ exhibits an AUC close to random performance, indicating that its anomaly ranking capability collapses after the concept shift. In contrast, VAE++DD achieves a substantially higher AUC, demonstrating that explicit drift detection and timely model adaptation effectively suppress drift-induced false positives and preserve robust ranking performance under evolving data distributions.

\begin{table*}[!t]
\caption{Average G-mean on the Datasets (The best-performing is shown in bold black, second-best in bold blue)}\label{tab:compare_gmean}
\begin{adjustbox}{width=1.0\textwidth}
\begin{tabular}{|c|cccc|cccccc|}
\hline
           & \multicolumn{4}{c|}{Synthetic datasets}                                                                                                                                                                                                                  & \multicolumn{6}{c|}{Real-world datasets}                                                                                                                                                                                                                                                                                                                                                         \\ \hline
Algorithm  & \multicolumn{1}{c|}{Sea}                                          & \multicolumn{1}{c|}{Sine}                                         & \multicolumn{1}{c|}{Circle}                                       & Vib                                          & \multicolumn{1}{c|}{Fraud}                                        & \multicolumn{1}{c|}{MNIST-01}                                     & \multicolumn{1}{c|}{MNIST-23}                                     & \multicolumn{1}{c|}{MNIST-multi}                                  & \multicolumn{1}{c|}{Forest}                                       & Arrhy                                        \\ \hline
Baseline   & \multicolumn{1}{c|}{0.640(0.000)}                                 & \multicolumn{1}{c|}{0.500(0.000)}                                 & \multicolumn{1}{c|}{0.467(0.000)}                                 & 0.884(0.000)                                 & \multicolumn{1}{c|}{0.741(0.000)}                                 & \multicolumn{1}{c|}{0.640(0.000)}                                 & \multicolumn{1}{c|}{0.490(0.000)}                                 & \multicolumn{1}{c|}{0.636(0.000)}                                 & \multicolumn{1}{c|}{0.308(0.000)}                                 & 0.836(0.000)                                 \\ \hline
iForest++  & \multicolumn{1}{c|}{0.743(0.000)}                                 & \multicolumn{1}{c|}{0.667(0.004)}                                 & \multicolumn{1}{c|}{0.630(0.010)}                                 & 0.930(0.002)                                 & \multicolumn{1}{c|}{0.787(0.003)}                                 & \multicolumn{1}{c|}{\textbf{0.747(0.006)}}                        & \multicolumn{1}{c|}{{\color[HTML]{00009B} \textbf{0.742(0.007)}}} & \multicolumn{1}{c|}{0.503(0.007)}                                 & \multicolumn{1}{c|}{0.472(0.007)}                                 & 0.843(0.002)                                 \\ \hline
LOF++      & \multicolumn{1}{c|}{0.771(0.000)}                                 & \multicolumn{1}{c|}{0.672(0.000)}                                 & \multicolumn{1}{c|}{{\color[HTML]{00009B} \textbf{0.706(0.000)}}} & \textbf{0.990(0.000)}                        & \multicolumn{1}{c|}{0.749(0.000)}                                 & \multicolumn{1}{c|}{0.465(0.000)}                                 & \multicolumn{1}{c|}{0.335(0.000)}                                 & \multicolumn{1}{c|}{0.445(0.000)}                                 & \multicolumn{1}{c|}{0.566(0.000)}                                 & 0.842(0.000)                                 \\ \hline
ARCUS      & \multicolumn{1}{c|}{\textbf{0.880(0.000)}}                        & \multicolumn{1}{c|}{0.763(0.000)} & \multicolumn{1}{c|}{0.579(0.000)}                                 & 0.940(0.000)                                 & \multicolumn{1}{c|}{0.619(0.000)}                                 & \multicolumn{1}{c|}{0.567(0.000)}                                 & \multicolumn{1}{c|}{0.663(0.000)}                                 & \multicolumn{1}{c|}{0.578(0.000)}                                 & \multicolumn{1}{c|}{0.311(0.000)}                                 & 0.657(0.000)                                 \\ \hline
Memstream  & \multicolumn{1}{c|}{0.739(0.014)}                                 & \multicolumn{1}{c|}{\color[HTML]{000000} \textbf{0.808(0.013)}}                                 & \multicolumn{1}{c|}{0.691(0.001)}                                 & 0.938(0.000)                                 & \multicolumn{1}{c|}{{\color[HTML]{000000} 0.786(0.001)}} & \multicolumn{1}{c|}{0.624(0.000)}                                 & \multicolumn{1}{c|}{0.454(0.000)}                                 & \multicolumn{1}{c|}{0.382(0.000)}                                 & \multicolumn{1}{c|}{0.226(0.000)}                                 & 0.802(0.000)                                 \\ \hline
METER      & \multicolumn{1}{c|}{0.580(0.014)}                                 & \multicolumn{1}{c|}{0.451(0.110)}                                 & \multicolumn{1}{c|}{0.556(0.000)}                                 & 0.734(0.179)                                 & \multicolumn{1}{c|}{0.729(0.030)}                                 & \multicolumn{1}{c|}{0.202(0.044)}                                 & \multicolumn{1}{c|}{0.606(0.100)}                                 & \multicolumn{1}{c|}{0.274(0.068)}                                 & \multicolumn{1}{c|}{0.474(0.043)}                                 & 0.816(0.025)                                 \\ \hline
SEAD       & \multicolumn{1}{c|}{0.681(0.013)}                                 & \multicolumn{1}{c|}{0.713(0.014)}                                 & \multicolumn{1}{c|}{0.391(0.022)}                                 & 0.930(0.052)                                 & \multicolumn{1}{c|}{0.744(0.009)}                                 & \multicolumn{1}{c|}{0.367(0.030)}                                 & \multicolumn{1}{c|}{0.592(0.012)}                                 & \multicolumn{1}{c|}{0.354(0.018)}                                 & \multicolumn{1}{c|}{{\color[HTML]{00009B} \textbf{0.724(0.015)}}}                                 & 0.719(0.012)                                 \\ \hline
CPOCEDS    & \multicolumn{1}{c|}{0.516(0.000)}                                 & \multicolumn{1}{c|}{0.544(0.000)}                                 & \multicolumn{1}{c|}{0.557(0.000)}                                 & 0.869(0.000)                                 & \multicolumn{1}{c|}{0.704(0.000)}                                 & \multicolumn{1}{c|}{0.513(0.000)}                                 & \multicolumn{1}{c|}{0.481(0.000)}                                 & \multicolumn{1}{c|}{0.485(0.000)}                                 & \multicolumn{1}{c|}{0.290(0.000)}                                 & 0.743(0.000)                                 \\ \hline
StrAEm++DD & \multicolumn{1}{c|}{0.790(0.000)}                                 & \multicolumn{1}{c|}{0.704(0.000)}                                 & \multicolumn{1}{c|}{0.523(0.000)}                                 & 0.924(0.000)                                 & \multicolumn{1}{c|}{0.797(0.000)}                                 & \multicolumn{1}{c|}{0.726(0.000)}                                 & \multicolumn{1}{c|}{0.701(0.000)}                                 & \multicolumn{1}{c|}{{\color[HTML]{00009B} \textbf{0.736(0.000)}}} & \multicolumn{1}{c|}{0.627(0.000)}                                 & 0.801(0.000)                                 \\ \hline
VAE++ES    & \multicolumn{1}{c|}{0.655(0.000)}                                 & \multicolumn{1}{c|}{0.680(0.000)}                                 & \multicolumn{1}{c|}{0.552(0.000)}                                 & 0.910(0.000)                                 & \multicolumn{1}{c|}{{\color[HTML]{00009B} \textbf{0.814(0.000)}}}                                 & \multicolumn{1}{c|}{0.719(0.000)}                                 & \multicolumn{1}{c|}{0.715(0.000)}                                 & \multicolumn{1}{c|}{0.680(0.000)}                                 & \multicolumn{1}{c|}{0.677(0.000)} & {\color[HTML]{00009B} \textbf{0.844(0.000)}} \\ \hline
VAE++ESDD  & \multicolumn{1}{c|}{{\color[HTML]{00009B} \textbf{0.868(0.000)}}} & \multicolumn{1}{c|}{\color[HTML]{00009B}\textbf{0.791(0.000)}}                        & \multicolumn{1}{c|}{\textbf{0.760(0.000)}}                        & {\color[HTML]{00009B} \textbf{0.945(0.000)}} & \multicolumn{1}{c|}{\textbf{0.821(0.000)}}                        & \multicolumn{1}{c|}{{\color[HTML]{00009B} \textbf{0.744(0.000)}}} & \multicolumn{1}{c|}{\textbf{0.799(0.000)}}                        & \multicolumn{1}{c|}{\textbf{0.825(0.000)}}                        & \multicolumn{1}{c|}{\textbf{0.817(0.000)}}                        & \textbf{0.845(0.000)}                        \\ \hline
\end{tabular}
\end{adjustbox}
\end{table*}

\begin{table*}[!t]
\centering
\caption{Average Recall on the Datasets (The best-performing is shown in bold black, second-best in bold blue)}
\label{tab:compare_recall}
\begin{adjustbox}{width=1.0\textwidth}
\begin{tabular}{|c|cccc|cccccc|}
\hline
{\color[HTML]{000000} }           & \multicolumn{4}{c|}{{\color[HTML]{000000} Synthetic datasets}}                                                                                                                                                                                           & \multicolumn{6}{c|}{{\color[HTML]{000000} Real-world datasets}}                                                                                                                                                                                                                                                                                                                                  \\ \hline
{\color[HTML]{000000} Algorithm}  & \multicolumn{1}{c|}{{\color[HTML]{000000} Sea}}                   & \multicolumn{1}{c|}{{\color[HTML]{000000} Sine}}                  & \multicolumn{1}{c|}{{\color[HTML]{000000} Circle}}                & {\color[HTML]{000000} Vib}                   & \multicolumn{1}{c|}{{\color[HTML]{000000} Fraud}}                 & \multicolumn{1}{c|}{{\color[HTML]{000000} MNIST-01}}              & \multicolumn{1}{c|}{{\color[HTML]{000000} MNIST-23}}              & \multicolumn{1}{c|}{{\color[HTML]{000000} MNIST-multi}}           & \multicolumn{1}{c|}{{\color[HTML]{000000} Forest}}                & {\color[HTML]{000000} Arrhy}                 \\ \hline
{\color[HTML]{000000} Baseline}   & \multicolumn{1}{c|}{{\color[HTML]{000000} 0.884(0.000)}}          & \multicolumn{1}{c|}{{\color[HTML]{000000} 0.606(0.000)}}          & \multicolumn{1}{c|}{{\color[HTML]{000000} 0.701(0.000)}}          & {\color[HTML]{000000} 0.899(0.000)}          & \multicolumn{1}{c|}{{\color[HTML]{000000} 0.573(0.000)}}          & \multicolumn{1}{c|}{{\color[HTML]{000000} 0.792(0.000)}}          & \multicolumn{1}{c|}{{\color[HTML]{000000} 0.771(0.000)}}          & \multicolumn{1}{c|}{{\color[HTML]{000000} 0.712(0.000)}}          & \multicolumn{1}{c|}{{\color[HTML]{000000} 0.660(0.000)}}          & {\color[HTML]{000000} 0.697(0.000)}          \\ \hline
{\color[HTML]{000000} iForest++}  & \multicolumn{1}{c|}{{\color[HTML]{000000} \textbf{0.999(0.004)}}} & \multicolumn{1}{c|}{{\color[HTML]{000000} \textbf{1.000(0.000)}}} & \multicolumn{1}{c|}{{\color[HTML]{000000} \textbf{0.928(0.009)}}} & {\color[HTML]{000000} \textbf{1.000(0.000)}} & \multicolumn{1}{c|}{{\color[HTML]{000000} 0.669(0.004)}}          & \multicolumn{1}{c|}{{\color[HTML]{000000} 0.880(0.007)}}          & \multicolumn{1}{c|}{{\color[HTML]{000000} 0.939(0.003)}}          & \multicolumn{1}{c|}{{\color[HTML]{000000} 0.865(0.007)}}          & \multicolumn{1}{c|}{{\color[HTML]{000000} 0.243(0.005)}}          & {\color[HTML]{000000} 0.701(0.001)}          \\ \hline
{\color[HTML]{000000} LOF++}      & \multicolumn{1}{c|}{{\color[HTML]{000000} 0.657(0.000)}}          & \multicolumn{1}{c|}{{\color[HTML]{00009B} \textbf{0.947(0.000)}}} & \multicolumn{1}{c|}{{\color[HTML]{000000} 0.429(0.000)}}          & {\color[HTML]{000000} 0.889(0.000)}          & \multicolumn{1}{c|}{{\color[HTML]{000000} 0.681(0.000)}}          & \multicolumn{1}{c|}{{\color[HTML]{000000} 0.030(0.000)}}          & \multicolumn{1}{c|}{{\color[HTML]{000000} 0.270(0.000)}}          & \multicolumn{1}{c|}{{\color[HTML]{000000} 0.050(0.000)}}          & \multicolumn{1}{c|}{{\color[HTML]{000000} 0.656(0.000)}}          & {\color[HTML]{000000} 0.677(0.000)}          \\ \hline
{\color[HTML]{000000} ARCUS}      & \multicolumn{1}{c|}{{\color[HTML]{000000} 0.826(0.006)}}          & \multicolumn{1}{c|}{{\color[HTML]{000000} 0.645(0.001)}}          & \multicolumn{1}{c|}{{\color[HTML]{000000} 0.396(0.001)}}          & {\color[HTML]{000000} \textbf{1.000(0.000)}} & \multicolumn{1}{c|}{{\color[HTML]{000000} 0.620(0.000)}} & \multicolumn{1}{c|}{{\color[HTML]{000000} 0.410(0.002)}}          & \multicolumn{1}{c|}{{\color[HTML]{000000} 0.480(0.000)}}          & \multicolumn{1}{c|}{{\color[HTML]{000000} 0.410(0.001)}}          & \multicolumn{1}{c|}{{\color[HTML]{000000} 0.193(0.000)}}          & {\color[HTML]{000000} 0.641(0.000)}          \\ \hline
{\color[HTML]{000000} Memstream}  & \multicolumn{1}{c|}{{\color[HTML]{000000} 0.781(0.001)}}          & \multicolumn{1}{c|}{{\color[HTML]{000000} 0.676(0.001)}}          & \multicolumn{1}{c|}{{\color[HTML]{000000} 0.422(0.001)}}          & {\color[HTML]{000000} \textbf{1.000(0.000)}} & \multicolumn{1}{c|}{{\color[HTML]{000000} 0.715(0.000)}}          & \multicolumn{1}{c|}{{\color[HTML]{000000} 0.420(0.001)}}          & \multicolumn{1}{c|}{{\color[HTML]{000000} 0.372(0.000)}}          & \multicolumn{1}{c|}{{\color[HTML]{000000} 0.173(0.000)}}          & \multicolumn{1}{c|}{{\color[HTML]{000000} 0.071(0.000)}}          & {\color[HTML]{000000} 0.722(0.000)}          \\ \hline
{\color[HTML]{000000} METER}      & \multicolumn{1}{c|}{{\color[HTML]{000000} 0.000(0.000)}}          & \multicolumn{1}{c|}{{\color[HTML]{000000} 0.396(0.000)}}          & \multicolumn{1}{c|}{{\color[HTML]{000000} 0.001(0.000)}}          & {\color[HTML]{000000} \textbf{1.000(0.000)}} & \multicolumn{1}{c|}{{\color[HTML]{000000} 0.602(0.001)}}          & \multicolumn{1}{c|}{{\color[HTML]{000000} 0.055(0.001)}}          & \multicolumn{1}{c|}{{\color[HTML]{000000} \textbf{0.986(0.001)}}} & \multicolumn{1}{c|}{{\color[HTML]{000000} 0.086(0.000)}}          & \multicolumn{1}{c|}{{\color[HTML]{000000} 0.476(0.001)}}          & {\color[HTML]{000000} 0.916(0.001)}          \\ \hline
{\color[HTML]{000000} SEAD}       & \multicolumn{1}{c|}{{\color[HTML]{000000} 0.701(0.018)}}          & \multicolumn{1}{c|}{{\color[HTML]{000000} 0.658(0.017)}}          & \multicolumn{1}{c|}{{\color[HTML]{000000} 0.288(0.009)}}          & {\color[HTML]{000000} 0.997(0.000)}          & \multicolumn{1}{c|}{{\color[HTML]{000000} 0.670(0.004)}}          & \multicolumn{1}{c|}{{\color[HTML]{000000} 0.221(0.028)}}          & \multicolumn{1}{c|}{{\color[HTML]{000000} 0.490(0.014)}}          & \multicolumn{1}{c|}{{\color[HTML]{000000} 0.171(0.013)}}          & \multicolumn{1}{c|}{{\color[HTML]{000000} 0.747(0.021)}}          & {\color[HTML]{000000} 0.694(0.005)}          \\ \hline
{\color[HTML]{000000} CPOCEDS}    & \multicolumn{1}{c|}{{\color[HTML]{000000} 0.529(0.000)}}          & \multicolumn{1}{c|}{{\color[HTML]{000000} 0.545(0.000)}}          & \multicolumn{1}{c|}{{\color[HTML]{000000} 0.513(0.000)}}          & {\color[HTML]{000000} 0.456(0.000)}          & \multicolumn{1}{c|}{{\color[HTML]{000000} 0.610(0.000)}}          & \multicolumn{1}{c|}{{\color[HTML]{000000} 0.585(0.000)}}          & \multicolumn{1}{c|}{{\color[HTML]{000000} 0.160(0.000)}}          & \multicolumn{1}{c|}{{\color[HTML]{000000} 0.481(0.000)}}          & \multicolumn{1}{c|}{{\color[HTML]{000000} 0.088(0.000)}}          & {\color[HTML]{000000} 0.360(0.000)}          \\ \hline
{\color[HTML]{000000} StrAEm++DD} & \multicolumn{1}{c|}{{\color[HTML]{000000} 0.954(0.000)}}          & \multicolumn{1}{c|}{{\color[HTML]{000000} 0.745(0.000)}}          & \multicolumn{1}{c|}{{\color[HTML]{000000} 0.787(0.000)}}          & {\color[HTML]{000000} 0.991(0.000)}          & \multicolumn{1}{c|}{{\color[HTML]{000000} 0.733(0.000)}}          & \multicolumn{1}{c|}{{\color[HTML]{00009B} \textbf{0.910(0.000)}}} & \multicolumn{1}{c|}{{\color[HTML]{000000} 0.948(0.000)}}          & \multicolumn{1}{c|}{{\color[HTML]{00009B} \textbf{0.910(0.000)}}} & \multicolumn{1}{c|}{{\color[HTML]{000000} 0.508(0.000)}}          & {\color[HTML]{00009B} \textbf{0.851(0.000)}} \\ \hline
{\color[HTML]{000000} VAE++ES}    & \multicolumn{1}{c|}{{\color[HTML]{000000} 0.904(0.000)}}          & \multicolumn{1}{c|}{{\color[HTML]{000000} 0.750(0.000)}}          & \multicolumn{1}{c|}{{\color[HTML]{000000} 0.809(0.000)}}          & {\color[HTML]{000000} 0.920(0.000)}          & \multicolumn{1}{c|}{{\color[HTML]{00009B} \textbf{0.750(0.000)}}}          & \multicolumn{1}{c|}{{\color[HTML]{00009B} \textbf{0.910(0.000)}}} & \multicolumn{1}{c|}{{\color[HTML]{000000} 0.910(0.000)}}          & \multicolumn{1}{c|}{{\color[HTML]{000000} 0.909(0.000)}}          & \multicolumn{1}{c|}{{\color[HTML]{00009B} \textbf{0.820(0.000)}}} & {\color[HTML]{000000} 0.811(0.000)}          \\ \hline
{\color[HTML]{000000} VAE++ESDD}  & \multicolumn{1}{c|}{{\color[HTML]{00009B} \textbf{0.984(0.000)}}} & \multicolumn{1}{c|}{{\color[HTML]{000000} 0.766(0.000)}}          & \multicolumn{1}{c|}{{\color[HTML]{00009B} \textbf{0.845(0.000)}}} & {\color[HTML]{00009B} \textbf{0.998(0.000)}} & \multicolumn{1}{c|}{{\color[HTML]{000000} \textbf{0.755(0.000)}}} & \multicolumn{1}{c|}{{\color[HTML]{000000} \textbf{0.925(0.000)}}} & \multicolumn{1}{c|}{{\color[HTML]{00009B} \textbf{0.962(0.000)}}} & \multicolumn{1}{c|}{{\color[HTML]{000000} \textbf{0.924(0.000)}}} & \multicolumn{1}{c|}{{\color[HTML]{000000} \textbf{0.857(0.000)}}} & {\color[HTML]{000000} \textbf{0.861(0.000)}} \\ \hline
\end{tabular}
\end{adjustbox}
\end{table*}

\begin{table*}[!t]
\caption{Average Specificity on the Datasets (The best-performing is shown in bold black, second-best in bold blue)}
\label{tab:compare_speci}
\begin{adjustbox}{width=1.0\textwidth}
\begin{tabular}{|c|cccc|cccccc|}
\hline
{\color[HTML]{000000} }           & \multicolumn{4}{c|}{{\color[HTML]{000000} Synthetic datasets}}                                                                                                                                                                                           & \multicolumn{6}{c|}{{\color[HTML]{000000} Real-world datasets}}                                                                                                                                                                                                                                                                                                                                  \\ \hline
{\color[HTML]{000000} Algorithm}  & \multicolumn{1}{c|}{{\color[HTML]{000000} Sea}}                   & \multicolumn{1}{c|}{{\color[HTML]{000000} Sine}}                  & \multicolumn{1}{c|}{{\color[HTML]{000000} Circle}}                & {\color[HTML]{000000} Vib}                   & \multicolumn{1}{c|}{{\color[HTML]{000000} Fraud}}                 & \multicolumn{1}{c|}{{\color[HTML]{000000} MNIST-01}}              & \multicolumn{1}{c|}{{\color[HTML]{000000} MNIST-23}}              & \multicolumn{1}{c|}{{\color[HTML]{000000} MNIST-multi}}           & \multicolumn{1}{c|}{{\color[HTML]{000000} Forest}}                & {\color[HTML]{000000} Arrhy}                 \\ \hline
{\color[HTML]{000000} Baseline}   & \multicolumn{1}{c|}{{\color[HTML]{000000} 0.588(0.000)}}          & \multicolumn{1}{c|}{{\color[HTML]{000000} 0.484(0.000)}}          & \multicolumn{1}{c|}{{\color[HTML]{000000} 0.403(0.000)}}          & {\color[HTML]{000000} 0.803(0.000)}          & \multicolumn{1}{c|}{{\color[HTML]{00009B} \textbf{0.987(0.000)}}}          & \multicolumn{1}{c|}{{\color[HTML]{000000} 0.592(0.000)}}          & \multicolumn{1}{c|}{{\color[HTML]{000000} 0.443(0.000)}}          & \multicolumn{1}{c|}{{\color[HTML]{000000} 0.567(0.000)}}          & \multicolumn{1}{c|}{{\color[HTML]{000000} 0.146(0.000)}}          & {\color[HTML]{000000} 0.910(0.000)}          \\ \hline
{\color[HTML]{000000} iForest++}  & \multicolumn{1}{c|}{{\color[HTML]{000000} 0.542(0.008)}}          & \multicolumn{1}{c|}{{\color[HTML]{000000} 0.585(0.007)}}          & \multicolumn{1}{c|}{{\color[HTML]{000000} 0.506(0.009)}}          & {\color[HTML]{000000} 0.890(0.002)}          & \multicolumn{1}{c|}{{\color[HTML]{000000} 0.890(0.002)}}          & \multicolumn{1}{c|}{{\color[HTML]{000000} 0.819(0.006)}}           & \multicolumn{1}{c|}{{\color[HTML]{000000} 0.791(0.006)}}          & \multicolumn{1}{c|}{{\color[HTML]{000000} 0.818(0.006)}}          & \multicolumn{1}{c|}{{\color[HTML]{00009B} \textbf{0.972(0.002)}}}          & {\color[HTML]{000000} 0.907(0.002)}          \\ \hline
{\color[HTML]{000000} LOF++}      & \multicolumn{1}{c|}{{\color[HTML]{00009B} \textbf{0.944(0.000)}}} & \multicolumn{1}{c|}{{\color[HTML]{000000} \textbf{0.946(0.000)}}} & \multicolumn{1}{c|}{{\color[HTML]{00009B} \textbf{0.947(0.000)}}} & {\color[HTML]{000000} 0.885(0.000)}          & \multicolumn{1}{c|}{{\color[HTML]{000000} 0.885(0.000)}}          & \multicolumn{1}{c|}{{\color[HTML]{000000} \textbf{0.997(0.000)}}} & \multicolumn{1}{c|}{{\color[HTML]{000000} \textbf{0.993(0.000)}}} & \multicolumn{1}{c|}{{\color[HTML]{00009B} \textbf{0.996(0.000)}}} & \multicolumn{1}{c|}{{\color[HTML]{000000} 0.876(0.000)}}          & {\color[HTML]{00009B}\textbf{0.968(0.000)}}          \\ \hline
{\color[HTML]{000000} ARCUS}      & \multicolumn{1}{c|}{{\color[HTML]{000000} 0.942(0.000)}}          & \multicolumn{1}{c|}{{\color[HTML]{00009B} \textbf{0.917(0.000)}}} & \multicolumn{1}{c|}{{\color[HTML]{000000} 0.892(0.000)}}          & {\color[HTML]{000000} 0.888(0.000)}          & \multicolumn{1}{c|}{{\color[HTML]{000000} \textbf{0.998(0.000)}}} & \multicolumn{1}{c|}{{\color[HTML]{000000} 0.952(0.000)}}          & \multicolumn{1}{c|}{{\color[HTML]{000000} 0.947(0.000)}}          & \multicolumn{1}{c|}{{\color[HTML]{000000} 0.960(0.000)}}          & \multicolumn{1}{c|}{{\color[HTML]{000000} \textbf{0.989(0.000)}}} & {\color[HTML]{000000} \textbf{0.991(0.000)}} \\ \hline
{\color[HTML]{000000} Memstream}  & \multicolumn{1}{c|}{{\color[HTML]{000000} 0.653(0.003)}}          & \multicolumn{1}{c|}{{\color[HTML]{000000} 0.761(0.002)}}          & \multicolumn{1}{c|}{{\color[HTML]{000000} 0.781(0.002)}}          & {\color[HTML]{00009B} \textbf{0.950(0.000)}} & \multicolumn{1}{c|}{{\color[HTML]{000000} 0.954(0.003)}}          & \multicolumn{1}{c|}{{\color[HTML]{000000} 0.950(0.002)}}          & \multicolumn{1}{c|}{{\color[HTML]{00009B} \textbf{0.950(0.007)}}} & \multicolumn{1}{c|}{{\color[HTML]{000000} 0.949(0.000)}}          & \multicolumn{1}{c|}{{\color[HTML]{000000} \textbf{0.989(0.003)}}} & {\color[HTML]{000000} \textbf{0.991(0.000)}} \\ \hline
{\color[HTML]{000000} METER}      & \multicolumn{1}{c|}{{\color[HTML]{000000} \textbf{1.000(0.000)}}} & \multicolumn{1}{c|}{{\color[HTML]{000000} 0.522(0.000)}}          & \multicolumn{1}{c|}{{\color[HTML]{000000} \textbf{1.000(0.000)}}} & {\color[HTML]{000000} \textbf{0.994(0.001)}} & \multicolumn{1}{c|}{{\color[HTML]{000000} 0.914(0.001)}}          & \multicolumn{1}{c|}{{\color[HTML]{00009B} \textbf{0.995(0.000)}}} & \multicolumn{1}{c|}{{\color[HTML]{000000} 0.488(0.006)}}          & \multicolumn{1}{c|}{{\color[HTML]{000000} 0.081(0.006)}}          & \multicolumn{1}{c|}{{\color[HTML]{000000} 0.487(0.001)}}          & {\color[HTML]{000000} 0.846(0.000)}          \\ \hline
{\color[HTML]{000000} SEAD}       & \multicolumn{1}{c|}{{\color[HTML]{000000} 0.860(0.052)}}          & \multicolumn{1}{c|}{{\color[HTML]{000000} 0.810(0.058)}}          & \multicolumn{1}{c|}{{\color[HTML]{000000} 0.711(0.073)}}          & {\color[HTML]{000000} 0.902(0.051)}          & \multicolumn{1}{c|}{{\color[HTML]{000000} 0.862(0.051)}} & \multicolumn{1}{c|}{{\color[HTML]{000000} 0.722(0.056)}}          & \multicolumn{1}{c|}{{\color[HTML]{000000} 0.828(0.044)}}          & \multicolumn{1}{c|}{{\color[HTML]{000000} 0.860(0.065)}}          & \multicolumn{1}{c|}{{\color[HTML]{000000} 0.708(0.057)}} & {\color[HTML]{000000} 0.851(0.042)} \\ \hline
{\color[HTML]{000000} CPOCEDS}    & \multicolumn{1}{c|}{{\color[HTML]{000000} 0.502(0.000)}}          & \multicolumn{1}{c|}{{\color[HTML]{000000} 0.497(0.000)}}          & \multicolumn{1}{c|}{{\color[HTML]{000000} 0.595(0.000)}}          & {\color[HTML]{000000} 0.769(0.000)}          & \multicolumn{1}{c|}{{\color[HTML]{000000} 0.733(0.000)}}          & \multicolumn{1}{c|}{{\color[HTML]{000000} 0.499(0.000)}}          & \multicolumn{1}{c|}{{\color[HTML]{000000} 0.894(0.000)}}          & \multicolumn{1}{c|}{{\color[HTML]{000000} \textbf{0.998(0.000)}}} & \multicolumn{1}{c|}{{\color[HTML]{000000} 0.921(0.000)}}          & {\color[HTML]{000000} 0.747(0.000)}          \\ \hline
{\color[HTML]{000000} StrAEm++DD} & \multicolumn{1}{c|}{{\color[HTML]{000000} 0.740(0.000)}}          & \multicolumn{1}{c|}{{\color[HTML]{000000} 0.800(0.000)}}          & \multicolumn{1}{c|}{{\color[HTML]{000000} 0.436(0.000)}}          & {\color[HTML]{000000} 0.762(0.000)}          & \multicolumn{1}{c|}{{\color[HTML]{000000} 0.956(0.000)}}          & \multicolumn{1}{c|}{{\color[HTML]{000000} 0.650(0.000)}}          & \multicolumn{1}{c|}{{\color[HTML]{000000} 0.556(0.000)}}          & \multicolumn{1}{c|}{{\color[HTML]{000000} 0.612(0.000)}}          & \multicolumn{1}{c|}{{\color[HTML]{000000} 0.799(0.000)}}          & {\color[HTML]{000000} 0.801(0.000)}          \\ \hline
{\color[HTML]{000000} VAE++ES}    & \multicolumn{1}{c|}{{\color[HTML]{000000} 0.542(0.000)}}          & \multicolumn{1}{c|}{{\color[HTML]{000000} 0.719(0.000)}}          & \multicolumn{1}{c|}{{\color[HTML]{000000} 0.511(0.000)}}          & {\color[HTML]{000000} 0.757(0.000)}          & \multicolumn{1}{c|}{{\color[HTML]{000000} 0.976(0.000)}}          & \multicolumn{1}{c|}{{\color[HTML]{000000} 0.578(0.000)}}          & \multicolumn{1}{c|}{{\color[HTML]{000000} 0.567(0.000)}}          & \multicolumn{1}{c|}{{\color[HTML]{000000} 0.574(0.000)}}          & \multicolumn{1}{c|}{{\color[HTML]{000000} 0.688(0.000)}}          & {\color[HTML]{000000} 0.867(0.000)}          \\ \hline
{\color[HTML]{000000} VAE++ESDD}  & \multicolumn{1}{c|}{{\color[HTML]{000000} 0.768(0.000)}}          & \multicolumn{1}{c|}{{\color[HTML]{000000} 0.835(0.000)}}          & \multicolumn{1}{c|}{{\color[HTML]{000000} 0.810(0.000)}}          & {\color[HTML]{000000} 0.771(0.000)}          & \multicolumn{1}{c|}{{\color[HTML]{000000} 0.944(0.000)}}          & \multicolumn{1}{c|}{{\color[HTML]{000000} 0.670(0.000)}}          & \multicolumn{1}{c|}{{\color[HTML]{000000} 0.659(0.000)}}          & \multicolumn{1}{c|}{{\color[HTML]{000000} 0.670(0.000)}}          & \multicolumn{1}{c|}{{\color[HTML]{000000} 0.700(0.000)}}          & {\color[HTML]{000000} 0.886(0.000)}          \\ \hline
\end{tabular}
\end{adjustbox}
\end{table*}

\begin{table*}[!t]
\caption{Average PAUC on the Datasets (The best-performing is shown in bold black, second-best in bold blue)}
\label{tab:compare_pauc}
\begin{adjustbox}{width=1.0\textwidth}
\begin{tabular}{|c|cccc|cccccc|}
\hline
{\color[HTML]{000000} }           & \multicolumn{4}{c|}{{\color[HTML]{000000} Synthetic datasets}}                                                                                                                                                                                           & \multicolumn{6}{c|}{{\color[HTML]{000000} Real-world datasets}}                                                                                                                                                                                                                                                                                                                                  \\ \hline
{\color[HTML]{000000} Algorithm}  & \multicolumn{1}{c|}{{\color[HTML]{000000} Sea}}                   & \multicolumn{1}{c|}{{\color[HTML]{000000} Sine}}                  & \multicolumn{1}{c|}{{\color[HTML]{000000} Circle}}                & {\color[HTML]{000000} Vib}                   & \multicolumn{1}{c|}{{\color[HTML]{000000} Fraud}}                 & \multicolumn{1}{c|}{{\color[HTML]{000000} MNIST-01}}              & \multicolumn{1}{c|}{{\color[HTML]{000000} MNIST-23}}              & \multicolumn{1}{c|}{{\color[HTML]{000000} MNIST-multi}}           & \multicolumn{1}{c|}{{\color[HTML]{000000} Forest}}                & {\color[HTML]{000000} Arrhy}                 \\ \hline
{\color[HTML]{000000} Baseline}   & \multicolumn{1}{c|}{{\color[HTML]{000000} 0.633(0.000)}}          & \multicolumn{1}{c|}{{\color[HTML]{000000} 0.514(0.000)}}          & \multicolumn{1}{c|}{{\color[HTML]{000000} 0.467(0.000)}}          & {\color[HTML]{000000} 0.557(0.000)}          & \multicolumn{1}{c|}{{\color[HTML]{000000} 0.895(0.000)}}          & \multicolumn{1}{c|}{{\color[HTML]{000000} 0.786(0.000)}}          & \multicolumn{1}{c|}{{\color[HTML]{000000} 0.731(0.000)}}          & \multicolumn{1}{c|}{{\color[HTML]{000000} 0.550(0.000)}}          & \multicolumn{1}{c|}{{\color[HTML]{000000} 0.490(0.000)}}          & {\color[HTML]{000000} 0.629(0.000)}          \\ \hline
{\color[HTML]{000000} iForest++}  & \multicolumn{1}{c|}{{\color[HTML]{000000} 0.840(0.018)}} & \multicolumn{1}{c|}{{\color[HTML]{000000} \textbf{0.927(0.020)}}} & \multicolumn{1}{c|}{{\color[HTML]{000000} 0.770(0.018)}}          & {\color[HTML]{000000} 0.441(0.017)}          & \multicolumn{1}{c|}{{\color[HTML]{000000} \textbf{0.925(0.015)}}} & \multicolumn{1}{c|}{{\color[HTML]{000000} 0.696(0.024)}}          & \multicolumn{1}{c|}{{\color[HTML]{000000} \textbf{0.880(0.032)}}} & \multicolumn{1}{c|}{{\color[HTML]{000000} 0.538(0.026)}}          & \multicolumn{1}{c|}{{\color[HTML]{000000} 0.508(0.023)}}          & {\color[HTML]{000000} 0.586(0.016)}          \\ \hline
{\color[HTML]{000000} LOF++}      & \multicolumn{1}{c|}{{\color[HTML]{000000} 0.829(0.000)}}          & \multicolumn{1}{c|}{{\color[HTML]{00009B} \textbf{0.921(0.000)}}} & \multicolumn{1}{c|}{{\color[HTML]{000000} 0.710(0.000)}}          & {\color[HTML]{000000} 0.525(0.000)}          & \multicolumn{1}{c|}{{\color[HTML]{000000} 0.913(0.000)}}          & \multicolumn{1}{c|}{{\color[HTML]{000000} 0.708(0.000)}}          & \multicolumn{1}{c|}{{\color[HTML]{000000} 0.802(0.000)}}          & \multicolumn{1}{c|}{{\color[HTML]{000000} 0.456(0.000)}}          & \multicolumn{1}{c|}{{\color[HTML]{000000} 0.500(0.000)}}          & {\color[HTML]{000000} 0.616(0.000)}          \\ \hline
{\color[HTML]{000000} ARCUS}      & \multicolumn{1}{c|}{{\color[HTML]{000000} 0.577(0.041)}}          & \multicolumn{1}{c|}{{\color[HTML]{000000} 0.735(0.007)}}          & \multicolumn{1}{c|}{{\color[HTML]{000000} 0.529(0.022)}}          & {\color[HTML]{000000} 0.557(0.018)}          & \multicolumn{1}{c|}{{\color[HTML]{000000} 0.907(0.024)}}          & \multicolumn{1}{c|}{{\color[HTML]{000000} 0.578(0.019)}}          & \multicolumn{1}{c|}{{\color[HTML]{000000} 0.628(0.020)}}          & \multicolumn{1}{c|}{{\color[HTML]{00009B} \textbf{0.674(0.016)}}} & \multicolumn{1}{c|}{{\color[HTML]{000000} 0.486(0.025)}}          & {\color[HTML]{000000} 0.465(0.026)}          \\ \hline
{\color[HTML]{000000} Memstream}  & \multicolumn{1}{c|}{{\color[HTML]{000000} 0.790(0.104)}}          & \multicolumn{1}{c|}{{\color[HTML]{000000} 0.891(0.067)}}          & \multicolumn{1}{c|}{{\color[HTML]{000000} \textbf{0.823(0.066)}}} & {\color[HTML]{000000} 0.512(0.057)}          & \multicolumn{1}{c|}{{\color[HTML]{000000} 0.891(0.014)}}          & \multicolumn{1}{c|}{{\color[HTML]{000000} 0.586(0.008)}}          & \multicolumn{1}{c|}{{\color[HTML]{000000} 0.719(0.005)}}          & \multicolumn{1}{c|}{{\color[HTML]{000000} 0.574(0.006)}}          & \multicolumn{1}{c|}{{\color[HTML]{000000} 0.499(0.024)}}          & {\color[HTML]{000000} \textbf{0.781(0.008)}} \\ \hline
{\color[HTML]{000000} METER}      & \multicolumn{1}{c|}{{\color[HTML]{000000} 0.509(0.049)}}          & \multicolumn{1}{c|}{{\color[HTML]{000000} 0.504(0.021)}}          & \multicolumn{1}{c|}{{\color[HTML]{000000} 0.522(0.064)}}          & {\color[HTML]{000000} 0.453(0.001)}          & \multicolumn{1}{c|}{{\color[HTML]{000000} 0.905(0.019)}}          & \multicolumn{1}{c|}{{\color[HTML]{000000} 0.528(0.000)}}          & \multicolumn{1}{c|}{{\color[HTML]{000000} 0.562(0.001)}}          & \multicolumn{1}{c|}{{\color[HTML]{000000} 0.510(0.000)}}          & \multicolumn{1}{c|}{{\color[HTML]{000000} 0.490(0.019)}}          & {\color[HTML]{000000} 0.604(0.000)}          \\ \hline
{\color[HTML]{000000} SEAD}       & \multicolumn{1}{c|}{{\color[HTML]{000000} 0.495(0.075)}}          & \multicolumn{1}{c|}{{\color[HTML]{000000} 0.504(0.077)}}          & \multicolumn{1}{c|}{{\color[HTML]{000000} 0.508(0.085)}}          & {\color[HTML]{000000} 0.515(0.068)}          & \multicolumn{1}{c|}{{\color[HTML]{000000} 0.476(0.071)}}          & \multicolumn{1}{c|}{{\color[HTML]{000000} 0.473(0.088)}}          & \multicolumn{1}{c|}{{\color[HTML]{000000} 0.566(0.002)}}          & \multicolumn{1}{c|}{{\color[HTML]{000000} 0.494(0.089)}}          & \multicolumn{1}{c|}{{\color[HTML]{000000} 0.491(0.081)}}          & {\color[HTML]{000000} 0.429(0.069)}          \\ \hline
{\color[HTML]{000000} CPOCEDS}    & \multicolumn{1}{c|}{{\color[HTML]{000000} \textbf{0.941(0.000)}}}          & \multicolumn{1}{c|}{{\color[HTML]{000000} 0.836(0.000)}}          & \multicolumn{1}{c|}{{\color[HTML]{000000} 0.622(0.000)}}          & {\color[HTML]{000000} 0.424(0.000)}          & \multicolumn{1}{c|}{{\color[HTML]{000000} 0.904(0.000)}}          & \multicolumn{1}{c|}{{\color[HTML]{000000} 0.521(0.000)}}          & \multicolumn{1}{c|}{{\color[HTML]{000000} 0.752(0.000)}}          & \multicolumn{1}{c|}{{\color[HTML]{000000} 0.540(0.000)}}          & \multicolumn{1}{c|}{{\color[HTML]{000000} 0.490(0.000)}}          & {\color[HTML]{000000} 0.596(0.000)}          \\ \hline
{\color[HTML]{000000} StrAEm++DD} & \multicolumn{1}{c|}{{\color[HTML]{000000} 0.835(0.000)}}          & \multicolumn{1}{c|}{{\color[HTML]{000000} 0.886(0.000)}}          & \multicolumn{1}{c|}{{\color[HTML]{000000} 0.638(0.000)}}          & {\color[HTML]{000000} 0.663(0.000)}          & \multicolumn{1}{c|}{{\color[HTML]{000000} 0.899(0.000)}}          & \multicolumn{1}{c|}{{\color[HTML]{00009B} \textbf{0.822(0.000)}}} & \multicolumn{1}{c|}{{\color[HTML]{000000} 0.838(0.000)}}          & \multicolumn{1}{c|}{{\color[HTML]{000000} 0.673(0.000)}}          & \multicolumn{1}{c|}{{\color[HTML]{00009B} \textbf{0.629(0.000)}}} & {\color[HTML]{000000} 0.536(0.000)}          \\ \hline
{\color[HTML]{000000} VAE++ES}    & \multicolumn{1}{c|}{{\color[HTML]{000000} 0.749(0.000)}}          & \multicolumn{1}{c|}{{\color[HTML]{000000} 0.800(0.000)}}          & \multicolumn{1}{c|}{{\color[HTML]{000000} 0.532(0.000)}}          & {\color[HTML]{00009B} \textbf{0.746(0.000)}} & \multicolumn{1}{c|}{{\color[HTML]{000000} 0.901(0.000)}}          & \multicolumn{1}{c|}{{\color[HTML]{000000} 0.807(0.000)}}          & \multicolumn{1}{c|}{{\color[HTML]{000000} 0.843(0.000)}}          & \multicolumn{1}{c|}{{\color[HTML]{000000} 0.593(0.000)}}          & \multicolumn{1}{c|}{{\color[HTML]{000000} 0.509(0.000)}}          & {\color[HTML]{000000} 0.634(0.000)}          \\ \hline
{\color[HTML]{000000} VAE++ESDD}  & \multicolumn{1}{c|}{{\color[HTML]{00009B} \textbf{0.869(0.000)}}} & \multicolumn{1}{c|}{{\color[HTML]{000000} 0.903(0.000)}}          & \multicolumn{1}{c|}{{\color[HTML]{00009B} \textbf{0.788(0.000)}}} & {\color[HTML]{000000} \textbf{0.760(0000)}}  & \multicolumn{1}{c|}{{\color[HTML]{00009B} \textbf{0.922(0.000)}}} & \multicolumn{1}{c|}{{\color[HTML]{000000} \textbf{0.828(0.000)}}} & \multicolumn{1}{c|}{{\color[HTML]{00009B} \textbf{0.863(0.000)}}} & \multicolumn{1}{c|}{{\color[HTML]{000000} \textbf{0.783(0.000)}}} & \multicolumn{1}{c|}{{\color[HTML]{000000} \textbf{0.756(0.000)}}} & {\color[HTML]{00009B} \textbf{0.759(0.000)}} \\ \hline
\end{tabular}
\end{adjustbox}
\end{table*}

\subsection{Comparative study}\label{sec:compare}

This section presents a comparison of Baseline, VAE++ES, VAE++ESDD, StrAem++DD, iForest++, LOF++, ARCUS, METER, SEAD, CPOCEDS, and MemStream under severe class imbalance. Method details are provided in Sec.~\ref{sec:exp_setup}. Tables~\ref{tab:compare_gmean}–\ref{tab:compare_pauc} report the average G-mean, Recall, Specificity, and PAUC over the entire stream. For fair comparison, all baselines were re-evaluated using author-recommended settings with additional hyperparameter tuning, and the best-performing configurations are reported.

As shown in all tables, VAE++ESDD consistently outperforms StrAEm++DD, demonstrating the benefits of the proposed two-level ensembling mechanism. This trend is also evident in Fig.~\ref{fig:compare_recu}.

As shown in Table~\ref{tab:compare_gmean}, the proposed method achieves consistently high G-mean with zero standard deviation across runs, indicating strong robustness to random initialization. VAE++ESDD ranks among the top performers across all datasets and drift types. It can also handle recurrent concept drift (Fig.~\ref{fig:compare_recu}) provided that sufficient data are available between drift events to retrain the VAE and resume incremental learning. This assumption—that recurrent drifts do not occur too frequently—is standard in buffer-based streaming approaches.

Tables~\ref{tab:compare_recall} and~\ref{tab:compare_speci} show that several methods excel in one metric but fail in the other. For instance, MemStream achieves high specificity (0.950) but low recall (0.173) on MNIST-23, while iForest attains perfect recall (1.000) but low specificity (0.585) on Sine. In contrast, VAE++ESDD maintains recall above 0.75 and specificity above 0.65 across all datasets, indicating a more balanced trade-off. High specificity with low recall leads to missed anomalies, whereas high recall with low specificity causes frequent false alarms; neither is acceptable in practical monitoring scenarios. Achieving a balanced recall–specificity trade-off is therefore critical for reliable real-world anomaly detection.

The Average PAUC performance is summarized in Table~\ref{tab:compare_pauc}, where VAE++ESDD ranks within the top two methods in most cases. A higher PAUC reflects a model’s threshold-independent discriminative power, indicating its ability to consistently distinguish between normal and abnormal samples across time. This metric captures both the accuracy and stability of predictions in streaming environments.
In contrast, G-mean evaluates performance at a specific decision threshold, focusing on the balance between recall and specificity. Therefore, combining the two metrics provides a more comprehensive view of performance. The consistently high PAUC and G-mean values demonstrate that VAE++ESDD achieves both strong discriminative capability and robust adaptability, effectively handling anomaly detection and concept drift in dynamic data streams.

\section{Conclusion}\label{sec:conclusion}
Extracting patterns from data streams presents significant challenges, including the identification of anomalous events, the unavailability of ground truth, and adaptation to nonstationary environments. To address these challenges, we introduce one innovative approach, namely VAE++ESDD, which takes advantage of ensemble learning, VAE-based incremental learning, and drift detection, thus extending the application of autoencoders to online learning and incorporates a hybrid active-passive approach to handle concept drift. The proposed method can effectively identify anomalous events, detect concept drift, and does not rely on supervision. An extensive experimental study is conducted to evaluate the performance of VAE++ESDD. The results demonstrate that our proposed method VAE++ESDD outperforms existing baseline and advanced methods.

In this work, a new model is created and retrained once concept drift is detected. Although extensive experiments demonstrate strong performance, the current approach does not explicitly exploit the knowledge retained in previously learned models. Consequently, the proposed method is best suited for scenarios in which data distributions evolve continuously and rarely revert to earlier states. In contrast, under recurrent drift conditions, storing and selectively reusing historical models could reduce computational cost and help mitigate potential catastrophic forgetting. As part of future work, we aim to develop a more adaptive update strategy that selectively leverages past models and resets only the components affected by the detected drift, thereby improving both efficiency and adaptability. The current implementation does not distinguish between different types of drift, such as feature drift and concept drift. As part of future work, we plan to incorporate drift-type differentiation into the adaptation strategy, enabling more targeted, efficient, and drift-specific update mechanisms.

\bibliographystyle{model1-num-names}

\bibliography{cas-refs}





\end{document}